%% file: main.tex
\documentclass{article}

\PassOptionsToPackage{numbers,compress}{natbib}
\usepackage[preprint,main]{neurips_2026}
\makeatletter
\renewcommand{\@noticestring}{Preprint.}
\makeatother

\usepackage[utf8]{inputenc}
\usepackage[T1]{fontenc}
\usepackage{hyperref}
\hypersetup{hidelinks}
\usepackage{url}
\usepackage{booktabs}
\usepackage{array}
\usepackage{amsfonts}
\usepackage{amsmath}
\usepackage{amssymb}
\usepackage{graphicx}
\usepackage{nicefrac}
\usepackage{microtype}
\usepackage{xcolor}
\usepackage{enumitem}
\title{Precise but Uncoupled: Reviewer Precision Does Not Guarantee Critique Uptake in Multi-Agent Math Reasoning}

\author{%
Chih\mbox{-}Hsuan Yang\thanks{Corresponding author: \texttt{bellayang@anl.gov}.}\\
Argonne National Laboratory
\And
Jingyan Jiang\\
Argonne National Laboratory
\And
Vikram Vasudevan\\
Oregon State University
\AND
Cheng\mbox{-}Hau Yang\\
Argonne National Laboratory
\And
Huihuo Zheng\\
Argonne National Laboratory
\And
Le Chen\\
Argonne National Laboratory
\AND
Eliu A. Huerta\\
Argonne National Laboratory\\
University of Chicago
\And
Venkatram Vishwanath\\
Argonne National Laboratory
\And
Ian T. Foster\\
Argonne National Laboratory\\
University of Chicago
\AND
Rajeev Thakur\\
Argonne National Laboratory
}

\begin{document}

\maketitle

\begin{abstract}
\input{sections/00_abstract}
\end{abstract}

\input{sections/01_introduction}
\input{sections/02_related_work}
\input{sections/03_task_setting_and_protocols}
\input{sections/05_results}
\input{sections/06_conclusion}
\input{sections/07_limitations_and_broader_impact}

\section*{Acknowledgments}

This research used resources of the Argonne Leadership Computing Facility, a U.S. Department of Energy (DOE) Office of Science user facility at Argonne National Laboratory (ANL) operated under Contract No.~DE-AC02-06CH11357. The work was also supported under the same contract by the DOE Office of Science's Advanced Scientific Computing Research Program and by Laboratory Directed Research and Development (LDRD) funding from ANL, provided by the Director, DOE Office of Science.

\bibliographystyle{plainnat}
\bibliography{references}

\appendix
\input{sections/08_appendix}

\clearpage
\input{checklist}

\end{document}

%% file: sections/00_abstract.tex

Many recent math- and science-oriented agent systems adopt hierarchical designs with specialized reviewer roles, motivated by the idea that routing critique through a dedicated review stage should help turn wrong candidates into correct ones on hard problems. We study that expectation on 4{,}181 verifier-grounded Omni-MATH problems~\cite{omnimath,omnimath2}, a hard ten-tier benchmark with enough headroom to separate protocols, using matched \texttt{gpt-oss-120b} actors as the primary actor family. On tiers 1--2, multi-agent collaboration adds at most about 2 percentage points over the matched single-agent baseline. From tier 4 onward, the gains open sharply, reaching about 10 to 20 percentage points on tiers 6--9. In that regime, broadcast-style peer discussion attains higher final accuracy than a planner--executor--reviewer pipeline (PER). We use that divergence as a starting observation and ask when reviewer quality translates into effective solver updates. On this benchmark, the PER--broadcast accuracy gap is not accounted for by reviewer precision alone. Here \emph{reviewer precision} asks how often a reviewer warning truly points to a real error. PER's reviewer has higher precision (0.861 vs.\ 0.644), yet correct critique is much less likely to change the next candidate the protocol carries forward, and reviewer-guided repair is correspondingly lower. These results show that reviewer detection quality and critique uptake are empirically separable. Within matched PER interventions, forcing explicit acknowledgment lowers final accuracy instead of improving follow-through, while placing reviewer guidance directly in the solver's working context partially improves follow-through without closing the gap. Taken together, these interventions point in the same direction: critique appears more likely to be acted upon when it is presented more directly in the solver's working context, although this evidence is directional rather than causal. Under reviewer-centric evaluation, a system can look strong at spotting errors yet still fail to solve more problems if the protocol does not act on those critiques.

%% file: sections/01_introduction.tex

\section{Introduction}

\paragraph{A common design choice: use a specialized reviewer.}
Recent work on LLM reasoning has moved from single-chain prompting toward richer inference-time systems with multiple interacting calls, branches, or agents \cite{multiagentdebate,autogen,camel,masurvey,zhang2025ttssurvey}. Within that broader landscape, open peer deliberation---that is, agents publicly critiquing and revising a shared candidate---remains influential \cite{multiagentdebate,canagentsdebate}, but many recent math- and science-oriented systems adopt hierarchical designs with specialized reviewer roles. MALT \cite{malt} decomposes reasoning into generator, verifier, and refinement agents trained end-to-end; DPSDP \cite{dpsdp} trains an actor--critic pair through reinforcement learning; and recent surveys and scaling analyses argue that role heterogeneity can be a primary source of multi-agent gains \cite{scientificagentssurvey,scalingagentsystems,aiagentsmatter}. A common operational expectation in these systems is that routing critique through a dedicated review stage should improve repair---that is, convert a wrong candidate into a correct one---on hard problems. We refer to that more specific expectation as the \emph{role-specialization hypothesis}: when critique matters, separating reviewer and solver roles should improve repair.

\paragraph{A hidden failure mode: reviewer--solver coupling.}
That expectation leaves out a second question. A reviewer can detect an error correctly, and the downstream solver can still fail to act on that critique. Throughout the paper, \emph{solver} means the role that produces the next candidate answer; in PER this is the Executor role. We call the resulting failure \emph{reviewer--solver decoupling}. This failure mode remains hard to detect if evaluation reports only final pass rate or reviewer precision \cite{cannotselfcorrect,correctbench,decomposingselfcorrection,whymasfail}. We make that failure visible by separating reviewer quality from solver response: on evaluator-verified useful critique, does the next candidate change, and does that change lead to repair?

\begin{figure}[t]
\centering
\small
\setlength{\tabcolsep}{4pt}
\renewcommand{\arraystretch}{1.15}
\begin{tabular}{p{0.47\linewidth}p{0.47\linewidth}}
\toprule
\textbf{PER: critique is routed, but can remain non-binding} & \textbf{Broadcast: critique is shared, and approval is collective} \\
\midrule
Planner $\rightarrow$ Executor $\rightarrow$ Reviewer &
Peer discussion $\leftrightarrow$ shared candidate \\
Reviewer signal enters a separable \texttt{advice} field &
Critique enters the shared deliberation state \\
Solver can acknowledge the signal yet preserve the candidate &
Every peer sees the same candidate before submission \\
Submission decision is role-local after routed review &
Submission requires collective re-approval \\
\bottomrule
\end{tabular}
\caption{Why reviewer quality and realized repair can diverge. In PER, information flow and decision flow can separate: a reviewer can send a correct warning without changing the next candidate. In broadcast, critique is embedded in shared candidate state and group approval, making bypass harder by design.}
\label{fig:coupling_schematic}
\end{figure}

\paragraph{Main finding.}
We evaluate four protocols on 4{,}181 problems from the Omni-MATH benchmark family \cite{omnimath,omnimath2}, an olympiad-level setting where substantial headroom remains under current frontier models \cite{omnimath,omnimath2}. On the easiest tiers, collaboration adds little. Once the benchmark becomes hard enough for collaboration to matter, however, the two collaborative designs begin to diverge: broadcast achieves higher final accuracy than PER. We treat that divergence as the starting point for a process analysis: does the gap reflect reviewer quality, or does useful critique fail to reach the solver's next candidate? On reviewer \emph{detection}, PER's reviewer has higher precision than broadcast's (0.861 vs.\ 0.644). On reviewer-to-action \emph{uptake}, however, the ordering reverses: when the reviewer gives critique that the evaluator later confirms as useful, the protocol changes its carried-forward candidate only 33.6\% of the time in PER versus 93.5\% in broadcast, and ReviewerGuidedRepairRate is 0.051 versus 0.286. Under this evaluation, the PER--broadcast accuracy gap is therefore not accounted for by reviewer precision alone and is instead consistent with differences in critique uptake.

Relative to the matched single-agent iterative baseline, the collaboration gains open sharply from tier 4 onward. We track that gap with three primary diagnostics: whether useful critique is ignored (\emph{NeglectRate}), whether the next answer changes (\emph{CouplingRate}), and whether that change repairs the problem (\emph{ReviewerGuidedRepairRate}). Within PER, ACK-required is a failed compliance-style control. EMB, which presents reviewer guidance directly in the solver's working context, recovers only part of the PER--broadcast gap, as does a deeper-reflection variant. We treat the within-PER probes as directional support for this pattern, not as isolated causal proof.

\paragraph{Why Omni-MATH?}
On standard easy math benchmarks, strong systems are already close to ceiling, so it is harder to tell whether collaboration is helping. Omni-MATH gives this paper room to separate protocols: it contains 4{,}181 competition-level problems, remains far from saturated under current evaluations, and supplies ten benchmark-defined difficulty tiers rather than a scale tuned to one model family \cite{omnimath,omnimath2}.

\paragraph{The single-agent anchor.}
We read the tier curves relative to \emph{Single-Agent Iterative}, which is the paper's matched non-collaborative anchor. The single-agent baseline already has iterative opportunity and verifier interaction, so gain over Single-Agent isolates what deliberation and review add rather than mixing collaboration with the value of simply getting more attempts.

\paragraph{Scope of the claim.}
This is a \emph{test-time} study: we keep the underlying model family fixed and vary the protocol used at inference time, including role structure, feedback routing, and retry behavior. Our claim is not that one protocol universally wins, nor that PER--broadcast alone isolates a single causal variable. The PER--broadcast comparison is evidence about complete protocol designs that differ in shared state, approval rules, and discussion structure. Within that comparison, reviewer precision and critique uptake can separate sharply under a common PER-style reviewer-to-solver interface; the within-PER probes provide more limited direction-of-change evidence on how feedback delivery relates to the uptake gap. The appendix carries the fuller recovery, cost, and protocol-control details: a cross-evaluator disagreement replay, a higher-cost PER-inner6 stress test, and a reduced cross-family replication with Gemma~3 actors in which the overall ranking shifts but the precision--uptake separation persists, suggesting that uptake is an important evaluation axis rather than a complete account of protocol quality.

\paragraph{Contributions.}
\noindent We make four integrated contributions:
\begin{enumerate}[leftmargin=1.4em,itemsep=0pt]
  \item \textbf{Finding.} In this matched benchmark/model regime, hierarchical PER attains lower repair and lower final accuracy than broadcast even though PER's reviewer has higher precision; the clearest observable difference is much lower uptake of useful critique.
  \item \textbf{Interventions.} Within-PER probes that hold the model family, reviewer backbone, and budget fixed provide directional evidence consistent with the interface story: ACK-required is a failed compliance control, and EMB partially recovers the gap by placing reviewer guidance directly in the solver's working context.
  \item \textbf{Measurement.} We introduce a measurement framework that separates within-attempt review from verifier-guided repair, together with coupling-sensitive metrics that make detection, uptake, and realized repair measurable under matched protocol comparisons.
  \item \textbf{Implications.} The same analysis shifts evaluation away from reviewer quality alone and toward whether the protocol acts on critique, while the appendix gives a verifier-cost-aware deployment view for this benchmark regime rather than a universal protocol winner.
\end{enumerate}

%% file: sections/02_related_work.tex

\section{Related Work}

\paragraph{Inference-time structure and multi-agent reasoning.}
Prior work shows that reasoning can improve when inference is given more structure, whether through deliberate reasoning traces, branching search, or multi-agent interaction \cite{wei2022chainofthought,wang2023selfconsistency,yao2023treeofthoughts,multiagentdebate,autogen,camel}. Within multi-agent systems, open peer deliberation remains influential, but recent math- and science-oriented systems often move toward specialized or hierarchical role assignments \cite{malt,dpsdp,scientificagentssurvey,scalingagentsystems,aiagentsmatter}. Our paper focuses on one narrow question inside that trend: when a system includes a specialized reviewer, should reviewer quality alone predict whether collaboration helps?

\paragraph{Collaboration value.}
Recent agent evaluations also make a broader point: adding interaction or extra agents is not useful by itself unless collaboration improves outcomes relative to a matched non-collaborative baseline, and does so at acceptable coordination or verifier cost \cite{agentbench,aiagentsmatter,scalingagentsystems}. Our paper adopts that lens directly. We compare collaborative protocols not only by absolute accuracy, but also by gain over Single-Agent Iterative, which already has retry opportunity and verifier interaction. This lets us ask not only whether a protocol solves more problems, but whether deliberation itself adds value and through which stage that value is realized.

\paragraph{Self-correction, verifier-guided repair, and separable stages.}
A separate line of work shows that detection, revision, and successful repair are not the same event \cite{selfrefine,reflexion,correctbench,iterativeselfrepair,cannotselfcorrect,decomposingselfcorrection}. These results are important because they move evaluation away from the assumption that noticing an error is already most of the work. We extend that logic to a multi-agent reviewer-to-solver interface: a reviewer can identify a valid problem, yet the downstream solver can still fail to turn that critique into a better next candidate.

\paragraph{Process-aware evaluation.}
Recent evaluation work also argues that answer-only metrics miss important process differences, especially in agent systems where interaction, cost, and evaluator design matter \cite{evalreasonaction,agentbench,aiagentsmatter}. Our contribution is narrower than a benchmark or taxonomy paper. We focus on evaluator-verified useful critique and ask three linked questions: was the critique right, did it change the next candidate, and did that change repair the problem? Appendix~\ref{app:literature_map} gives the fuller literature map and term lineage.

%% file: sections/03_task_setting_and_protocols.tex

\section{Task Setting and Evaluation}
\label{sec:task_setting}

\subsection{Benchmark: Omni-MATH 2}
\label{ssec:omnimath2}
We use Omni-MATH~2 \cite{omnimath2}, a cleaned and filtered version of Omni-MATH \cite{omnimath}. Our filtered corpus contains 4{,}181 competition-level mathematics problems organized into ten benchmark-provided difficulty tiers, with tier~10 the hardest. Each problem includes evaluator-usable final answers and reference material, which lets us study verifier-grounded recovery rather than free-form judge preferences.

This benchmark is useful here for two reasons. First, it remains far from saturated under current evaluations \cite{omnimath,omnimath2}, so protocol differences still have room to appear on both first-pass and multi-attempt outcomes. Second, the benchmark-provided tier structure lets us ask \emph{where} collaboration matters, not only whether one protocol has the highest aggregate score. We therefore use Omni-MATH not as a generic math benchmark, but as a hard setting in which reviewer precision, critique uptake, and verifier-guided repair can separate. Tier counts and broader literature context are deferred to Appendix~\ref{app:tier_breakdown} and Appendix~\ref{app:literature_context}.

\subsection{Task Setting}
We study verifier-gated mathematical reasoning on exact-answer or equivalence-checkable olympiad problems \cite{omnimath,omnimath2}. A protocol may deliberate internally, submit a final answer, receive evaluator feedback, and optionally retry under a fixed outer-loop budget. Throughout the paper, \emph{protocol} means the full interaction procedure: agent roles, communication topology, revision policy, evaluator-submission rule, and feedback-routing rule.

This paper separates two stages that are easy to conflate if we look only at final accuracy. The first stage is \emph{within-attempt review}: before any external verdict, agents may critique a current candidate and revise it. The second stage is \emph{verifier-guided retry}: after a failed submission, the protocol may use evaluator feedback to attempt recovery on a later outer-loop attempt. That distinction matters because a protocol can appear strong for different reasons: it can produce better first-pass candidates, recover better after failure, or spend substantially more verifier interaction to reach the same endpoint. The main paper therefore keeps outcomes, review-to-action transitions, and verifier burden separate.

\subsection{Compared Protocols}
We compare four protocols arranged as a ladder from no collaboration to open peer deliberation. \textbf{Baseline LLM} is a direct one-shot solver with no iterative repair memory and no explicit interaction. \textbf{Single-Agent Iterative} uses the same base model but allows repeated outer-loop attempts with question-scoped failure and candidate-revision memory, without introducing a separate reviewer. Our \textbf{structured role-based MAS} instantiates a Planner--Executor--Reviewer pipeline (PER): the planner decomposes the problem, the executor produces the next candidate answer, and the reviewer decides whether that candidate is ready for evaluator submission or should be sent back for revision. In later sections we use \emph{solver} as the name for this candidate-producing role. Finally, \textbf{broadcast deliberation} uses three peers in a hybrid interaction pattern: confidence polling allocates one public speaker at a time during discussion, but once a current candidate exists, all three peers enter an approval phase that can approve, reject, or revise that candidate before submission; under our matched configuration, submission requires unanimous approval \cite{multiagentdebate,autogen,camel,masurvey}.

These four protocols let us separate three things that are often mixed together in agent papers: the value of getting more attempts, the value of adding a specialized reviewer, and the value of making critique part of a shared deliberation state. PER and broadcast instantiate two common design patterns: role-specialized review pipelines and shared-state peer deliberation. PER routes reviewer feedback as a separable signal inside a hierarchical pipeline, whereas broadcast embeds critique inside shared peer context and treats the current candidate as protocol-level state.

\subsection{Metrics, Controls, and Evaluator}
\label{ssec:model_evaluator}
The paper keeps three kinds of quantity in view. \emph{Outcome} asks whether the protocol solves the problem. \emph{Mechanism} asks what happened between critique and the next candidate. \emph{Verifier burden} asks how much external feedback the protocol consumes to get there. We summarize outcomes with FinalPassRate and Pass@1. We summarize mechanism with NeglectRate, CouplingRate, and ReviewerGuidedRepairRate, which follow the chain detection $\rightarrow$ uptake $\rightarrow$ repair: did the reviewer correctly flag a real problem, did that critique change the solver's next candidate, and did the change actually fix the problem? We track verifier burden with AvgEvalCalls/problem, the average number of evaluator (verifier) calls per problem, because protocol differences in recovery and coordination can shift verifier cost even when token cost rises more slowly. CouplingRate is evaluator-grounded because it is computed only on evaluator-verified useful revise episodes. It is an operational answer-transition statistic, not a semantic measure of reasoning alignment: it asks whether reviewer feedback changes the next candidate the protocol actually carries forward, which is a necessary but not sufficient condition for critique utilization. In that sense it is a protocol-level uptake or binding metric: under broadcast, uptake can be realized through shared candidate updates and collective re-approval, whereas under PER it depends more on routed reviewer-to-solver follow-through. Appendix~\ref{app:metrics} gives the exact definition together with deterministic normalization and equivalence-aware audits, and shows that the observed uptake gap is stable under those alternative answer-change definitions.

Across the matched configurations, all acting agents and the evaluator use \texttt{gpt-oss-120b}\footnote{Specifically, the public OpenAI checkpoint \texttt{openai/gpt-oss-120b}, released under Apache 2.0 and downloadable from \url{https://huggingface.co/openai/gpt-oss-120b}; see the official release and model documentation in \citet{openai_gptoss_release,openai_gptoss_model}.}, a public open-weight reasoning model, served through public GPU-backed HPC infrastructure via an OpenAI-compatible server, with temperature $0$ throughout. We also hold the evaluator contract, outer-loop budget, and memory reset policy fixed across protocols. Even with those controls, PER and broadcast still differ in how the full system operates: who can see the current candidate, how agents communicate, how discussion proceeds, and who must approve before submission. We therefore use the PER--broadcast comparison as evidence about whole protocol designs, not as an isolation of one factor at a time. More targeted direction-of-change evidence comes from the within-PER probes. As a robustness check, Appendix~\ref{app:evaluator} replays all saved submitted answers with two other open evaluators under the same prompt/parser; pairwise disagreement remains low on the full benchmark (3.38\%--5.76\%; $\kappa=0.850$--$0.915$), where $\kappa$ denotes Cohen's chance-corrected agreement coefficient \cite{cohen1960agreement}. That range is generally read as strong agreement \cite{landis1977observer,mchugh2012kappa}, so we treat the replay as trustworthy as a robustness check rather than as a replacement judge; disagreement rises mainly on the hard collaborative slice where answer equivalence is most brittle.

To probe the reviewer-to-solver interface more directly, we also test two full-benchmark PER variants while holding the model family, reviewer setup, and outer-loop budget fixed. ACK-required adds mandatory verbal acknowledgment of critique while keeping routed advice unchanged; EMB places reviewer guidance directly in the solver working context. Appendix~\ref{app:coupling_probe_details} gives the exact interface changes.

%% file: sections/05_results.tex

\section{Results}
\label{sec:results}

We organize the evaluation around four questions: outcomes over a matched single-agent baseline (RQ1), post-failure recovery (RQ2), whether the PER--broadcast accuracy gap is better explained by reviewer detection quality or critique uptake (RQ3), and how protocol choice changes when verifier calls are cheap or expensive (RQ4).

\subsection{Outcomes, Cost, and Recovery}
\label{ssec:rq1}

Across all 4{,}181 Omni-MATH problems, the three multi-step protocols substantially exceed the one-shot baseline. Within this matched setting, broadcast achieves the highest FinalPassRate and Pass@1, followed by PER and then the single-agent iterative baseline (Table~\ref{tab:main_outcome_cost}). We treat that ranking as the starting observation for the process analysis below, not as a universal verdict on protocol design.

Omni-MATH defines these ten tiers independently of the models we evaluate, so the clearer tier-level comparison is gain over Single-Agent rather than raw protocol rank. Excluding the tiny tails at tiers 3 and 10, collaboration gains are near zero on tiers 1--2 and then open sharply from tier 4 onward. On tiers 6--9, PER improves FinalPassRate by about 10 to 16 percentage points over Single-Agent, while Broadcast improves it by about 14 to 20 percentage points. Pass@1 gains follow the same pattern, showing that the hard-tier value of collaboration appears before verifier-guided retries.

In this benchmark/model regime, Single-Agent Iterative is the efficiency leader, PER occupies a higher-cost middle position, and broadcast is the accuracy leader at the high-cost end. PER uses more evaluator calls per problem than broadcast (2.19 vs.\ 1.35), so it is verifier-heavier despite generating fewer tokens overall.

Broadcast's aggregate edge comes mostly from avoiding first failure rather than from dramatically stronger post-failure recovery. After one failure, broadcast and Single-Agent have nearly identical conditional recovery on the second attempt, at 37.2\% and 37.0\%, while PER recovers less efficiently at 32.4\%. Single-Agent leads on cost-per-solve, while broadcast becomes the better choice once verifier cost makes Pass@1 the primary objective; Appendix~\ref{app:efficiency} and Appendix~\ref{app:verifier_cost_model} give the fuller deployment view.

\begin{table}[t]
\centering
\small
\caption{Overall outcome and cost comparison over $N$ = 4{,}181 problems. FinalPassRate and Pass@1 are computed over all problems, not averaged by tier. Pass@1 is the fraction of problems solved on the first submitted attempt, so across protocols it reflects each protocol's own pre-submission deliberation budget rather than a compute-matched one-step score. ``Tokens per Extra Solve'' means the average number of generated tokens required to solve one additional problem relative to Baseline LLM, the one-shot entry point for the overall cost ladder; tier-level collaboration gains elsewhere are reported relative to Single-Agent Iterative. ``Average evaluator calls'' means the average number of evaluator (verifier) calls per problem. Appendix Figure~\ref{fig:supp_rq_overview} gives a fuller multi-metric view.}
\label{tab:main_outcome_cost}
\resizebox{\linewidth}{!}{
\begin{tabular}{lccccc}
\toprule
Protocol & FinalPassRate & Pass@1 & Avg. Tokens & Tokens per Extra Solve & Avg. Evaluator Calls \\
\midrule
Baseline LLM (one-shot)  & 56.8\% & 56.8\% & \phantom{00}18{,}385 & ---  & 1.00 \\
Single-Agent Iterative   & 78.8\% & 57.4\% & \phantom{0}48{,}123 & 135K & 1.69 \\
PER (role-based MAS)     & 85.2\% & 72.8\% & 400{,}351 & 1.34M & 2.19 \\
Broadcast deliberation   & 89.2\% & 78.6\% & 616{,}499 & 1.84M & 1.35 \\
\bottomrule
\end{tabular}
}
\end{table}

\begin{figure}[t]
\centering
\includegraphics[width=\linewidth]{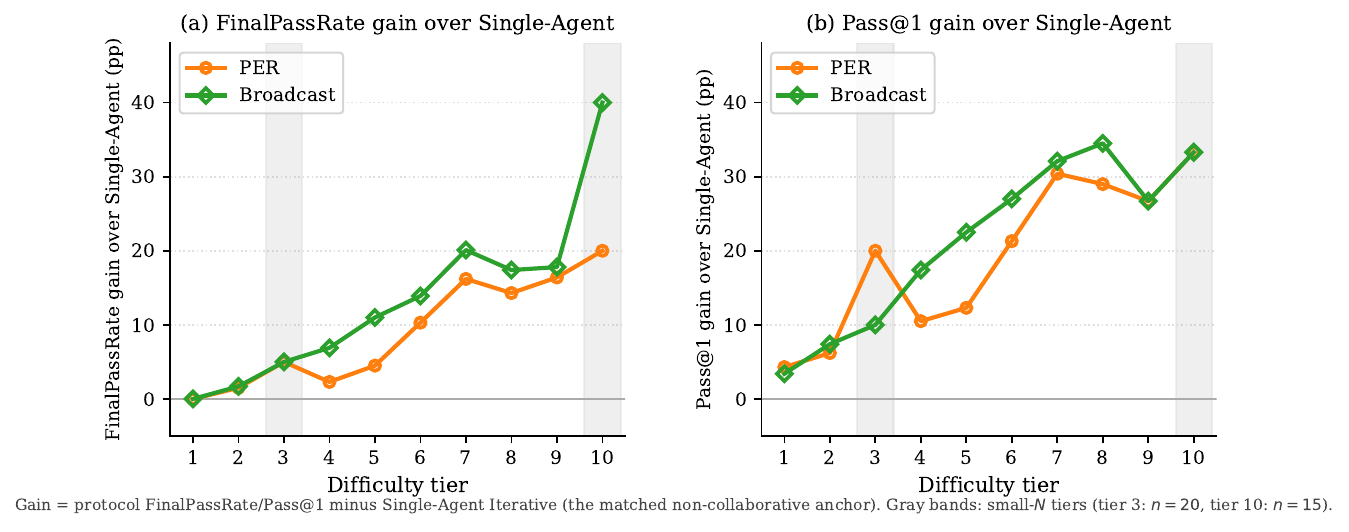}
\caption{\textbf{Tier-level collaboration gain over Single-Agent Iterative.} Single-Agent Iterative is the matched non-collaborative anchor: it already has iterative opportunity and verifier interaction, so gain over Single-Agent more directly reflects what deliberation and review contribute. \textbf{(a) FinalPassRate gain.} Gains are near zero on tiers 1--2 and open sharply from tier 4 onward. \textbf{(b) Pass@1 gain.} The same pattern holds before any verifier-guided retry, confirming that the hard-tier advantage is earned during pre-submission deliberation rather than from extra outer-loop attempts. All gains are measured in percentage points. Gray bands mark the benchmark's small-$N$ tiers (tier 3: $n=20$; tier 10: $n=15$).}
\label{fig:collaboration_gain_by_tier}
\end{figure}

\subsection{Mechanism: The Reviewer--Solver Decoupling Gap}
\label{ssec:rq3}

\paragraph{The role-specialization hypothesis under difficulty.}
The next question is which process difference best tracks the observed PER--broadcast accuracy gap in this setting. If specialized reviewers are supposed to help most on problems that genuinely require critique to be integrated, then the key issue is not just whether PER's reviewer is accurate, but whether the protocol converts that signal into solver action. We organize RQ3 around a simple chain: \textbf{detection}, \textbf{uptake}, and \textbf{repair}. The clearest separation appears in uptake.

The pattern appears even before we ask whether revisions succeed. Among problems with a wrong initial candidate, broadcast repairs 25.7\% and neglects 26.2\%, whereas PER repairs only 11.0\% and neglects 48.8\%; try-but-fail rates are much closer (48.1\% vs.\ 40.2\%). For PER, neglect is the dominant failure mode, not inability to repair after attempting revision. That pattern persists across difficulty. On the more populated hard tiers 6--9, PER reviewer precision remains high (0.81--0.87), yet even evaluator-verified useful critique changes the next candidate only 0.32--0.37 of the time, while broadcast remains around 0.93--0.96. Those same tiers are also where the matched single-agent baseline drops below 70\% final accuracy, so the growing collaboration gain does not coincide with any drop in reviewer precision. Appendix Figure~\ref{fig:coupling_mechanism_by_tier} gives the full tier view.

\begin{figure}[t]
\centering
\includegraphics[width=0.96\linewidth]{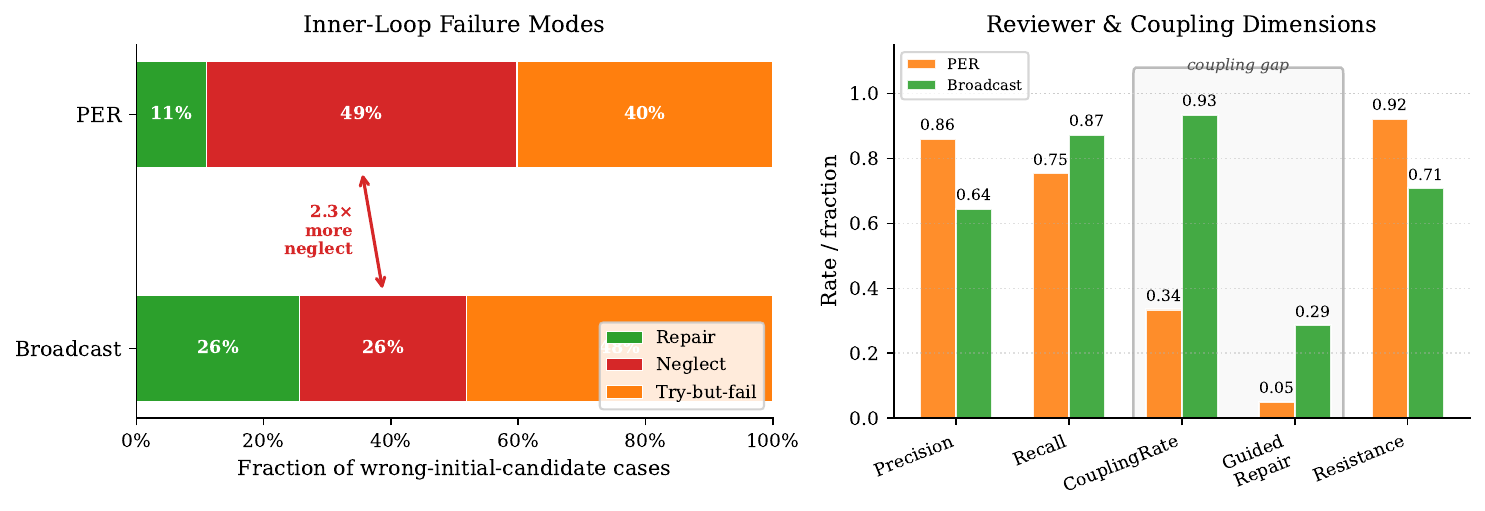}
\caption{\textbf{What observable process difference tracks the PER--broadcast gap?} Not reviewer precision alone, but whether evaluator-verified useful critique changes the next candidate the protocol carries forward. \textbf{Left:} NeglectRate is the dominant inner-loop failure mode for PER (48.8\% of wrong-initial-candidate cases neglect available critique), while TryButFailRate is much closer across protocols. \textbf{Right:} the aggregate metrics show the core paradox: PER has \emph{higher} reviewer precision (0.861 vs.\ 0.644), yet evaluator-verified useful critique changes the protocol's next candidate only 33.6\% of the time under PER versus 93.5\% under broadcast (CouplingRate), and ReviewerGuidedRepairRate is correspondingly 5.6$\times$ lower (5.1\% vs.\ 28.6\%). The shaded box highlights the coupling and repair dimensions where the gap between protocols is largest.}
\label{fig:inner_loop_coupling}
\end{figure}

\paragraph{Review-conditioned decomposition: the coupling gap.}
Figure~\ref{fig:inner_loop_coupling} makes the decomposition explicit. On \textbf{detection}, PER's reviewer has higher precision (0.861 vs.\ 0.644). On \textbf{uptake}, the contrast is sharper: under PER, evaluator-verified useful revise episodes produce an immediate answer transition only 33.6\% of the time; under broadcast, they do so 93.5\% of the time. On \textbf{repair}, the same gap reappears: ReviewerGuidedRepairRate is 0.051 for PER versus 0.286 for broadcast. In the main comparison, the higher-coupling protocol is also the higher-repair protocol, so these answer-transition events align with realized correction rather than appearing as spurious or cosmetic changes. PER is also more resistant to misleading review: it declines to revise correct answers 92\% of the time when the reviewer falsely flags them, versus 71\% for broadcast. Because CouplingRate is a protocol-level answer-transition statistic, part of this contrast reflects each protocol's own update and submission rules, not only the model's internal response to critique. The same interface that makes useful critique easier to bypass also makes false alarms easier to resist.

The same mechanism is visible in individual traces. In one tier-5 PER trace, the reviewer proposes a corrected boxed answer yet the pipeline still submits the earlier candidate unchanged. In a tier-6 broadcast trace with the same qualitative structure, peer critique replaces the live candidate before re-approval and submission. Appendix~\ref{app:trace_examples} gives the full trace excerpts.

We treat PER versus broadcast as evidence about complete protocol designs, not routing alone; the within-PER probes below test whether more direct critique presentation increases uptake.

\begin{figure}[t]
\centering
\includegraphics[width=0.96\linewidth]{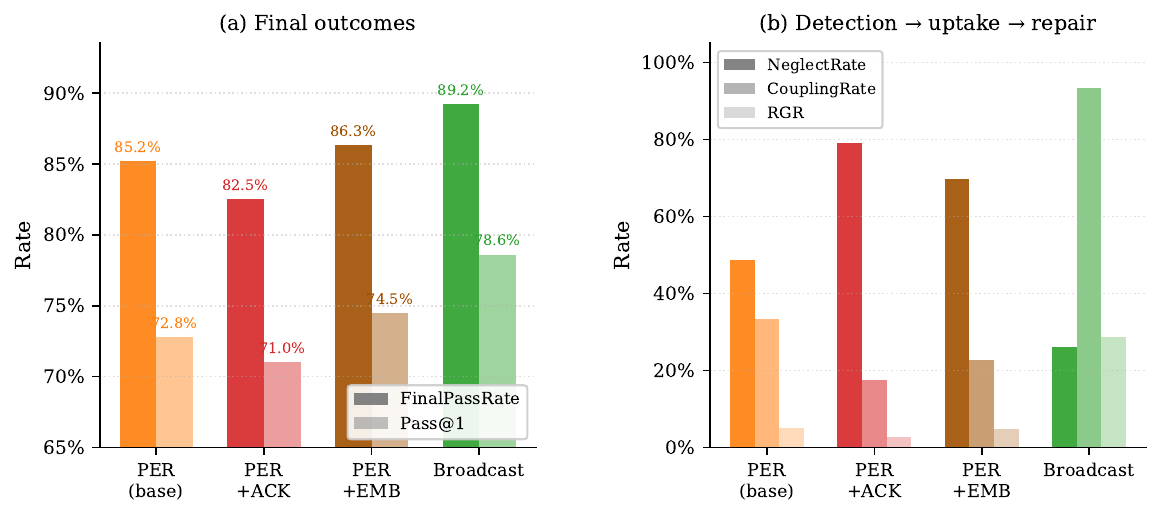}
\caption{\textbf{Within-PER intervention evidence for the uptake story.} All values are full-benchmark aggregates ($N=4{,}181$); the model family, reviewer setup, and outer-loop budget are held fixed across conditions, varying only how critique reaches the solver. \textbf{(a) Final outcomes.} ACK-required lowers FinalPassRate (85.2\%$\to$82.5\%) and Pass@1, while EMB partially recovers both. \textbf{(b) Detection$\to$uptake$\to$repair.} ACK-required sharply worsens NeglectRate (0.488$\to$0.792) and CouplingRate (0.336$\to$0.175), confirming it as a failed compliance-style control. EMB reduces neglect and partially restores coupling (0.698 NeglectRate, 0.227 CouplingRate), consistent with presenting reviewer feedback more directly improving follow-through within the PER family, while still falling short of broadcast's shared-state design. Read as directional evidence for the uptake story, not as a single clean ablation.}
\label{fig:coupling_intervention}
\end{figure}

\paragraph{Within-PER coupling interventions.}
The uptake gap also shifts within PER itself. These within-PER probes are the source of the paper's directional evidence that more direct critique delivery improves uptake. We hold the model family, reviewer setup, and outer-loop budget fixed and vary only how critique reaches the solver. Figure~\ref{fig:coupling_intervention} shows the direction of movement. ACK-required adds mandatory verbal acknowledgment but keeps routed advice unchanged; it does not rescue PER: FinalPassRate falls from 85.2\% to 82.5\%, NeglectRate rises from 0.488 to 0.792, and CouplingRate drops from 0.336 to 0.175. In this prompt configuration, the solver can satisfy the new instruction without changing the candidate itself. This pattern is consistent with superficial compliance: verbal acknowledgment can separate from an actual update to the underlying candidate.

EMB places reviewer feedback directly in the solver's working context. Relative to the matched ACK-required control, FinalPassRate rises from 82.5\% to 86.3\%, Pass@1 from 71.0\% to 74.5\%, and NeglectRate falls from 0.792 to 0.698. Compared with exact base PER, the gain is more selective: end-to-end outcomes improve, but direct uptake diagnostics do not uniformly dominate base PER. We read \emph{Base PER $\rightarrow$ EMB $\rightarrow$ Broadcast} as a sequence of designs with progressively more direct reviewer feedback rather than as a single clean ablation: ACK-required does not fix the uptake gap, while more direct presentation of reviewer feedback is associated with better follow-through.

\paragraph{A deeper-reflection PER stress test.}
We also test whether PER is simply under-iterated locally. A separate full-benchmark PER variant with a deeper local reflective budget (PER-inner6) improves over base PER on Pass@1, FinalPassRate, and HintAwareCorrectionScore, but it still trails broadcast and still uses more evaluator calls per problem than broadcast (2.07 vs.\ 1.35). We read deeper local reflection as a partial recovery of PER rather than as a full explanation of the broadcast advantage. Details appear in Appendix Tables~\ref{tab:completed_coupling_intervention_appendix} and~\ref{tab:coupling_intervention_tier_finalpass}, Figure~\ref{fig:supp_per_inner6}, and Appendix~\ref{app:per_inner6}.

%% file: sections/06_conclusion.tex

\section{Conclusion}
\label{sec:conclusion}

Specialized reviewers help only when correct critique changes the next candidate the protocol carries forward. In our study, PER's reviewer has higher precision, yet the next answer changes far less often after evaluator-verified useful critique, and realized repair is correspondingly lower. NeglectRate, CouplingRate, and ReviewerGuidedRepairRate separate detection, uptake, and repair so that each step is independently measurable. The main finding is more limited than a protocol ranking: in this hard benchmark and primary shared-model setting, higher reviewer precision does not by itself guarantee better collaborative reasoning, and reviewer precision and critique uptake can come apart.

This result matters both for evaluation design and for system design. A system that detects errors but fails to act on the corresponding critique can appear strong under reviewer-centric evaluation while failing to improve actual problem solving. Improving reviewer quality alone may therefore not improve outcomes unless the interface helps correct feedback change the next candidate.

The within-PER intervention suite then adds directional support for that account. In the tested prompt configuration, ACK-required is a failed compliance-style control, EMB is a partial recovery step, and deeper local reflection helps only partially while still using more evaluator calls than broadcast. Taken together, these results are consistent with critique presented more directly in the solver's working context playing an important role inside hierarchical pipelines, even though they do not isolate a single causal factor or imply that uptake alone determines the final ranking. The current evidence also does not yet distinguish cleanly between two nearby failure modes inside PER: critique that is not fully incorporated at all, and critique that is seen but treated as less authoritative than the solver's existing reasoning. A reduced cross-family replication with Gemma~3~27B actors preserves the same precision--uptake separation even as the overall ranking shifts. In that replication, PER's reviewer still has higher precision (PER: 0.881, Broadcast: 0.722), while Broadcast shows much higher useful-critique uptake (Broadcast: 0.742, PER: 0.092) and a correspondingly larger repair rate. This suggests that uptake is an important hidden evaluation axis, but not by itself a universal predictor of protocol quality across model families; see Appendix~\ref{app:gemma3_replication}. Within hierarchical agent pipelines, reviewer quality alone is not a sufficient design principle; performance also depends on whether the protocol turns correct critique into the next candidate it carries forward.

%% file: sections/07_limitations_and_broader_impact.tex

\section{Limitations and Broader Impact}
\label{sec:limitations}

This paper studies protocol structure in a deliberately restricted but rigorous setting: verifier-grounded olympiad-level math from the Omni-MATH benchmark family \cite{omnimath,omnimath2}. Our conclusions are therefore about protocol behavior under this benchmark, not about universal superiority of one interaction style across every domain. The main benchmark evidence comes from one primary actor family, while Appendix~\ref{app:gemma3_replication} adds a reduced cross-family replication with Google's open-weight Gemma~3~27B actors from a different family while holding the evaluator fixed. Even with that addition, protocol rankings may still vary with model family or deployment details such as prompt templates and context budgets. Our claim is thus about a common PER-style feedback interface under the matched configurations used in this study, not about the impossibility of hierarchical specialization itself. We expect the detection--uptake--repair decomposition to generalize more readily than the exact rate values, which may differ in domains where correctness is less binary than math benchmarking. CouplingRate is an operational answer-transition statistic on evaluator-verified useful revise episodes, not a free-form semantic judgment about whether the solver ``understood'' the critique. Our current within-PER probes also do not isolate whether lower uptake comes primarily from information formatting, role-hierarchy weighting, or both. Appendix~\ref{app:metrics} and Appendix~\ref{app:evaluator} provide the stricter answer-transition audit and cross-evaluator replay.

The broader impact is two-sided. On the positive side, coupling-aware evaluation can reduce overclaiming by showing when valid critique fails to become action, which matters for benchmark design and settings where corrective feedback is costly or safety-critical \cite{scalableoversight}. In safety-critical settings, correct critique that is ignored can be functionally equivalent to no critique at all. On the other hand, stronger feedback uptake can also increase effective capability. We therefore view clearer mechanistic evaluation as a tool for both capability assessment and more trustworthy oversight.

\paragraph{Data and code.}
The public release website is available at \url{https://huggingface.co/spaces/AgentsSci/scientific-agent-protocol-traces-site}. Code and data will be released after institutional approval.

%% file: sections/08_appendix.tex

\section*{Appendix}

This supplementary material is organized in two layers. The first appendix pages defend the paper's main claim with the highest-priority robustness checks; the later sections provide the fuller metric catalog, protocol configurations, extended results tables, trace examples, and process diagnostics referenced from the main paper.

\section{Robustness Checks for the Main Claim}
\label{app:robustness_summary}

Because some reviewers may skim only the first appendix pages, we front-load the checks that most directly protect the paper's core claim. Table~\ref{tab:robustness_summary_appendix} summarizes the four main concerns and the appendix evidence that addresses them. The rest of the appendix then serves the second job: reference material, fuller metric definitions, protocol documentation, and extended plots.

\begin{table}[h]
\centering
\small
\caption{High-priority robustness checks for the paper's main claim. These are the appendix items most directly aimed at protecting the precision--uptake story rather than supplying background reference material.}
\label{tab:robustness_summary_appendix}
\resizebox{\linewidth}{!}{
\begin{tabular}{p{3.1cm}p{4.9cm}p{5.6cm}p{2.4cm}}
\toprule
\textbf{Concern} & \textbf{Check} & \textbf{Main result} & \textbf{Where} \\
\midrule
Coupling metric is brittle & Deterministic and equivalence-aware answer-transition audits on the same useful-revise episodes & Strict and legacy CouplingRate are numerically identical for both PER and Broadcast, with zero strict-vs-legacy disagreement & Table~\ref{tab:strict_coupling_audit} \\
\addlinespace
Single evaluator artifact & Cross-evaluator replay with \texttt{gpt-oss-120b}, Llama, and Gemma under the same prompt/parser & Full-benchmark disagreement stays at 3.38\%--5.76\% with $\kappa=0.850$--$0.915$; disagreement rises on the hard collaborative slice but does not reverse the main pattern & Appendix~\ref{app:evaluator} \\
\addlinespace
PER just needs more local reflection & Higher-cost PER-inner6 stress test & PER-inner6 improves over base PER (FinalPassRate 85.2\%$\to$86.3\%) but still trails Broadcast (89.2\%) and still uses more evaluator calls (2.07 vs.\ 1.35) & Appendix~\ref{app:per_inner6} \\
\addlinespace
Single actor family only & Reduced cross-family replication with Google Gemma~3~27B actors and fixed \texttt{gpt-oss-120b} evaluator & The overall ranking shifts, but the precision--uptake separation remains strong (PER precision 0.881 vs.\ 0.722; uptake 0.092 vs.\ 0.742) & Appendix~\ref{app:gemma3_replication} \\
\bottomrule
\end{tabular}
}
\end{table}

\subsection{Coupling Metric Robustness}

The paper-facing CouplingRate is an operational immediate answer-transition statistic on evaluator-verified useful revise episodes. The main robustness question is whether the PER--broadcast uptake gap depends on a fragile answer-normalization choice. To check that, we re-audit the same episodes with two stricter definitions: a deterministic normalized answer key and an equivalence-aware exact answer checker. The practical result is stable: for both PER and Broadcast, the strict deterministic audit matches the legacy paper-facing CouplingRate exactly, and the equivalence-aware variant changes only a few PER cases without altering the qualitative gap.

\begin{table}[h]
\centering
\footnotesize
\setlength{\tabcolsep}{3pt}
\renewcommand{\arraystretch}{1.08}
\caption{Pooled strict answer-transition audit for the main PER and broadcast comparison. Legacy CouplingRate is the paper-facing uptake metric used in the main text; Strict CouplingRate re-audits the same useful-revise episodes with a deterministic normalized answer-key rule; EquivalenceAware CouplingRate replaces that rule with the exact deterministic answer-equivalence checker. The practical point is that the uptake gap is definition-stable: for both protocols, strict and legacy rates are numerically identical and strict-vs-legacy disagreement is zero on all eligible useful-revise episodes. This shows that the PER--broadcast coupling gap does not depend on a fragile normalization choice or on one particular legacy pipeline label.}
\label{tab:strict_coupling_audit}
\begin{tabular}{@{}>{\raggedright\arraybackslash}p{1.2cm}>{\centering\arraybackslash}p{1.55cm}>{\centering\arraybackslash}p{2.35cm}>{\centering\arraybackslash}p{2.35cm}>{\centering\arraybackslash}p{2.85cm}>{\centering\arraybackslash}p{1.8cm}@{}}
\toprule
\textbf{Protocol} & \textbf{\shortstack[c]{Useful revise\\episodes}} & \textbf{\shortstack[c]{Legacy\\CouplingRate}} & \textbf{\shortstack[c]{Strict\\CouplingRate}} & \textbf{\shortstack[c]{EquivalenceAware\\CouplingRate}} & \textbf{\shortstack[c]{Strict-vs-legacy\\disagreement}} \\
\midrule
PER & 5{,}253 & 0.336 (1{,}766/5{,}253) & 0.336 (1{,}766/5{,}253) & 0.340 (1{,}787/5{,}253) & 0 / 5{,}253 \\
Broadcast & 5{,}214 & 0.935 (4{,}876/5{,}214) & 0.935 (4{,}876/5{,}214) & 0.935 (4{,}876/5{,}214) & 0 / 5{,}214 \\
\bottomrule
\end{tabular}
\end{table}

\section{Evaluator Selection and Leakage Controls}
\label{app:evaluator}

\begin{itemize}
\item Evaluator correctness matters in verifier-gated math benchmarks because judge errors can distort model and protocol comparisons as systems improve \cite{omnimath2,strongverifiers,xverify,jetts}.
\item \texttt{gpt-oss-120b} is a public OpenAI open-weight reasoning model released under Apache 2.0, with public downloads and deployment documentation \cite{openai_gptoss_release,openai_gptoss_model}.
\item In smoke-test disagreements between \texttt{gpt-oss-120b} and a smaller evaluator, manual review by a PhD-level annotator with math background favored \texttt{gpt-oss-120b}. We therefore use \texttt{gpt-oss-120b} as the evaluator.
\item To mitigate same-model-family coupling, correctness judgment is restricted to the submitted boxed answer, the ground-truth answer, and minimal reference explanation when needed for answer equivalence. The evaluator does not read agents' reasoning traces when deciding pass/fail.
\item Only after a fixed \texttt{FAIL} verdict, and only for hinted feedback, may a hint-generation pass inspect compact reasoning context to produce a short non-answer-revealing repair hint. A hint scanner is applied before any hint is forwarded to acting agents.
\item The same evaluator and judging mechanism are used across all protocols for fairness.
\end{itemize}

As a robustness check against single-evaluator sensitivity, we replay the exact saved submitted answers under the same evaluator prompt/parser with three open models: \texttt{gpt-oss-120b}, \texttt{Meta-Llama-3.1-70B-Instruct}, and \texttt{gemma-3-27b-it}. The first replay covers the full paper experiment output (16{,}724 submission instances; 10{,}590 unique candidate answers after deduplication). Table~\ref{tab:evaluator_disagreement_appendix} shows that overall pairwise agreement remains high: instance-level disagreement stays between 3.38\% and 5.76\%, with Cohen's $\kappa$ between 0.850 and 0.915. Cohen's $\kappa$ is a chance-corrected inter-rater agreement coefficient \cite{cohen1960agreement}; by common heuristics, the observed range is generally read as strong agreement \cite{landis1977observer,mchugh2012kappa}. We therefore treat the replay as trustworthy as a robustness check rather than as a second ground-truth source, and we report raw disagreement alongside $\kappa$ because agreement coefficients can still be affected by label prevalence. The disagreements are not uniform, however. They rise on the collaborative modes and on the hard tiers where answer equivalence and partially correct boundary cases are more common. For example, \texttt{gpt-oss-120b} versus Llama disagrees on 4.38\% of baseline outputs but on 6.21\% of PER outputs and 6.57\% of broadcast outputs. On the hard tiers 7--10, the corresponding per-mode disagreement reaches 9.95\% for PER and 9.47\% for broadcast in the full replay.

Because those hard collaborative cases are exactly where evaluator sensitivity matters most, we also run a second replay restricted to PER and broadcast on tiers 7--10 (1{,}694 submission instances). This focused replay is not a replacement benchmark-wide result; it is a stress test that removes the remaining missing judgments and gives all three evaluators 100\% valid coverage on the same hard slice. Under that restriction, overall pairwise disagreement rises to 9.68\% for \texttt{gpt-oss-120b} versus Llama, 9.39\% for \texttt{gpt-oss-120b} versus Gemma, and 5.25\% for Llama versus Gemma, with $\kappa=0.755$, $0.761$, and $0.875$, respectively. The largest tier/mode cell is tier-9 PER, where \texttt{gpt-oss-120b} versus Llama reaches 12.33\% disagreement on $n=146$ instances; tier 8 is also near 9--10\% for both PER and broadcast. Tier 10 has only 30 total focused-slice instances and should not be overinterpreted. Across the focused replay, 206 of 1{,}694 instances are contested by at least one evaluator pair, split almost evenly between PER (104) and broadcast (102). We therefore use \texttt{gpt-oss-120b} as the main evaluator and treat the replay suite as a robustness check rather than a replacement judge: disagreement is real and concentrates where the benchmark is hardest, but the overall agreement levels are high enough that the main protocol story is unlikely to be a trivial artifact of one evaluator.

\begin{table}[h]
\centering
\small
\caption{Cross-evaluator replay over saved paper-experiment submissions. The first block is the full benchmark; the second is the focused hard collaborative slice (PER/Broadcast, tiers 7--10). Coverage is the fraction of submission instances for which both evaluators returned a valid binary judgment. Disagreement is computed on submission instances rather than deduplicated candidates; the unique-candidate replay is slightly harsher but does not change the qualitative picture.}
\label{tab:evaluator_disagreement_appendix}
\begin{tabular}{@{}llccc@{}}
\toprule
\textbf{Slice} & \textbf{Evaluator pair} & \textbf{Coverage} & \textbf{Disagreement} & \textbf{$\kappa$} \\
\midrule
\textbf{Full benchmark} & \texttt{gpt-oss-120b} vs.\ Llama & 99.83\% & 5.76\% & 0.850 \\
& \texttt{gpt-oss-120b} vs.\ Gemma & 99.45\% & 4.35\% & 0.884 \\
& Llama vs.\ Gemma & 99.53\% & 3.38\% & 0.915 \\
\addlinespace
\textbf{Focused hard slice} & \texttt{gpt-oss-120b} vs.\ Llama & 100.0\% & 9.68\% & 0.755 \\
& \texttt{gpt-oss-120b} vs.\ Gemma & 100.0\% & 9.39\% & 0.761 \\
& Llama vs.\ Gemma & 100.0\% & 5.25\% & 0.875 \\
\bottomrule
\end{tabular}
\end{table}

\begin{figure}[h]
\centering
\includegraphics[width=\linewidth]{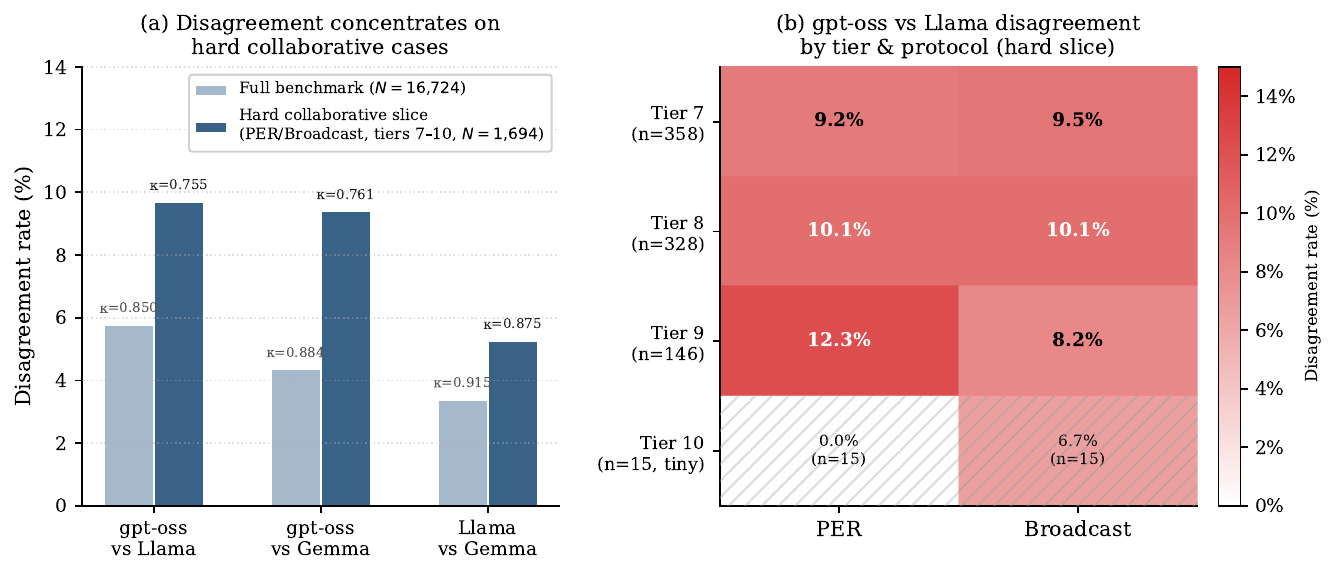}
\caption{\textbf{Cross-evaluator disagreement robustness.} \textbf{(a)} Grouped bars compare pairwise disagreement rates on the full benchmark ($N=16{,}724$; light bars) versus the focused hard collaborative slice (PER and Broadcast, tiers 7--10, $N=1{,}694$; dark bars). Cohen's $\kappa$ is annotated above each bar. Overall agreement is high (disagreement 3.4--5.8\%, $\kappa=0.850$--$0.915$), but disagreement roughly doubles on the hard collaborative slice (5.3--9.7\%, $\kappa=0.755$--$0.875$). \textbf{(b)} \texttt{gpt-oss-120b} vs.\ Llama disagreement heatmap for the hard collaborative slice, stratified by tier and protocol. Disagreement concentrates at tiers 8--9 ($\sim$9--12\%); tier 10 has only 15 instances per protocol and its lower rate should not be overinterpreted (hatched). The pattern confirms that the main protocol story is not a trivial single-evaluator artifact, but evaluator sensitivity is real exactly where answer equivalence is most ambiguous.}
\label{fig:supp_eval_disagreement}
\end{figure}

\section{Deeper-Reflection PER Stress Test}
\label{app:per_inner6}

We also test whether PER's local reflective loop is simply too shallow. To stress-test that hypothesis, we ran a separate full-benchmark PER variant with a deeper local reflection budget (PER-inner6). This probe is intentionally distinct from ACK-required and EMB: it changes inner-loop depth rather than the critique surface. We therefore do not treat it as a missing cell in the within-PER coupling-intervention continuum.

The resulting pattern is informative rather than destabilizing. PER-inner6 improves over base PER on first-pass and final outcomes, suggesting that deeper local reflection can partially recover PER. But it still does not match broadcast, and it does not remove the practical verifier-cost asymmetry. In particular, PER-inner6 remains cheaper than broadcast in generation tokens yet still uses more evaluator calls per problem. The small drop in AvgEvalCalls/problem from base PER (2.19) to PER-inner6 (2.07) should therefore be read as a second-order consequence of better early repair: more problems are solved before an external verifier failure can trigger another outer retry, even though the protocol spends substantially more inner-loop reasoning on each surviving hard case. The cleanest reading is therefore that deeper local reflection helps, but does not make the feedback-uptake interface issue disappear.

\begin{table}[h]
\centering
\small
\setlength{\tabcolsep}{3pt}
\renewcommand{\arraystretch}{1.08}
\caption{Deeper-reflection PER stress test over the completed full benchmark ($N=4{,}181$). PER-inner6 increases the local reflective budget but does not change the acting/evaluator model family or benchmark slice. This is a stress test of the shallow-local-reflection hypothesis, not a new main protocol family.}
\label{tab:per_inner6}
\begin{tabular}{@{}lccccccc@{}}
\toprule
\textbf{Protocol} & \textbf{\shortstack[c]{Final\\PassRate}} & \textbf{Pass@1} & \textbf{\shortstack[c]{HintAware\\CorrectionScore}} & \textbf{AvgTokens} & \textbf{\shortstack[c]{AvgEvalCalls/\\problem}} & \textbf{\shortstack[c]{AvgReflective\\Rounds}} & \textbf{\shortstack[c]{Misleading\\ReviewHarm}} \\
\midrule
PER                        & 85.2\% & 72.8\% & 70.4\% & 400K & 2.19 & 1.17 & 0.083 \\
PER-inner6                 & 86.3\% & 74.8\% & 73.9\% & 557K & 2.07 & 2.06 & 0.183 \\
Broadcast                  & 89.2\% & 78.6\% & 81.8\% & 616K & 1.35 & 4.23 & 0.313 \\
\bottomrule
\end{tabular}
\end{table}

\begin{figure}[h]
\centering
\includegraphics[width=\linewidth]{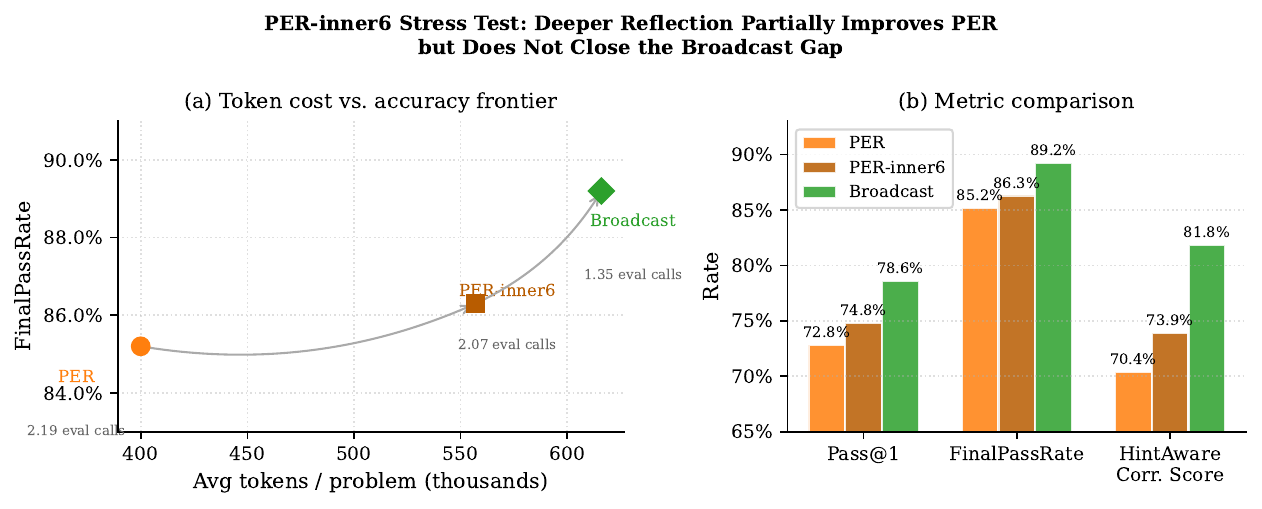}
\caption{Deeper-reflection PER stress test: PER-inner6 (brown-orange) increases local reflective rounds from 1.17 to 2.06 per problem. Panel~(a) shows the token-cost vs.\ accuracy frontier; annotations give AvgEvalCalls/problem beside each point. Panel~(b) compares Pass@1, FinalPassRate, and HintAwareCorrectionScore across the three protocols. PER-inner6 partially closes the accuracy gap to Broadcast (FinalPassRate: 86.3\% vs.\ 85.2\% base PER, 89.2\% Broadcast) while still using more evaluator calls per problem (2.07) than Broadcast (1.35). The gap is narrowed but not closed; deeper local reflection helps, while the verifier-cost asymmetry persists.}
\label{fig:supp_per_inner6}
\end{figure}

\section{Cross-Family Partial Replication with Gemma 3 Actors}
\label{app:gemma3_replication}

To probe whether the main precision--uptake separation depends entirely on the primary actor family, we ran a reduced four-mode replication with Google's open-weight \texttt{gemma-3-27b-it} actors, a different 27B actor family from the primary \texttt{gpt-oss-120b} run, while keeping the evaluator fixed at \texttt{gpt-oss-120b}. This addendum uses the tier-sampled subset currently available from that sweep ($N=835$ problems across the same ten benchmark tiers, with tiers 3 and 10 still small-$N$). The subset contains a random sample of 100 problems per tier when available; for tiers with fewer than 100 problems, we include the full tier. Table~\ref{tab:gemma3_replication} reports the stable cross-family signal we rely on in the paper: the raw protocol ranking shifts, but reviewer precision and critique uptake remain sharply separated. To keep this appendix addition conservative, we summarize the pooled RQ1 outcomes together with the review-conditioned RQ3 metrics; we do not use the current reduced-run RQ2 export in the paper text until its attempt-depth columns are fully reconciled with the launch budget.

\begin{table}[h]
\centering
\small
\setlength{\tabcolsep}{3pt}
\renewcommand{\arraystretch}{1.08}
\caption{Reduced cross-family replication with Gemma~3 actors and a fixed \texttt{gpt-oss-120b} evaluator ($N=835$ tier-sampled problems). FinalPassRate and Pass@1 are computed over all sampled problems. Avg.\ Model Calls is the average number of model calls per problem. Reviewer precision asks how often a reviewer warning truly points to a real error. Useful-critique uptake is the paper-facing CouplingRate on evaluator-verified useful-critique episodes. Reviewer-guided repair is the fraction of those useful-review episodes that the next candidate actually fixes. The ranking differs from the primary \texttt{gpt-oss-120b} actor run, but the precision--uptake separation remains strong.}
\label{tab:gemma3_replication}
\begin{tabular}{@{}lcccccc@{}}
\toprule
\textbf{Protocol} & \textbf{\shortstack[c]{Final\\PassRate}} & \textbf{Pass@1} & \textbf{\shortstack[c]{Avg.\ Model\\Calls}} & \textbf{\shortstack[c]{Reviewer\\precision}} & \textbf{\shortstack[c]{Useful-critique\\uptake}} & \textbf{\shortstack[c]{Reviewer-guided\\repair}} \\
\midrule
Baseline LLM (one-shot) & 42.6\% & 42.6\% & 2.62 & --- & --- & --- \\
Single-Agent Iterative & 60.5\% & 42.5\% & 6.92 & --- & --- & --- \\
PER (role-based MAS) & 65.6\% & 43.4\% & 22.83 & 0.881 & 0.092 & 0.009 \\
Broadcast deliberation & 58.7\% & 38.3\% & 130.11 & 0.722 & 0.742 & 0.174 \\
\bottomrule
\end{tabular}
\end{table}

The cross-family takeaway is narrower than a replication of the primary ranking. On this smaller-model subset, PER has the higher final accuracy, yet Broadcast remains much more likely to turn correct critique into the next candidate the protocol carries forward and to realize repair from that critique. We therefore interpret the Gemma addendum as evidence that reviewer precision and critique uptake remain separable even when protocol ranking shifts across actor families, not as a second protocol leaderboard.

\section{Notation}
\label{app:notation}

\begin{itemize}
\item Let $N$ denote the number of evaluated problems.
\item Let $y_i^{\mathrm{final}} \in \{0,1\}$: eventually solved within budget.
\item Let $y_i^{(1)} \in \{0,1\}$: first outer-loop submission passes.
\item Let $a_i$: index of first successful outer-loop submission.
\item Let $\mathcal{T}$ denote the set of difficulty tiers, and let $D_t=\{i:d(i)=t\}$ be the subset of problems in tier $t$.
\item Let $f_i = a_i - 1$: number of failed outer-loop submissions before first success.
\item Let $h_i$: number of hint-bearing feedback rounds before first success.
\item Let $R_i$: number of reflective-refinement rounds on problem $i$.
\item Let $z_i^{(0)}, z_i^{(1)} \in \{0,1\}$: correctness of initial and final internal candidates in the inner loop.
\item Let $x_i^{(0)}, x_i^{(1)}$: extracted initial and final candidate answers (for NeglectRate splitting).
\item Let $r_i \in \{\mathrm{Revise}, \mathrm{Agree}\}$: extracted review action on review-labeled subset.
\item Let $g_i \in \{0,1\}$: whether evaluator feedback was incorporated in the next revision.
\item Let $t_i$: total token cost for problem $i$. Let $c_i$: model/API calls. Let $e_i$: evaluator calls. Let $w_i$: wall-clock time.
\item Let $d(i)$: difficulty tier. Let $s(i)$: mathematical sub-domain.
\end{itemize}

\section{Extended Term Lineage}
\label{app:metric_lineage}

This section expands the brief paper-facing terminology into its longer literature lineage and the specific measurement gaps each term is meant to expose.

\begin{table}[h]
\centering
\footnotesize
\setlength{\tabcolsep}{3pt}
\renewcommand{\arraystretch}{1.1}
\caption{Extended term lineage table. This version keeps the literature anchors and the specific measurement gaps; the main paper uses only the brief terminology needed to follow the mechanism argument.}
\begin{tabular}{@{}>{\raggedright\arraybackslash}p{2.5cm}>{\raggedright\arraybackslash}p{3.2cm}>{\raggedright\arraybackslash}p{3.2cm}>{\raggedright\arraybackslash}p{3.1cm}@{}}
\toprule
\textbf{Term} & \textbf{Definition} & \textbf{Literature anchor} & \textbf{What prior work misses} \\
\midrule
Within-attempt review (inner loop) & Reviewer critiques solver draft; solver may revise---all before any external verifier verdict is issued & Self-Refine \cite{selfrefine}; intrinsic self-correction \cite{correctbench} & Multi-agent extension: reviewer and solver are separate agents \\
\addlinespace
Verifier-guided repair (outer loop) & System retries after the external verifier issues a FAIL signal & Extrinsic correction \cite{correctbench}; iterative self-repair \cite{iterativeselfrepair} & Protocol-level comparison of who benefits from verifier feedback \\
\addlinespace
ReflectiveRepairRate & Among wrong initial candidates, fraction corrected by end of the inner loop \emph{(extended)} & Protocol-level transition \cite{selfrefine,selfcorrection} & Not separated from reviewer quality \\
\addlinespace
NeglectRate \emph{(new)} & Among reviewer-flagged wrong answers, fraction where solver makes no revision (submits same answer) & ``Self-correction blind spot'' \cite{selfcorrectionblindspot} (single-agent analog) & Never measured in multi-agent settings \\
\addlinespace
TryButFailRate \emph{(new)} & Among reviewer-flagged wrong answers, fraction where solver revises but answer remains wrong & Error depth hypothesis \cite{decomposingselfcorrection} (single-agent) & Not separated from neglect---opposite implications for system design \\
\addlinespace
Reviewer--Solver Coupling \emph{(new)} & Whether a reviewer's correct flag reliably produces solver revision (high coupling) or not (decoupling) & No prior work measures this at the agent-interface level & Root cause of PER's repair gap \\
\addlinespace
ReviewerGuided\allowbreak RepairRate \emph{(new)} & Among problems where reviewer correctly identifies a wrong answer (TP), fraction that the solver actually fixes & Decomposes reviewer precision from solver compliance & Prior work treats reviewer quality and system repair as one \\
\addlinespace
MisleadingResistance \emph{(new)} & Among problems where reviewer incorrectly flags a correct answer (FP), fraction the solver resists revising & Review false-positive handling in prior work is usually implicit & Trade-off between safety against false alarms and responsiveness to valid critique \\
\addlinespace
ConditionalRecoveryRate@$k$ \emph{(new)} & Count of problems first solved at attempt $k$ divided by the count that reached attempt $k$ & Hazard-rate framing of outer-loop recovery; survival analysis analog & Marginal gain masks diminishing returns; hazard rate models verifier-call value directly \\
\bottomrule
\end{tabular}
\end{table}

\section{Literature Map for the Evaluation Lens}
\label{app:literature_map}

This section situates the paper's evaluation lens relative to the adjacent multi-agent, self-correction, and benchmark-design strands it draws on.

\begin{table}[h]
\centering
\small
\caption{Compact literature map for the evaluation lens of this paper. The goal is not exhaustive coverage, but to show the main strands our framing draws on and the specific gap we target.}
\resizebox{\linewidth}{!}{
\begin{tabular}{p{3.7cm}p{5.3cm}p{4.1cm}p{3.9cm}}
\toprule
\textbf{Strand} & \textbf{Representative papers} & \textbf{Typical emphasis} & \textbf{Gap we target} \\
\midrule
Peer deliberation and debate & Multi-Agent Debate \cite{multiagentdebate}, AutoGen \cite{autogen}, CAMEL \cite{camel}, Can Agents Debate? \cite{canagentsdebate} & Whether interaction among peers improves final task performance or factuality & Usually reports outcome gains, not whether critique is actually acted on \\
\addlinespace
Hierarchical and role-specialized MAS & MALT \cite{malt}, DPSDP \cite{dpsdp}, scientific-agent surveys \cite{scientificagentssurvey}, Scaling Agent Systems \cite{scalingagentsystems} & Specialized roles, reviewer/critic decomposition, and coordination gains & Usually treats reviewer quality and realized repair as one combined effect \\
\addlinespace
Self-correction and reflection & Self-Refine \cite{selfrefine}, Reflexion \cite{reflexion}, CorrectBench \cite{correctbench}, Cannot Self-Correct \cite{cannotselfcorrect}, Decomposing Self-Correction \cite{decomposingselfcorrection} & Intrinsic vs.\ extrinsic correction, error detection, revision, and repair & Mostly single-agent; does not isolate reviewer--solver coupling across distinct agents \\
\addlinespace
Agent evaluation and cost-aware benchmarking & AgentBench \cite{agentbench}, Reasoning + Action evaluation \cite{evalreasonaction}, AI Agents That Matter \cite{aiagentsmatter}, rigorous benchmark design \cite{zhu2025abc,yehudai2025agentsurvey} & Outcome validity, cost accounting, and benchmark discipline & Lacks this paper's inner/outer-loop decomposition on a verifier-gated olympiad benchmark \\
\bottomrule
\end{tabular}
}
\end{table}

\section{Full Metric Catalog}
\label{app:metrics}

\subsection{Outcome and Efficiency Metrics}

\paragraph{Final success and first-pass success.}
\begin{equation}
\mathrm{FinalPassRate} = \frac{1}{N}\sum_{i=1}^{N} y_i^{\mathrm{final}},
\qquad
\mathrm{Pass@1} = \frac{1}{N}\sum_{i=1}^{N} y_i^{(1)}.
\end{equation}

\paragraph{Average resource metrics and paper display labels.}
The runtime metric code emits \texttt{AvgTotalTokens}, \texttt{AvgModelCalls}, and \texttt{TotalEvaluatorAttempts}. In the paper tables we shorten these to AvgTokens, AvgModelCalls, and AvgEvalCalls/problem. AvgWallClock is an analysis-side summary built from recorded wall times rather than a core runtime metric.
\begin{align}
\mathrm{AvgTokens}
\equiv
\mathrm{AvgTotalTokens}
&= \frac{1}{N}\sum_{i=1}^{N} t_i,\\
\mathrm{AvgModelCalls}
&= \frac{1}{N}\sum_{i=1}^{N} c_i,\\
\mathrm{AvgEvalCalls/problem}
&= 
\frac{1}{N}\sum_{i=1}^{N} e_i
=
\frac{\mathrm{TotalEvaluatorAttempts}}{N},
\\
\mathrm{AvgWallClock}
&= \frac{1}{N}\sum_{i=1}^{N} w_i.
\end{align}

\paragraph{Tokens per extra solved problem (relative to a reference protocol).}
For a protocol $p$ relative to a reference protocol $b$ (Baseline LLM in the main paper), evaluated on the same matched problem set, we report the report-side derived quantity
\begin{equation}
\mathrm{Tokens/ExtraSolved}(p \mid b)
=
\frac{\sum_i \left(t_i^{(p)}-t_i^{(b)}\right)}
{\sum_i \left(y_i^{\mathrm{final},(p)}-y_i^{\mathrm{final},(b)}\right)}.
\end{equation}
On a matched set this is equivalent to the difference in average token cost divided by the difference in final pass rate. It answers: how many additional generation tokens are spent, on average, for each extra solved problem relative to the reference.

\subsection{Outer-Loop Metrics}

\paragraph{Cumulative and conditional recovery.}
\begin{equation}
\mathrm{OuterPass@}k = \frac{1}{N}\sum_{i=1}^{N}\mathbf{1}[y_i^{\mathrm{final}} = 1 \wedge a_i \le k],
\end{equation}
\begin{equation}
\Delta@k
\equiv
\mathrm{MarginalRecoveryGain}_{k}
=
\mathrm{OuterPass@}k - \mathrm{OuterPass@}(k-1),
\end{equation}
\begin{equation}
\mathrm{CRR@}k = \frac{\mathrm{OuterPass@}k - \mathrm{OuterPass@}(k-1)}{1 - \mathrm{OuterPass@}(k-1)}.
\end{equation}

\paragraph{Attempt-sensitive accuracy.}
\begin{equation}
\mathrm{MRR}_{\mathrm{attempt}} = \frac{1}{N}\sum_{i=1}^{N}\begin{cases}\frac{1}{a_i}, & y_i^{\mathrm{final}}=1,\\0, & \text{otherwise.}\end{cases}
\end{equation}

\paragraph{Correction scores.}
\begin{equation}
\mathrm{HintAwareCorrectionScore}_i = \begin{cases}\frac{1}{1+f_i+\lambda h_i}, & y_i^{\mathrm{final}}=1,\\0,&\text{otherwise.}\end{cases}
\end{equation}

\subsection{Inner-Loop Metrics}

\paragraph{Transition rates.}
Let
\begin{equation}
L_{\mathrm{inner}}
=
\sum_i \mathbf{1}[z_i^{(0)} \text{ and } z_i^{(1)} \text{ are both labeled}]
\end{equation}
denote the number of labeled inner transitions.
\begin{align}
\mathrm{RepairRate}
\equiv
\mathrm{ReflectiveRepairRate}
&=
\frac{\sum_i \mathbf{1}[z_i^{(0)}=0 \wedge z_i^{(1)}=1]}{\sum_i \mathbf{1}[z_i^{(0)}=0]},\\
\mathrm{PreservationRate}
\equiv
\mathrm{ReflectivePreservationRate}
&=
\frac{\sum_i \mathbf{1}[z_i^{(0)}=1 \wedge z_i^{(1)}=1]}{\sum_i \mathbf{1}[z_i^{(0)}=1]},\\
\mathrm{HarmRate}
\equiv
\mathrm{ReflectiveHarmRate}
&=
\frac{\sum_i \mathbf{1}[z_i^{(0)}=1 \wedge z_i^{(1)}=0]}{\sum_i \mathbf{1}[z_i^{(0)}=1]},\\
\mathrm{NetRepairGain}
\equiv
\mathrm{ReflectiveNetRepairGain}
&=
\frac{\sum_i \mathbf{1}[z_i^{(0)}=0 \wedge z_i^{(1)}=1]-\sum_i \mathbf{1}[z_i^{(0)}=1 \wedge z_i^{(1)}=0]}
{L_{\mathrm{inner}}}.
\end{align}

\paragraph{NeglectRate and TryButFailRate (StillWrong decomposition).}
\begin{align}
\mathrm{NeglectRate} &= \frac{\sum_i \mathbf{1}[z_i^{(0)}=0 \wedge x_i^{(1)}=x_i^{(0)} \wedge z_i^{(1)}=0]}{\sum_i \mathbf{1}[z_i^{(0)}=0]},\\
\mathrm{TryButFailRate} &= \frac{\sum_i \mathbf{1}[z_i^{(0)}=0 \wedge x_i^{(1)}\neq x_i^{(0)} \wedge z_i^{(1)}=0]}{\sum_i \mathbf{1}[z_i^{(0)}=0]}.
\end{align}
\begin{align}
\mathrm{StillWrongTotal}
\equiv
\mathrm{ReflectiveStillWrongRate}
&=
\mathrm{NeglectRate} + \mathrm{TryButFailRate}\\
&=
1-\mathrm{RepairRate}.
\end{align}
Neglect measures pure signal non-response (solver submits identical answer); TryButFail measures effort without success---opposite implications for system design.

\begin{table}[h]
\centering
\small
\caption{Within-attempt reflective-refinement transitions, computed over all labeled inner-loop observations (one per outer-loop attempt per problem), not averaged by tier. RepairRate, NeglectRate, and TryButFailRate are fractions over problems whose initial inner-loop candidate is wrong. NeglectRate means the solver keeps the same wrong answer. TryButFailRate means the solver changes the answer but it remains wrong. StillWrongTotal is the paper display label for ReflectiveStillWrongRate, so Neglect plus TryButFail equals StillWrongTotal. NetRepairGain is the net correctness change produced by the inner review loop: wrong-to-correct transitions minus correct-to-wrong transitions, divided by the number of labeled inner-loop transitions. Tier-by-tier breakdowns are in Figure~\ref{fig:supp_inner_loop_by_tier}.}
\label{tab:inner_loop_transition}
\resizebox{\linewidth}{!}{
\begin{tabular}{lccccccc}
\toprule
Protocol & RepairRate & NeglectRate & TryButFailRate & StillWrongTotal & PreservationRate & HarmRate & NetRepairGain \\
\midrule
PER (role-based)       & 11.0\% & 48.8\% & 40.2\% & 89.0\% & 96.5\% & 3.5\% & $+$5.8 pp \\
Broadcast deliberation & 25.7\% & 26.2\% & 48.1\% & 74.3\% & 93.5\% & 6.5\% & $+$6.7 pp \\
\bottomrule
\end{tabular}
}
\end{table}

\subsection{Review-Conditioned Decomposition}
\label{app:review_conditioned}

Define the count tensor:
\begin{equation}
C_{u,r,v} = \sum_{i=1}^{N}\mathbf{1}[z_i^{(0)}=u \wedge r_i=r \wedge z_i^{(1)}=v], \quad u,v\in\{0,1\},\; r\in\{\mathrm{Revise,Agree}\}.
\end{equation}

\paragraph{Reviewer detection counts.}
\begin{align}
\mathrm{TP} &= C_{0,\mathrm{Revise},1}+C_{0,\mathrm{Revise},0}, \quad \mathrm{FP} = C_{1,\mathrm{Revise},1}+C_{1,\mathrm{Revise},0},\\
\mathrm{FN} &= C_{0,\mathrm{Agree},1}+C_{0,\mathrm{Agree},0}, \quad \mathrm{TN} = C_{1,\mathrm{Agree},1}+C_{1,\mathrm{Agree},0}.
\end{align}

\paragraph{Detection rates and response rates.}
\begin{align}
\mathrm{Precision}
&=
\frac{\mathrm{TP}}{\mathrm{TP}+\mathrm{FP}},
\qquad
\mathrm{Recall}
=
\frac{\mathrm{TP}}{\mathrm{TP}+\mathrm{FN}},\\
\mathrm{FAR}
&=
\frac{\mathrm{FP}}{\mathrm{FP}+\mathrm{TN}},
\qquad
\mathrm{BalAcc}
=
\tfrac{1}{2}(\mathrm{Recall}+1-\mathrm{FAR}),\\
\mathrm{ReviewerGuidedRepairRate}
&=
\frac{C_{0,\mathrm{Revise},1}}{\mathrm{TP}},\\
\mathrm{ReviewerDetectedButNotFixedRate}
&=
\frac{C_{0,\mathrm{Revise},0}}{\mathrm{TP}},\\
\mathrm{MisleadingReviewResistanceRate}
&=
\frac{C_{1,\mathrm{Revise},1}}{\mathrm{FP}},\\
\mathrm{MisleadingReviewHarmRate}
&=
\frac{C_{1,\mathrm{Revise},0}}{\mathrm{FP}}.
\end{align}
\paragraph{Feedback-follow-through metrics and paper-facing aliases.}
The implementation also extracts whether reviewer feedback produces an immediate answer transition in the next revision step. Here ``useful'' and ``misleading'' are evaluator-verified properties of the pre-review candidate: useful revise episodes begin from an evaluator-labeled wrong candidate, while misleading revise episodes begin from an evaluator-labeled correct candidate. Let $F_{\mathrm{elig}}$ denote eligible reviewer-feedback episodes, $F_{\mathrm{useful}}$ evaluator-verified useful revise-feedback episodes, and $F_{\mathrm{mislead}}$ evaluator-verified misleading revise-feedback episodes.
\begin{equation}
\begin{aligned}
\mathrm{ReviewerFeedback}\\
\qquad \mathrm{IncorporationRate}
&=
\frac{\#(\text{followed feedback})}{F_{\mathrm{elig}}}.
\end{aligned}
\end{equation}
\begin{equation}
\begin{aligned}
\mathrm{LegacyCouplingRate}
&\equiv \mathrm{UsefulReviseFeedback}\\
&\qquad \mathrm{IncorporationRate}
=
\frac{\#(\text{followed useful feedback})}{F_{\mathrm{useful}}}.
\end{aligned}
\end{equation}
\begin{equation}
\begin{aligned}
\mathrm{MisleadingReview}\\
\qquad \mathrm{SusceptibilityRate}
&=
\frac{\#(\text{followed misleading feedback})}{F_{\mathrm{mislead}}}.
\end{aligned}
\end{equation}
In the main paper we shorten \emph{LegacyCouplingRate} to \emph{CouplingRate}, because this is the uptake statistic used throughout the review-conditioned decomposition. Operationally, a useful revise episode $i$ is counted as followed if the extracted answer changes from the pre-review candidate $a_i^{-}$ to the immediately following candidate $a_i^{+}$. Extraction is deterministic: we compare boxed answers when present and otherwise compare normalized final-answer strings.

In the main-paper tables, \emph{MisleadingResistance} is a shortened display label for \emph{MisleadingReviewResistanceRate}. Likewise, \emph{CouplingRate} is the paper-facing display label for the legacy uptake statistic above. In plain language, CouplingRate is the fraction of evaluator-verified useful revise episodes in which the immediately following candidate answer differs from the rejected pre-review answer. It is therefore an answer-transition statistic, not a free-form human judgment about whether the solver ``really understood'' the critique. ReviewerGuidedRepairRate and NeglectRate remain separate on purpose: together they distinguish answer change, non-response, and realized repair rather than collapsing them into a single uptake score. The stricter answer-transition audits used to test this metric's stability are front-loaded in Table~\ref{tab:strict_coupling_audit}.

If $M=\mathrm{TP}+\mathrm{FP}+\mathrm{FN}+\mathrm{TN}$ is the total number of review-labeled instances, then the net quantity used in the main-paper net-value comparison is
\begin{equation}
\mathrm{NetCouplingValuePer1000}
=
1000\cdot\frac{C_{0,\mathrm{Revise},1}-C_{1,\mathrm{Revise},0}}{M}.
\end{equation}
This is equivalent to \((\mathrm{TP}\times\mathrm{ReviewerGuidedRepairRate})-(\mathrm{FP}\times\mathrm{MisleadingReviewHarmRate})\) per 1{,}000 review instances.

\begin{table}[h]
\centering
\small
\caption{Supplemental review-conditioned counts and detection summary. The main paper focuses on precision, recall, coupling, and repair, so this appendix table reports the detection counts needed to contextualize those quantities without introducing partially populated columns.}
\label{tab:appendix_review_conditioned_full}
\resizebox{\linewidth}{!}{
\begin{tabular}{lcccc}
\toprule
Protocol & Review instances & Precision & Recall & AvgEvalCalls/problem \\
\midrule
PER (role-based) & 10{,}804 & 0.861 & 0.754 & 2.19 \\
Broadcast deliberation & 11{,}587 & 0.644 & 0.872 & 1.35 \\
\bottomrule
\end{tabular}
}
\end{table}

\begin{table}[h]
\centering
\small
\setlength{\tabcolsep}{2pt}
\renewcommand{\arraystretch}{1.12}
\caption{Review-conditioned decomposition for the main PER and broadcast comparison. Precision and Recall characterize reviewer detection quality. CouplingRate is the paper-facing immediate answer-change statistic on evaluator-verified useful review episodes. ReviewerGuidedRepairRate measures realized repair after reviewer-positive wrong cases. MisleadingResistance is the paper display label for MisleadingReviewResistanceRate.}
\label{tab:review_conditioned_decomposition_appendix}
\begin{tabular}{@{}>{\raggedright\arraybackslash}p{2.7cm}cc>{\centering\arraybackslash}p{2.2cm}>{\centering\arraybackslash}p{3.1cm}>{\centering\arraybackslash}p{2.8cm}@{}}
\toprule
\textbf{Protocol} & \textbf{Precision} & \textbf{Recall} & \textbf{\shortstack[c]{Coupling\\Rate}} & \textbf{\shortstack[c]{ReviewerGuided\\RepairRate}} & \textbf{\shortstack[c]{Misleading\\Resistance}} \\
\midrule
PER (role-based) & 0.861 & 0.754 & 0.336 & 0.051 & 0.921 \\
Broadcast deliberation & 0.644 & 0.872 & 0.935 & 0.286 & 0.708 \\
\bottomrule
\end{tabular}
\end{table}

\section{Protocol Configurations}
\label{app:protocol_configs}

All descriptions in this appendix section refer to the matched paper configs rather than to alternate development configs elsewhere in the broader codebase. Under those paper configs, every acting role and the evaluator use \texttt{gpt-oss-120b} at temperature $0$. Because \texttt{gpt-oss-120b} is public and open-weight, reproducing the model access path does not require proprietary weights; the paper abstracts only over the local OpenAI-compatible serving layer used to host the matched runs on GPU-backed HPC infrastructure.

\begin{table}[h]
\centering
\small
\setlength{\tabcolsep}{2pt}
\renewcommand{\arraystretch}{1.15}
\caption{Matched paper protocol knobs at a glance. This table is intended as a quick reference for the exact comparison ladder used in the paper.}
\label{tab:paper_protocol_knobs}
\begin{tabular}{@{}>{\raggedright\arraybackslash}p{1.5cm}>{\raggedright\arraybackslash}p{1.5cm}>{\raggedright\arraybackslash}p{3.35cm}>{\raggedright\arraybackslash}p{2.95cm}>{\raggedright\arraybackslash}p{3.1cm}@{}}
\toprule
\textbf{Protocol} & \textbf{Outer budget} & \textbf{Inner/pre-submission structure} & \textbf{Feedback surface} & \textbf{Submission gate} \\
\midrule
Baseline LLM & 1 outer attempt & None & Plain evaluator pass/fail only & Direct one-shot solver submission \\
Single-Agent Iterative & Up to 3 outer\newline attempts & No reviewer;\newline question-scoped repair ledgers\newline reused across retries & Hinted evaluator feedback\newline plus prior submitted answer & Same solver resubmits\newline after each failure \\
PER & Up to 3 outer\newline attempts & 2 inner rounds per outer attempt:\newline Planner$\rightarrow$Executor$\rightarrow$\newline Reviewer & Reviewer/evaluator critique\newline routed through separable\newline \texttt{\$\{advice\}} fields & Reviewer can route back\newline to Planner/Executor or\newline approve submission \\
Broadcast & Up to 3 outer\newline attempts & 4 discussion rounds\newline plus up to 2 approval\newline rounds & Shared candidate state\newline plus shared peer history\newline and hinted evaluator\newline feedback & Unanimous re-approval\newline before submission after\newline deliberation \\
\bottomrule
\end{tabular}
\end{table}

\subsection{Prompt Surfaces and Output Contracts}
\label{app:prompt_surfaces}

This subsection records the exact information surfaces that matter for scientific interpretation and for reproduction. We do not print every full template inline because the release package will include the exact config files, but we do make the visible fields and output contracts explicit.

\paragraph{Baseline LLM.}
The baseline solver sees only \texttt{\$\{task\_description\}} and is instructed to solve in one shot. Its output contract is strict: the final two non-empty lines must be
\texttt{Final Answer: <answer>} and then \texttt{\textbackslash boxed\{<answer>\}}. The evaluator sees \texttt{\$\{task\_description\}}, the extracted solver submission \texttt{\$\{solution\}}, and evaluator-only reference material \texttt{\$\{result\}}; it must output exactly \texttt{Verdict}, \texttt{Correctness}, and \texttt{Response}.

\paragraph{Single-Agent Iterative.}
The first attempt uses the same direct-solve interface as the baseline. Later attempts additionally expose \texttt{\$\{former\_solution\}} and a compact question-scoped memory block in \texttt{\$\{advice\}} that summarizes prior failure information without revealing the gold answer. The solver keeps the same two-line final-answer contract as the baseline. The evaluator contract is the same as in the baseline, except that hinted repair feedback may be returned after a fixed \texttt{FAIL} verdict.

\paragraph{PER.}
PER uses three different acting prompt surfaces. The Planner sees \texttt{\$\{task\_description\}}, \texttt{\$\{previous\_plan\}}, and \texttt{\$\{advice\}}, and must open with a \texttt{DELTA:} section that states concrete plan updates without producing a final answer. The Executor sees \texttt{\$\{task\_description\}}, the Planner output in \texttt{\$\{solution\}}, and \texttt{\$\{advice\}}; its final line must be exactly one \texttt{\textbackslash boxed\{<answer>\}}. The Reviewer sees \texttt{\$\{task\_description\}}, \texttt{\$\{previous\_plan\}}, the current candidate in \texttt{\$\{preliminary\_solution\}}, and \texttt{\$\{advice\}}. It must emit a short diagnosis/fix block and then exactly one final route tag: \texttt{[Route:Planner]}, \texttt{[Route:Executor]}, or \texttt{[Agree]}. A \texttt{Final Answer For Evaluator} block is allowed only on \texttt{[Agree]}. The evaluator again sees only \texttt{\$\{task\_description\}}, the extracted candidate \texttt{\$\{solution\}}, and evaluator-only reference material \texttt{\$\{result\}}.

\paragraph{Broadcast deliberation.}
Each deliberator sees \texttt{\$\{task\_description\}}, outer-loop feedback in \texttt{\$\{advice\}}, the current shared candidate \texttt{\$\{candidate\_answer\}}, its role description, private deferred notes \texttt{\$\{private\_memory\}}, a current turn label, its own speaking history, the current phase, and a phase-specific instruction. Deliberators may emit confidence polls and may optionally propose
\texttt{Candidate Answer: \textbackslash boxed\{<answer>\}} inside discussion. The shared transcript, candidate state, and approval history together determine whether the protocol can submit. The evaluator contract is parallel to PER: it judges the extracted final submission against evaluator-only reference material and returns \texttt{Verdict}, \texttt{Correctness}, and \texttt{Response}.

\paragraph{Representative prompt excerpts.}
To make the paper-facing interface more concrete, we include two short literal excerpts here. The release package will contain the full exact YAML templates.

\begin{quote}
\small
\textbf{PER Reviewer excerpt.} ``You are the Reviewer in a Planner$\rightarrow$Executor$\rightarrow$Reviewer pipeline for difficult math problems.'' The prompt then makes the routing contract explicit: ``If \texttt{\$\{advice\}} contains a failure indicator such as \texttt{FAIL} or \texttt{Score: False}, you MUST end with ONLY \texttt{[Route:Planner]} or \texttt{[Route:Executor]}. You MUST NOT output \texttt{[Agree]}.'' It also fixes the response shape with a strict format of \texttt{Diagnosis:}, \texttt{Fix Instruction:}, an optional \texttt{Final Answer For Evaluator:} block only on agreement, and ``End with EXACTLY ONE tag on the last line.''
\end{quote}

\begin{quote}
\small
\textbf{Broadcast deliberator excerpt.} ``You are one peer in a small math deliberation group.'' The shared-state surface is then exposed directly: ``Outer-loop feedback from the evaluator or prior attempt: \texttt{\$\{advice\}}'' and ``Current shared candidate answer: \texttt{\$\{candidate\_answer\}}.'' The prompt also makes proposal behavior explicit: ``If you want the group to consider a concrete final answer, include \texttt{Candidate Answer: \textbackslash boxed\{<answer>\}}.'' The current phase and phase-specific instruction are visible in the same active prompt context as the candidate answer.
\end{quote}

\begin{quote}
\small
\textbf{Evaluator excerpt.} ``You are the Evaluator in a Planner$\rightarrow$Executor$\rightarrow$Reviewer loop for olympiad-style mathematics.'' The prompt then fixes the judge contract: ``Decide whether the submitted candidate final answer is mathematically equivalent to the ground truth,'' ``Do not reveal, quote, restate, or hint at the ground truth,'' and ``Output a clear verdict label: \texttt{Verdict: PASS} or \texttt{Verdict: FAIL}.'' The required output footer is exactly three lines: \texttt{Verdict: PASS or FAIL}, \texttt{Correctness: 0 or 1}, and \texttt{Response: <brief non-leaking repair hint>}.
\end{quote}

\paragraph{Answer extraction and normalization example.}
The paper-critical extraction path first takes the last well-formed \texttt{\textbackslash boxed\{...\}} answer if one exists; otherwise it falls back to the last non-empty line and strips a trailing period. For example, if a solver ends with
\begin{quote}
\small\ttfamily
Final Answer: 1/2\\
\boxed{1/2}.
\end{quote}
the extracted answer is \texttt{1/2}. If no boxed answer is present and the final line is \texttt{n \textbackslash neq 5.}, the fallback extractor returns \texttt{n \textbackslash neq 5}. For numeric-only exact matching, commas are removed and decimal strings are normalized through \texttt{Decimal(...).normalize()}, so strings such as \texttt{1}, \texttt{1.0}, and \texttt{1.} compare equal in the numeric path. Symbolic and set-valued Omni-MATH judgments continue through the evaluator/rule path rather than this numeric-only fallback.

\subsection{Reproducibility Artifact, Compute Path, and Asset Licenses}
\label{app:artifact_release}

The acting/evaluator model is the public OpenAI checkpoint \texttt{openai/gpt-oss-120b}. In our paper runs, those public weights are served through a public OpenAI-compatible inference endpoint on GPU-backed HPC infrastructure, while paper-scale benchmark orchestration runs on standard GPU-backed HPC nodes that submit batched benchmark shards to that endpoint rather than performing model inference locally. Our jobs use the CPU side of those nodes while calling the remote inference endpoint.

Paper-scale first-pass jobs use one mixed CPU/GPU node per submitted batch job, with worker parallelism 10 for broadcast and 20 for PER, single-agent iterative, and baseline. Representative first-pass batch-job wall times range from 00:27 to 03:53; the appendix and report artifact also record per-tier/per-mode wall-clock summaries, while the main paper reports protocol-level token cost, evaluator calls, and hard-tier wall-clock behavior.

The public release website is available at \url{https://huggingface.co/spaces/AgentsSci/scientific-agent-protocol-traces-site}. Code and data will be released after institutional approval. The staged release package will contain the exact paper-critical config, runtime, evaluation, and reporting files needed to reproduce the matched benchmark results. The package is organized around a shallow top level---\texttt{README.md}, \texttt{configs/paper/}, \texttt{data/omni-math-2-filtered/}, \texttt{couplingbench/}, \texttt{couplingbench\_cli/}, \texttt{scripts/}, \texttt{report/}, and \texttt{figures/}---so that readers can see the paper-critical path immediately. Paper-scale throughput used public GPU-backed HPC infrastructure, but small-scale spot checks do not require those exact systems once the public checkpoint is served through a compatible endpoint.

\begin{table}[h]
\centering
\footnotesize
\caption{Staged release artifact inventory for the paper-critical runtime and analysis path. Paths are relative to the release package root.}
\label{tab:artifact_inventory}
\begin{tabular}{p{3.5cm}p{4.3cm}p{4.5cm}}
\toprule
Artifact path(s) & Contents & Role in reproduction \\
\midrule
\path{configs/paper/} & Exact matched configs for baseline, single-agent iterative, PER, broadcast, ACK-required, and EMB (config key \texttt{embedded-advice}) & Reproduces the protocol ladder and within-PER intervention suite used in the paper \\
\path{README.md} plus top-level dependency files & Environment assumptions, smoke-test commands, full-benchmark command patterns, and analysis regeneration steps & Gives the release package an executable entry point rather than only a file dump \\
\path{couplingbench_cli/benchmark.py} & Main benchmark runner with trace and artifact-copy support & Executes a paper config on a dataset shard or full manifest \\
\path{couplingbench_cli/rq1_outcome_cost.py} & RQ1 outcome/cost analysis & Regenerates the main outcome and efficiency tables \\
\path{couplingbench_cli/rq2_recovery.py} & RQ2 recovery analysis & Rebuilds OuterPass@$k$ / CRR@$k$ summaries \\
\path{couplingbench_cli/rq3_protocol_refinement.py}, \path{couplingbench_cli/collaboration_decomposition.py} & Review-conditioned and collaboration-decomposition analysis & Rebuilds the main RQ3 repair and coupling tables \\
\path{couplingbench_cli/strict_coupling_rate.py} & Stricter answer-transition robustness analysis & Produces strict and equivalence-aware coupling summaries plus audit samples \\
\path{scripts/bootstrap_problem_clustered_metrics.py} & Problem-level and problem-clustered bootstrap uncertainty analysis over saved paper results & Recomputes paired PER--broadcast outcome and mechanism intervals without rerunning model inference \\
\path{scripts/export_paper_per_problem_tables.py} plus \path{report/per_problem/} & Compact per-problem outcome and review-summary exports keyed by stable paper IDs & Lets reviewers inspect problem-level behavior and reconnect aggregate tables to individual benchmark items without opening raw traces \\
\path{couplingbench/evaluation/} & Evaluator logic, answer extraction, and equivalence handling & Fixes the evaluator contract and answer-judging path \\
\path{scripts/preprocess_omni_math.py} plus the paper manifest & Dataset filtering and manifest construction & Recreates the exact 4{,}181-problem benchmark slice and tier assignments used in the paper \\
\path{scripts/} & Batch orchestration and report rebuilding utilities & Reconstructs top-level tier/mode report folders and aggregate CSVs from per-shard outputs \\
\path{report/} and figure-input CSVs & Generated paper tables and figure inputs packaged with the staged release artifact, including compact per-problem tables and paper-facing intervention summaries & Lets readers verify the exact tables and plots shown in the paper without rerunning the full benchmark first \\
\path{figures/plot_all_figures.py} & Figure-regeneration helper over packaged report tables & Rebuilds the paper-facing figures from the saved aggregate CSVs \\
\bottomrule
\end{tabular}
\end{table}

\paragraph{Reproduction path.}
The release package will support three reproduction levels. First, a smoke test reruns any paper config on a small shard once \texttt{openai/gpt-oss-120b} is served through an OpenAI-compatible endpoint. Second, a paper-scale rerun executes \texttt{couplingbench\_cli/benchmark.py} over the paper manifest or over tier shards and then rebuilds the aggregate report folders with \texttt{build\_tier\_mode\_report\_folder.py}. Third, table and figure regeneration reruns the RQ1/RQ2/RQ3 analysis scripts, the stricter coupling analysis, the problem-clustered bootstrap script, the compact per-problem export script, and the associated plotting utilities over those rebuilt reports. The lightweight package does not currently ship the full raw trace tree for the completed intervention reruns; instead it includes paper-facing intervention summary CSVs alongside the main four-mode aggregate tables. The exact commands, expected environment variables, packaged report-table filenames, and example output directory structure will be listed in the release \texttt{README.md}.

We use only public assets. \texttt{gpt-oss-120b} is released under Apache 2.0 together with the OpenAI gpt-oss usage policy \cite{openai_gptoss_release,openai_gptoss_model}. The public dataset cards for Omni-MATH (\url{https://huggingface.co/datasets/omniomni/omni-math}) and Omni-MATH-2 (\url{https://huggingface.co/datasets/martheballon/Omni-MATH-2}) list MIT licensing; in the paper we use the filtered Omni-MATH-2 subset described in Section~\ref{ssec:omnimath2} while crediting both the original Omni-MATH and Omni-MATH-2 benchmark authors \cite{omnimath,omnimath2}. The evaluator replay and reduced cross-family addendum additionally use the public models \texttt{Meta-Llama-3.1-70B-Instruct} and \texttt{gemma-3-27b-it} for robustness analysis, under their respective model licenses and usage terms.

\subsection{Exact Within-PER Coupling Probe Details}
\label{app:coupling_probe_details}

This subsection makes the two within-PER interventions explicit at the prompt/interface level. Across both variants, we keep the same model family, full benchmark, planner prompt, reviewer prompt, evaluator contract, inner-round budget, and outer-loop budget. The only intended change is how reviewer critique is surfaced to the downstream solver.

\paragraph{ACK-required.}
Base PER already routes reviewer/evaluator critique through a separable downstream advice field. The ACK-required variant keeps that routing path unchanged and flips the paper-config control \texttt{require\_explicit\_critique\_uptake=true}. Operationally, this augments the solver-side instruction with a mandatory acknowledgment preface before the next candidate. In the blind paper config, the solver is required to begin its response with a sentence of the form ``I acknowledge that the reviewer noted \ldots'' and then provide the revised candidate. This probe therefore tests whether verbal compliance enforcement alone can increase substantive uptake while leaving the rest of the PER routing structure intact. A useful way to read the negative result is that the new sentence creates an additional cheap completion target under the same routed-advice interface: the solver can satisfy the acknowledgment requirement locally without making the critique bind on the candidate state. In the main paper we therefore interpret its failure as being specific to this prompt configuration rather than as proof that every acknowledgment-style control must fail.

\paragraph{EMB.}
EMB is the paper-facing short name for the config key \texttt{embedded-advice}. This variant changes the critique surface rather than adding a new acknowledgment obligation. It flips the paper-config control \texttt{embed\_action\_feedback\_in\_history=true}. Reviewer feedback is inserted directly into the executor's primary working context together with the current candidate, so the next solver step reads the critique as part of the same active context used to generate the revision. The role hierarchy remains planner--executor--reviewer, but critique is no longer presented only as a separable role-local advice block. This is the smallest shared-context intervention we test inside the PER family. It therefore probes whether a more shared feedback surface improves follow-through and repair without switching all the way to full broadcast deliberation. When this variant improves final outcomes without uniformly improving the narrower review-conditioned subset metrics relative to base PER, we read that as an upstream-trajectory effect rather than a contradiction: the set of evaluator-verified useful-review episodes is itself protocol-induced, so some of the end-to-end gain can come from solving or simplifying trajectories before they enter that subset.

\paragraph{Controls held fixed across the probes.}
Across ACK-required and EMB, the acting/evaluator model family, full benchmark, evaluator prompt, evaluator correctness contract, outer-loop budget, inner-round budget, and PER role topology remain fixed. The intended manipulation is therefore the critique-to-solver interface, not a change in model family, dataset coverage, or retry budget.

\paragraph{Completed suite status.}
The current intervention results in the main paper come from completed full-benchmark reruns over all 4{,}181 problems. ACK-required and EMB remain directly comparable on identical questions throughout the full benchmark. In the main paper we therefore use ACK-required versus EMB as the cleanest within-family causal comparison, while also showing base PER and broadcast as exact full-benchmark baselines to illustrate the broader \emph{Base PER $\rightarrow$ EMB $\rightarrow$ Broadcast} progression toward more directly shared feedback surfaces.

\begin{table}[h]
\centering
\footnotesize
\setlength{\tabcolsep}{3pt}
\renewcommand{\arraystretch}{1.08}
\caption{Completed within-PER intervention suite aggregate over the full benchmark ($N=4{,}181$). Base PER and broadcast are exact full-benchmark baselines; ACK-required and EMB are directly matched on identical questions. UsefulCoupling is the paper-facing display label for UsefulReviseFeedbackIncorporationRate. EMB should be read as the smallest shared-context step inside PER rather than as a full replacement for broadcast-style shared deliberation.}
\label{tab:completed_coupling_intervention_appendix}
\begin{tabular}{@{}lccccccc@{}}
\toprule
\textbf{Protocol} & \textbf{\shortstack[c]{Final\\PassRate}} & \textbf{Pass@1} & \textbf{\shortstack[c]{HintAware\\Correction}} & \textbf{AvgTokens} & \textbf{NeglectRate} & \textbf{\shortstack[c]{ReviewerGuided\\Repair}} & \textbf{\shortstack[c]{Useful\\Coupling}} \\
\midrule
PER                        & 85.2\% & 72.8\% & 70.4\% & 400K & 0.488 & 0.051 & 0.336 \\
PER + ACK-required         & 82.5\% & 71.0\% & 68.9\% & 402K & 0.792 & 0.032 & 0.200 \\
PER + Embedded             & 86.3\% & 74.5\% & 72.0\% & 422K & 0.698 & 0.044 & 0.247 \\
Broadcast                  & 89.2\% & 78.6\% & 81.8\% & 616K & 0.262 & 0.286 & 0.935 \\
\bottomrule
\end{tabular}
\end{table}

\begin{table}[h]
\centering
\small
\caption{FinalPassRate by tier for the completed full-benchmark intervention suite. The small-bin caution remains the same as in the main benchmark: tier 3 has $n=20$ and tier 10 has $n=15$, so both should be interpreted cautiously.}
\label{tab:coupling_intervention_tier_finalpass}
\begin{tabular}{lcccc}
\toprule
Tier & Base PER & PER + ACK-required & PER + Embedded & Broadcast \\
\midrule
1  & 99.1\% & 98.3\% & 99.1\% & 99.1\% \\
2  & 95.0\% & 93.8\% & 94.8\% & 95.2\% \\
3  & 85.0\% & 90.0\% & 90.0\% & 85.0\% \\
4  & 88.0\% & 86.1\% & 89.6\% & 92.6\% \\
5  & 82.7\% & 79.2\% & 84.7\% & 89.2\% \\
6  & 79.3\% & 77.7\% & 80.7\% & 82.9\% \\
7  & 76.8\% & 72.4\% & 76.8\% & 80.7\% \\
8  & 83.5\% & 78.0\% & 82.9\% & 86.6\% \\
9  & 82.2\% & 78.8\% & 82.9\% & 83.6\% \\
10 & 73.3\% & 73.3\% & 86.7\% & 93.3\% \\
\bottomrule
\end{tabular}
\end{table}

\subsection{Broadcast Interaction Topology}
\label{app:broadcast_topology}

The current broadcast protocol is best understood as a hybrid rather than as unrestricted open discussion. Each discussion turn begins with a confidence poll from all deliberators. That poll is used to choose a single public speaker for the turn, but it can also surface a candidate answer before a long public exchange occurs. Once a current candidate exists, the protocol switches into an all-peer approval phase in which every deliberator may approve, reject, or propose a revision. Review-conditioned broadcast quantities in the paper therefore arise from this approval phase, not from the one-speaker public discussion turns.

Candidate ownership in broadcast is similarly protocol-level rather than speaker-level. The current candidate can be promoted from the selected speaker's public message, from the selected speaker's poll candidate, or from the strongest poll candidate available at that turn; later, the approval stage may replace it through the configured peer-revision rule. For the paper configs, this means that broadcast should be read as a protocol that combines turn allocation, shared candidate state, and peer approval, rather than as a simple ``speaker proposes, group judges'' loop.

\begin{table}[h]
\centering
\small
\caption{Critique routing and approval structure in PER and broadcast. This table summarizes the protocol differences most relevant to the coupling gap: how critique is delivered, whether the downstream candidate-producing role can bypass it, and what approval constraints apply before submission. Because the protocols also differ in prompts and memory surfaces, the table should be read as mechanistic support for the coupling interpretation, not as a pure ablation.}
\label{tab:signal_routing}
\begin{tabular}{lll}
\toprule
Architectural dimension & PER & Broadcast \\
\midrule
Reviewer signal delivery & Separable \texttt{\$\{advice\}} field & Embedded in shared chat history \\
Solver exposure to critique & Independent (can skip \texttt{\$\{advice\}}) & Mandatory (part of shared context) \\
After verifier \texttt{FAIL} & Hard-route back to reviewer & Unanimous re-approval required \\
Inner review rounds & 2 & 4 discussion + 2 approval \\
Approval policy & Planner--Executor decision & Unanimous (all deliberators) \\
Decoupling risk & High & Low \\
\bottomrule
\end{tabular}
\end{table}

\subsection{Implementation-Level Memory Contract}
\label{app:memory_contracts}

The main paper describes memory only at the level needed for scientific comparison. This subsection records the implementation details that matter for reproducibility and for reviewer questions about state visibility, persistence, and fairness.

\paragraph{Shared per-question repair ledgers.}
Single-Agent Iterative, PER, and broadcast all maintain the same two structured question-scoped ledgers. The \emph{failed-attempt memory} stores one compact record per rejected evaluator submission: extracted submitted answer, evaluator signal, compact evaluator hint, repair summary, and source stage. The \emph{candidate-revision memory} stores a compact trajectory of candidate proposals and revisions: event id, stage, source, action type, normalized candidate answer, optional review or rationale text, optional parent revision link, and evaluator signal when present. In the paper configs, the failed-attempt ledger keeps at most three records and the candidate-revision ledger keeps at most five. These ledgers are the main mechanism that persists repair-relevant information across outer attempts within a question.

\paragraph{Baseline LLM and Single-Agent Iterative.}
The one-shot baseline instantiates standard per-run chat-memory objects for its solver and evaluator, but because the protocol has only one attempt it does not populate any iterative repair ledger and never routes prior failure information back to the solver. Single-Agent Iterative reuses the same solver/evaluator pattern, but before each new outer attempt it explicitly resets the solver's raw chat history. Cross-attempt persistence therefore comes from the two structured question-level ledgers, not from carrying the full prior transcript forward. This keeps retry behavior tied to compact rejected-answer and candidate-trajectory state rather than to an ever-growing conversational context.

\paragraph{PER.}
PER uses the same failed-attempt and candidate-revision ledgers, but now shares them across Planner, Executor, and Reviewer. Reviewer or evaluator feedback is routed into later role invocations through the \texttt{\$\{advice\}} field together with the current question-memory context, rather than through a shared multi-agent transcript. In addition, the Planner and Reviewer use summary-enabled chat history in the paper config: \texttt{has\_summary=true}, summary updates every three messages, a 300-token summary cap, and a 200-token recent-message budget. The Executor uses standard chat history without summary compression. PER therefore combines shared structured repair memory with role-local summarized conversational state.

\paragraph{Broadcast deliberation.}
Broadcast uses the same two shared question-level ledgers, but adds protocol-specific per-deliberator state. Each deliberator maintains a public chat history rendered through the \texttt{summary\_plus\_recent} path; a private deferred-note buffer capped at eight entries; that agent's own speaking-history buffer capped at eight labeled turns; and a current turn label. In the current paper config, however, deliberators use \texttt{chat\_history} without enabling learned summary updates, so \texttt{summary\_plus\_recent} should be read as the chosen render path rather than as a true running summary. In practice it supplies a bounded recent shared transcript view, with a five-message fallback and heuristics that preserve messages around final-answer markers. The private notes are also specific in purpose: they are deferred contribution notes created for non-selected, high-confidence agents, storing the agent's reason to speak and intended contribution when it was not chosen for the public turn. The paper-config prompt surface exposes evaluator advice, the current shared candidate answer, private deferred notes, current turn label, and own speaking history. This is the main memory distinction between PER and broadcast. PER routes critique as role-specific advice inside a pipeline; broadcast embeds critique and candidate evolution into shared peer discussion while still preserving a small amount of agent-private context.

\paragraph{Reset semantics and evaluator-feedback routing.}
All question-scoped memories, raw chat histories, summaries, deferred notes, and speaking histories are reset between dataset examples when a new problem begins. Acting agents never receive ground-truth answers or official solutions. The evaluator's correctness verdict is computed from the extracted submitted final answer together with evaluator-only reference material. If hint mode is enabled and the verdict is \texttt{FAIL}, a separate post-verdict pass may read compact reasoning summaries to produce a non-answer-revealing repair hint, but the correctness decision is already fixed before that hint step. In broadcast, the resulting evaluator feedback is inserted into the shared deliberation memory as a public message and is also returned as the next outer-round advice string. In Single-Agent Iterative and PER, the evaluator feedback is returned as structured advice and reintroduced together with the question-level repair ledgers on the next attempt.

\paragraph{Prompt-control interpretation.}
Neither PER nor broadcast instructs agents that reviewer signals may be unreliable. The two protocols also use genuinely different role-specific prompt scaffolds and memory surfaces, so we do not interpret their gap as a pure prompt-control ablation. Instead, under matched paper configs, we treat the coupling difference as architectural evidence about how review signals are routed, remembered, and acted on.

\paragraph{Blind-safe implementation and compute setup.}
All experiments in the paper configs were executed through an OpenAI-compatible inference endpoint on GPU-backed HPC infrastructure, with batched runs orchestrated over benchmark shards rather than by interactive single-problem prompting. In the public release artifact path, this corresponds to public \texttt{gpt-oss-120b} weights served through a public inference endpoint with paper-scale batch orchestration on public HPC compute. The scientific conclusions do not depend on a particular serving stack: all protocols use the same acting model, the same evaluator model, the same question set, and the same paper-config attempt budgets, while protocol-level cost is reported directly through generated tokens, model/API calls, evaluator calls, and wall-clock summaries.

\subsection{Artifact Paths For Metric Reproduction}
\label{app:artifact_paths}

The matched experiment configs referenced throughout the paper live in the release package's \texttt{configs/paper/} directory. The benchmark runner copies the resolved config into each run directory for traceability, while the main-paper metrics and review-conditioned summaries are reproduced through the analysis entry points listed in Table~\ref{tab:artifact_inventory}. In particular, RQ1 tables come from \texttt{rq1\_outcome\_cost.py}, RQ2 recovery tables from \texttt{rq2\_recovery.py}, the review-conditioned repair tables from \texttt{rq3\_protocol\_refinement.py} and \texttt{collaboration\_decomposition.py}, and stricter answer-transition robustness from \texttt{strict\_coupling\_rate.py}. The figure-generation utilities used for the draft are likewise included in the release package so that readers can regenerate both the tables and the plots from the same matched paper runs.

For the two new ``still wrong'' metrics, the implementation logic is intentionally simple:
\begin{enumerate}[leftmargin=1.5em,itemsep=0.15em]
  \item Filter to labeled inner-loop transitions with both initial and final correctness labels.
  \item Restrict to transitions whose initial candidate is wrong.
  \item Count \emph{Neglect} when the final candidate is still wrong and the normalized final answer is unchanged from the initial one.
  \item Count \emph{TryButFail} when the final candidate is still wrong but the normalized final answer changed.
  \item Divide both counts by the number of wrong-initial transitions to obtain NeglectRate and TryButFailRate.
\end{enumerate}
This corresponds exactly to the implementation of the paper's inner-loop metric module: \texttt{ReflectiveNeglectRate} is ``wrong $\rightarrow$ same wrong answer,'' while \texttt{ReflectiveTryButFailRate} is ``wrong $\rightarrow$ changed-but-still-wrong answer.'' Readers who want to port these metrics to another stack therefore need only candidate-answer normalization, initial/final correctness labels, and review-episode boundaries rather than the full experimental runtime.

\section{Tier Sample Breakdown and Domain Distribution}
\label{app:tier_breakdown}

\begin{itemize}
\item Full per-tier problem counts, domain distribution, and sample size notes.
\item The ten-tier difficulty axis is inherited from the original Omni-MATH metadata rather than constructed solely for LLM benchmarking. In Omni-MATH, the benchmark designers first organize sources by real Olympiad/contest hierarchy, then assign per-problem difficulty scores mainly from AoPS community ratings; only problems not covered by AoPS:Ratings receive GPT-4o-based difficulty assignment \cite{omnimath}. Omni-MATH~2-Filtered keeps these inherited metadata while cleaning solvability and evaluation issues \cite{omnimath2}.
\item As a result, the tier variable is a hybrid metadata scale: mostly human- and community-grounded, with limited LLM assistance for missing ratings. The resulting tier imbalance is inherited from the benchmark distribution itself rather than introduced by our protocol evaluation. The tier axis is therefore useful for stratified analysis, but it should not be overread as a perfectly smooth model-calibrated latent difficulty axis. This is one reason local wiggles in tier-wise protocol curves are not, by themselves, unexpected.
\item Per-tier counts in the filtered benchmark are: tier 1 = 116, tier 2 = 517, tier 3 = 20, tier 4 = 1{,}106, tier 5 = 1{,}078, tier 6 = 497, tier 7 = 358, tier 8 = 328, tier 9 = 146, tier 10 = 15.
\item Tiers 3 and 10 are especially small and should be interpreted cautiously; tier 1 is moderate rather than tiny, and tiers 2 and 4--9 are well populated.
\item Domain metadata are retained as fine-grained sub-domain paths from Omni-MATH-2-filtered; any coarser rollups such as number theory, combinatorics, algebra, geometry, analysis, and other are analysis-side aggregations rather than raw dataset labels.
\end{itemize}

\section{Difficulty-Tier Extended Results}
\label{app:crr_tier}

\begin{figure}[t]
\centering
\includegraphics[width=\linewidth]{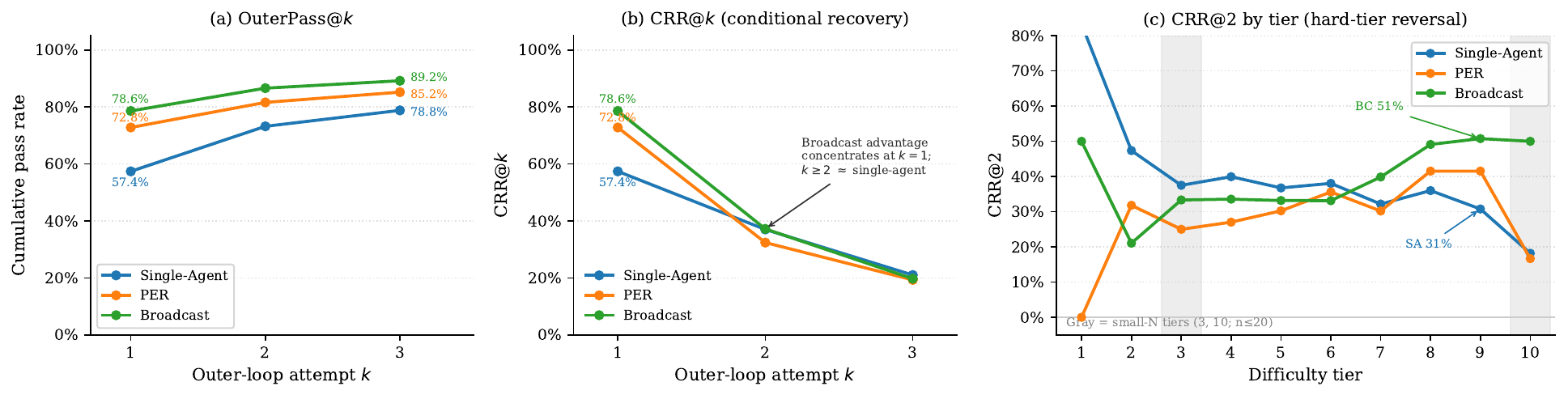}
\caption{\textbf{(a)} Cumulative outer-loop pass rate across three outer attempts. \textbf{(b)} Conditional recovery rate at attempt $k$ (CRR@$k$), meaning the probability of solving the problem at attempt $k$ given failure on all earlier attempts. Broadcast's advantage concentrates at the first attempt (78.6\% first-pass success), while CRR@2 is nearly tied with Single-Agent (37.2\% vs.\ 37.0\%). This makes CRR@2 the key quantity when verifier calls are expensive: a second verifier call helps only if the protocol is meaningfully more likely to solve the problem on that attempt. \textbf{(c)} CRR@2 by difficulty tier. The aggregate tie hides a large broadcast advantage on hard tiers 7--10, including about a 20 percentage point gap on tier 9. Gray bands mark the small-$N$ tiers (3 and 10; $n\le20$).}
\label{fig:rq2_recovery}
\end{figure}

\begin{figure}[h]
\centering
\includegraphics[width=\linewidth]{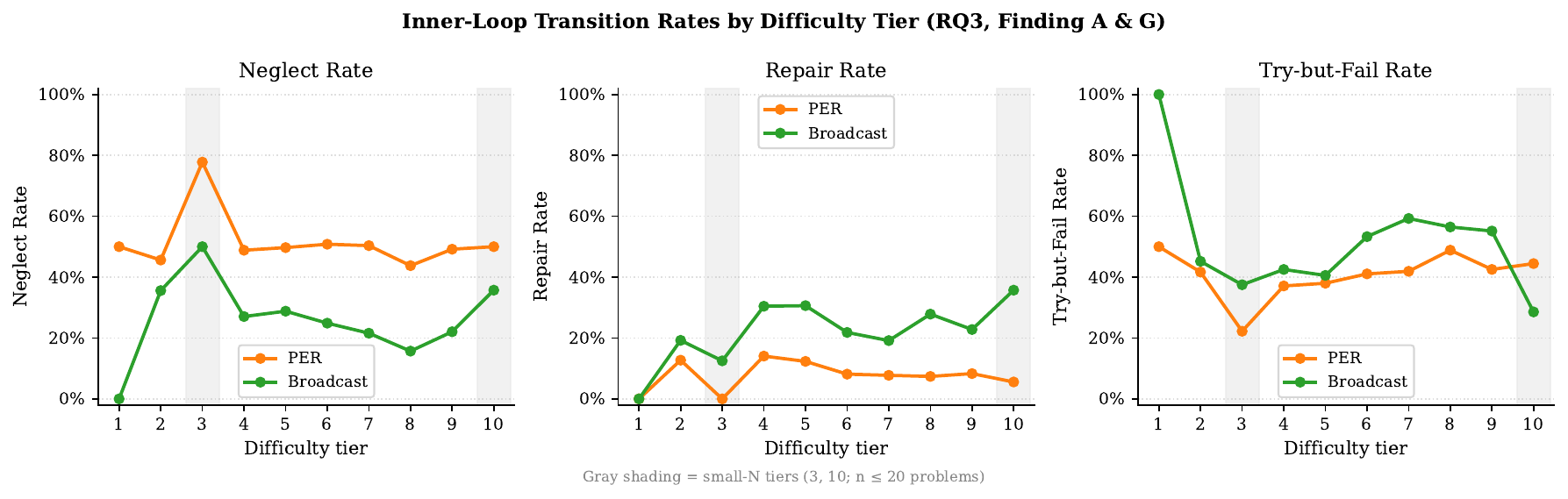}
\caption{Inner-loop transition rates by difficulty tier for PER and Broadcast. The three rates are RepairRate, NeglectRate, and TryButFailRate. PER's NeglectRate stays substantially higher across the benchmark, ranging from 0.44 to 0.78; over the better-populated tiers 4--9, the middle 50\% of PER values lie between 0.45 and 0.51, compared with 0.16 to 0.29 for Broadcast. Because the paper configs do not vary by tier, this gap is consistent with a structural rather than difficulty-specific explanation.}
\label{fig:supp_inner_loop_by_tier}
\end{figure}

\begin{figure}[h]
\centering
\includegraphics[width=\linewidth]{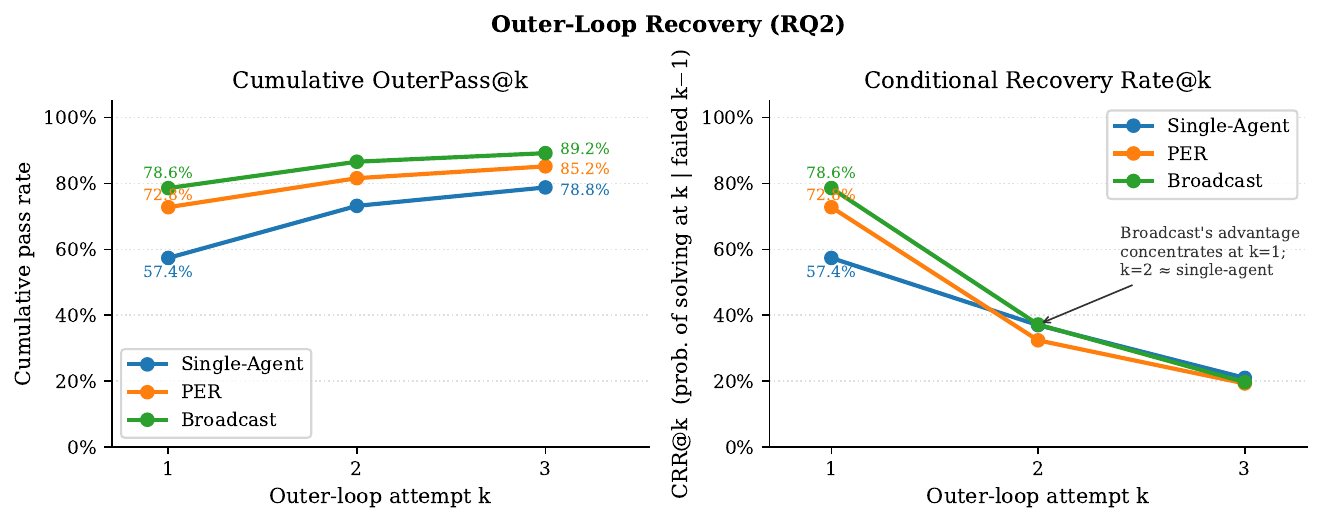}
\caption{Outer-loop recovery curves for Single-Agent Iterative, PER, and Broadcast across three outer attempts. The left panel shows cumulative pass rate after each attempt. The right panel shows conditional recovery rate, meaning the probability of solving the problem on attempt $k$ given failure on all earlier attempts. Broadcast achieves the highest cumulative pass rate, while its conditional recovery advantage over Single-Agent is concentrated on harder tiers rather than appearing uniformly at every attempt.}
\label{fig:supp_recovery_gain}
\end{figure}

\begin{itemize}
\item Full per-tier FinalPassRate, Pass@1, OuterPass@$k$, CRR@$k$ for all protocols.
\item CRR@2 by tier (main paper Figure~\ref{fig:rq2_recovery}(c)) reveals the difficulty-gated broadcast advantage: tiers 7--10 show broadcast CRR@2 consistently above single-agent; tier 9: broadcast 50.8\%, single 30.8\%.
\item Figure~\ref{fig:supp_recovery_gain} (below) shows the full-size aggregate recovery curves (panels (a) and (b) of Figure~\ref{fig:rq2_recovery}).
\item Figure~\ref{fig:supp_inner_loop_by_tier} shows inner-loop transition rates (RepairRate, NeglectRate, TryButFailRate) per tier for PER and broadcast.
\end{itemize}

\begin{table}[h]
\centering
\small
\caption{Verifier-guided outer-loop recovery under hinted feedback (default), $N=4{,}181$, with at most three outer attempts. $\Delta$@$k$ is the increase in cumulative pass rate from attempt $k{-}1$ to attempt $k$, measured in percentage points. CRR@$k$ is the probability of solving at attempt $k$ given failure on all prior attempts. OuterPass@3 equals FinalPassRate for each protocol.}
\label{tab:outer_loop_recovery}
\resizebox{\linewidth}{!}{
\begin{tabular}{lccccccc}
\toprule
Protocol & OuterPass@1 & $\Delta$@2 (pp) & OuterPass@2 & $\Delta$@3 (pp) & OuterPass@3 & CRR@2 & CRR@3 \\
\midrule
Single-Agent Iterative & 57.4\% & $+$15.7 & 73.2\% & $+$5.6 & 78.8\% & 37.0\% & 21.0\% \\
PER (role-based)       & 72.8\% & $+$\phantom{1}8.8  & 81.6\% & $+$3.5 & 85.2\% & 32.4\% & 19.3\% \\
Broadcast deliberation & 78.6\% & $+$\phantom{1}8.0  & 86.6\% & $+$2.7 & 89.2\% & 37.2\% & 19.7\% \\
\bottomrule
\end{tabular}
}
\end{table}

\begin{table}[h]
\centering
\small
\caption{Final pass rate, in percent, by protocol and difficulty tier. Bold values mark the highest accuracy within each tier. A horizontal rule separates easy tiers (1--3) from medium and hard tiers (4--10). Broadcast is the accuracy leader on tiers 4--10, while the protocols stay within 5 percentage points of one another on the easy tiers. Gray entries show the one-shot Baseline LLM.}
\label{tab:tier_accuracy}
\begin{tabular}{lrrrr}
\toprule
Tier & \textcolor{gray}{Baseline} & Single & PER & Broadcast \\
\midrule
1  & \textcolor{gray}{93} & \textbf{99} & \textbf{99} & \textbf{99} \\
2  & \textcolor{gray}{87} & 93          & \textbf{95} & \textbf{95} \\
3  & \textcolor{gray}{60} & 80          & \textbf{85} & \textbf{85} \\
\midrule
4  & \textcolor{gray}{70} & 86 & 88 & \textbf{93} \\
5  & \textcolor{gray}{56} & 78 & 83 & \textbf{89} \\
6  & \textcolor{gray}{38} & 69 & 79 & \textbf{83} \\
7  & \textcolor{gray}{29} & 61 & 77 & \textbf{81} \\
8  & \textcolor{gray}{30} & 69 & 84 & \textbf{87} \\
9  & \textcolor{gray}{21} & 66 & 82 & \textbf{84} \\
10 & \textcolor{gray}{20} & 53 & 73 & \textbf{93} \\
\bottomrule
\end{tabular}
\end{table}

\begin{table}[h]
\centering
\small
\caption{Protocol guidance by difficulty-tier group. ``Best accuracy'' is the protocol with the highest FinalPassRate. ``Cost-efficient alternative'' is the cheapest protocol within 5 percentage points of that best result. Guidance is split by verifier-cost regime. On tiers 4--10, broadcast is the accuracy leader, while PER is the cost-efficient fallback when verifier calls are cheap.}
\label{tab:reader_guidance}
\resizebox{\linewidth}{!}{
\begin{tabular}{lllll}
\toprule
Tier group & Best accuracy & Cost-efficient alt.\ ($\le$5 pp gap) & Cheap verifier & Expensive verifier \\
\midrule
Easy (tier 1--3) & Single / PER / Broadcast & Baseline or Single & Use Baseline or Single & Use Single (high CRR@1, low token cost) \\
Medium (tier 4--7) & Broadcast & PER & Use PER (1.5$\times$ cheaper than Broadcast) & Use Broadcast (highest OuterPass@1) \\
Hard (tier 8--10) & Broadcast & PER & Use PER if budget-constrained & Use Broadcast (highest first-pass rate) \\
\bottomrule
\end{tabular}
}
\end{table}

\begin{figure}[h]
\centering
\includegraphics[width=\linewidth]{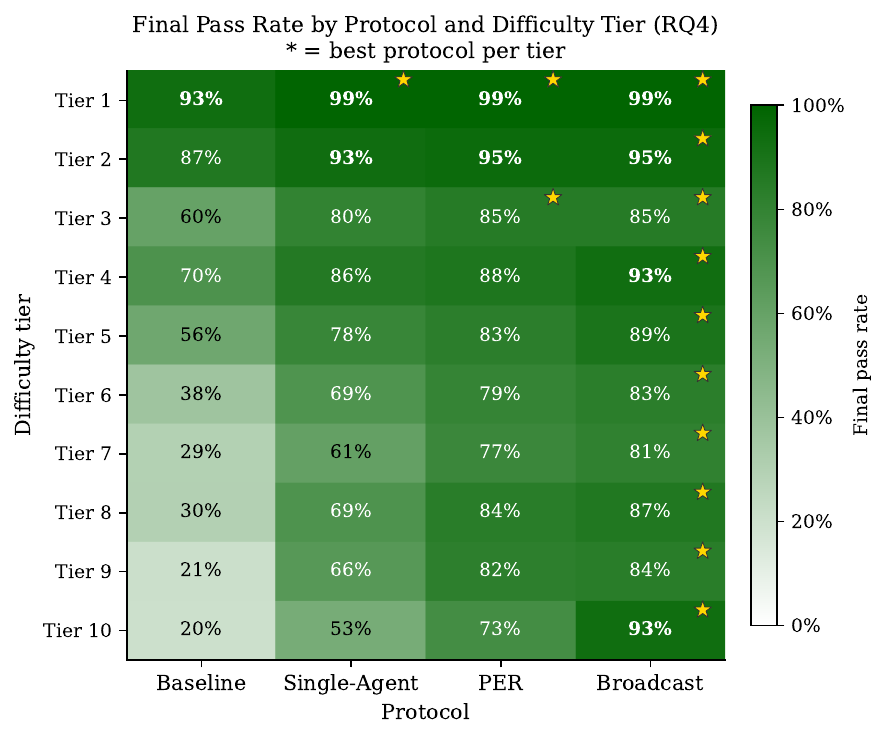}
\caption{FinalPassRate heatmap by protocol and difficulty tier. Each cell shows the percentage of problems solved, and darker color means higher accuracy. Broadcast is the accuracy leader from tier 4 onward; the protocols cluster within 5 percentage points on easy tiers 1--3. Tiers 3 and 10, marked with gray borders, have small sample sizes ($n\le20$) and should be interpreted cautiously.}
\label{fig:tier_heatmap}
\end{figure}

\section{NeglectRate by Tier and Tier-Stratified Signal Routing}
\label{app:neglect_by_tier}

\begin{figure}[h]
\centering
\includegraphics[width=\linewidth]{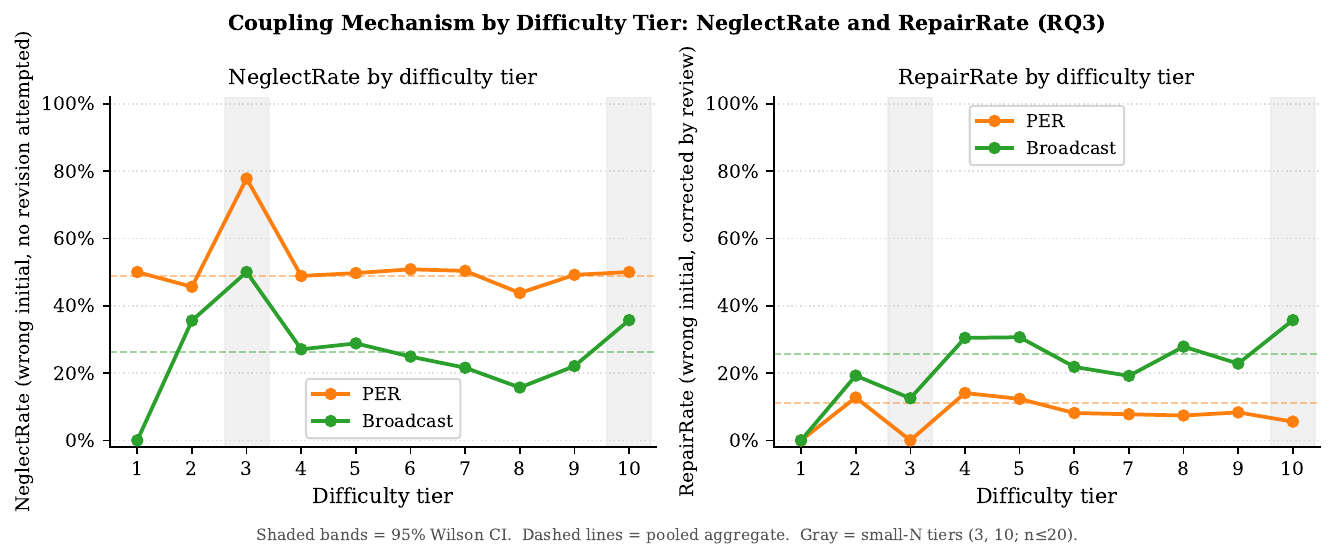}
\caption{NeglectRate (left) and RepairRate (right) by difficulty tier for PER and Broadcast. Shaded bands show 95\% Wilson confidence intervals computed from per-tier transition counts, and dashed horizontal lines show the pooled aggregate values from Table~\ref{tab:inner_loop_transition}. Gray bands mark the small-$N$ tiers 3 and 10 ($n\le20$). PER's NeglectRate stays elevated across the full difficulty range, while Broadcast's RepairRate advantage appears at every tier. The gap does not narrow on the hard tiers, where the role-specialization hypothesis would predict structured review to help most.}
\label{fig:coupling_mechanism_by_tier}
\end{figure}

Figure~\ref{fig:coupling_mechanism_by_tier} shows the two coupling-mechanism metrics that most directly explain the PER--broadcast repair gap: NeglectRate (left) and RepairRate (right), stratified by difficulty tier with 95\% Wilson confidence intervals. Dashed horizontal lines mark the pooled aggregate values reported in Table~\ref{tab:inner_loop_transition}.

The NeglectRate panel shows that PER's neglect is structurally elevated across the full difficulty range, not concentrated at a single tier. For tiers 4--9 (the well-populated tiers, IQR 0.45--0.51), PER's NeglectRate is consistently above 0.44, while broadcast's IQR is 0.16--0.29. The RepairRate panel mirrors this: broadcast's repair advantage is visible at every tier, and the gap persists even at the hardest well-populated tiers (7--9) where problems are presumably harder and all protocols struggle. Together these two panels support the structural interpretation of Finding~A: the coupling gap is not an artefact of easy problems where review is unnecessary, nor of a few extreme tiers, but a consistent property of PER's architecture across the full difficulty spectrum sampled by this benchmark.

\begin{itemize}
\item Per-tier NeglectRate for PER (consistently 0.44--0.78) and Broadcast (0.00--0.50); confidence intervals are widest at tiers 3 and 10 due to small $N$.
\item Full tier-stratified reviewer confusion: TP/FP/FN/TN counts per tier per protocol.
\end{itemize}

\begin{figure}[h]
\centering
\includegraphics[width=\linewidth]{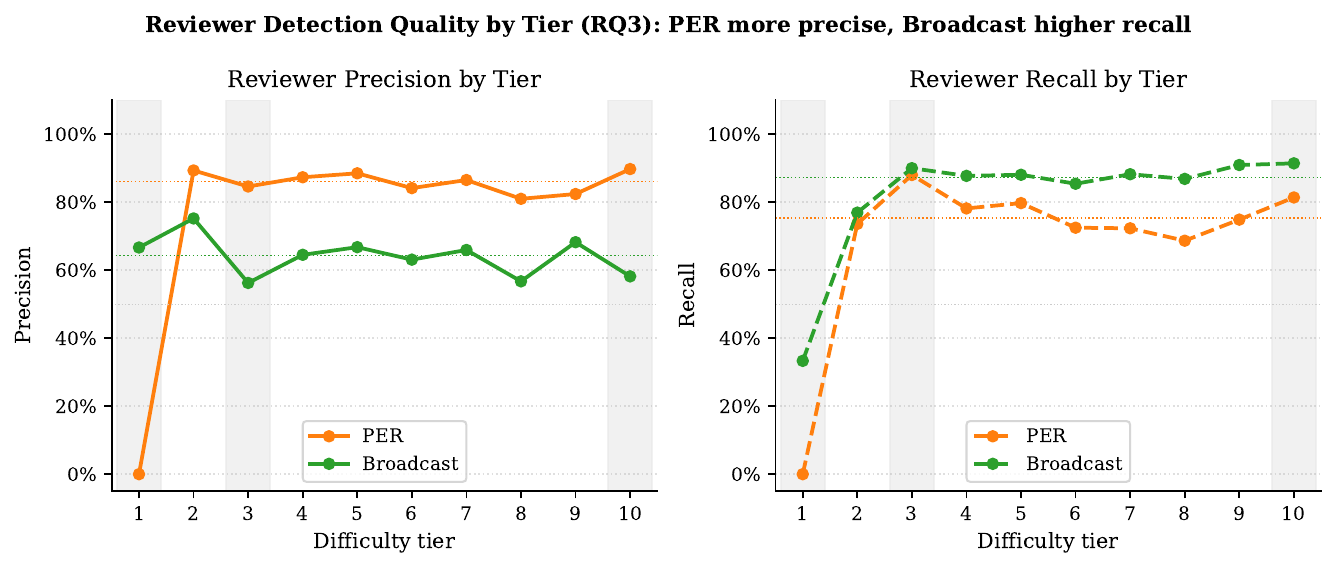}
\caption{Reviewer confusion matrix proportions (TP/FP/FN/TN) by difficulty tier for all protocols.}
\label{fig:supp_reviewer_confusion}
\end{figure}

\begin{figure}[h]
\centering
\includegraphics[width=\linewidth]{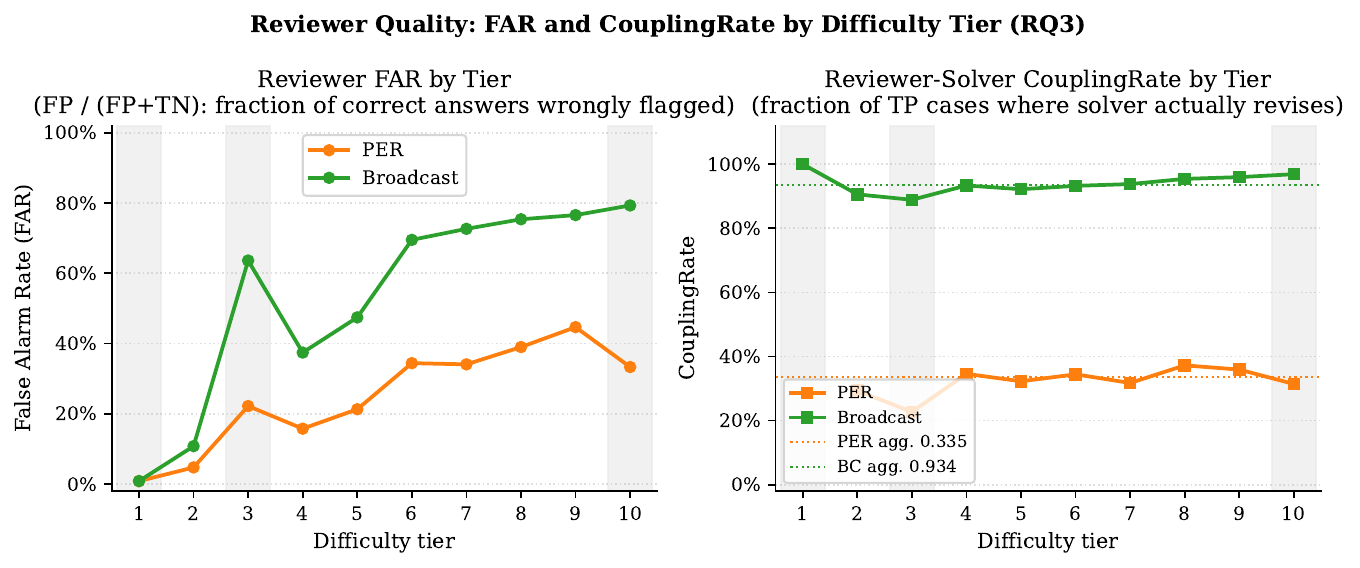}
\caption{Reviewer false alarm rate (FAR) by difficulty tier. FAR measures the fraction of correct answers incorrectly flagged for revision. PER has lower FAR and thus higher precision, but higher NeglectRate.}
\label{fig:supp_reviewer_far}
\end{figure}

\section{Efficiency: Evaluator Calls and Wall-Clock Time}
\label{app:efficiency}

\subsection{Explicit Verifier-Cost Threshold Model}
\label{app:verifier_cost_model}

To make RQ4 operational, we scalarize each protocol $p$ by its average generation-token cost $T_p$, average verifier-call count $V_p$, and final success rate $A_p$. A simple deployment objective is expected cost per solved problem:
\[
K_p(\lambda) \;=\; \frac{T_p + \lambda V_p}{A_p},
\]
where $\lambda$ is the cost of one verifier call measured in generation-token units. The numerator is the expected scalarized cost incurred on one benchmark problem under protocol $p$, and the division by $A_p$ converts that into expected cost \emph{per successful final solve}. Equivalently, over $N$ benchmark problems the protocol incurs total expected cost $N(T_p+\lambda V_p)$ and returns $NA_p$ successful final solves, so the ratio simplifies to the expression above. This is intended as a compact decision aid rather than as a claim that deployment should optimize only one scalar objective. In particular, it treats token cost and verifier cost explicitly, while leaving factors such as implementation preference for explicit role structure outside the objective.

For settings where remaining wrong at the end is itself costly, a richer extension is
\[
J_p(\lambda,\mu) \;=\; T_p + \lambda V_p + \mu (1 - A_p),
\]
where $\mu$ is the penalty for final failure. Here the objective is not ``cost per solved problem,'' but expected scalarized deployment loss \emph{per benchmark problem}: generation cost, plus verifier cost, plus a failure penalty charged exactly when the final output is wrong. This second form is closer to autonomous-science or engineering settings in which an incorrect final output can waste a simulation, robot cycle, or wet-lab assay. The simple $K_p(\lambda)$ model is enough for the paper's main crossover calculations, while $J_p(\lambda,\mu)$ clarifies why high-failure-penalty domains push toward more accurate first-pass protocols even before verifier costs become dominant.

\begin{table}[h]
\centering
\small
\caption{Verifier-cost crossover thresholds under the simple cost-per-solved objective $K_p(\lambda)=\bigl(T_p+\lambda V_p\bigr)/A_p$, using the full-benchmark aggregates from Table~\ref{tab:main_outcome_cost}. Here $\lambda^\star$ is the verifier-call cost, measured in generation-token units, at which the two protocols tie.}
\label{tab:verifier_cost_thresholds}
\begin{tabular}{p{3.0cm}p{2.7cm}p{7.2cm}}
\toprule
Comparison & $\lambda^\star$ (tokens / verifier call) & Interpretation \\
\midrule
PER vs.\ Broadcast & $\approx 2.1\times10^5$ & Below this threshold, PER is cheaper per solved problem on the scalarized objective; above it, Broadcast's higher success rate and lower verifier-call count outweigh its larger generation budget. \\
Single-Agent vs.\ Broadcast & $\approx 9.9\times10^5$ & Broadcast only overtakes single-agent when verifier calls are extremely expensive relative to generation, because single-agent is far cheaper in tokens but has lower first-pass and final success. \\
Single-Agent vs.\ PER & no positive crossover & Under this scalarized objective, single-agent remains cheaper per solved problem on the full-benchmark aggregates. PER is therefore best read as a structured intermediate design point rather than as the global minimizer of $K_p(\lambda)$ under the current paper configs. \\
\bottomrule
\end{tabular}
\end{table}

\begin{table}[h]
\centering
\small
\caption{Difficulty-aware verifier-cost crossover thresholds under the same objective $K_p(\lambda)=\bigl(T_p+\lambda V_p\bigr)/A_p$. Tier-group aggregates are \emph{problem-count weighted}, not simple averages over tier-level metrics: for each group $G$, we compute $T_{p,G}=\sum_{t\in G}N_tT_{p,t}/\sum_{t\in G}N_t$, $V_{p,G}=\sum_{t\in G}N_tV_{p,t}/\sum_{t\in G}N_t$, and $A_{p,G}=\sum_{t\in G}N_tA_{p,t}/\sum_{t\in G}N_t$, where $N_t$ is the number of analyzed problems in tier $t$. The group sizes are $N_{1\text{--}4}=1{,}759$, $N_{5\text{--}6}=1{,}575$, and $N_{7\text{--}10}=847$. The overall row reproduces the full-benchmark aggregate; the grouped rows expose how the scalarized crossover shifts with difficulty.}
\label{tab:tier_group_cost_thresholds}
\begin{tabular}{p{2.4cm}p{2.7cm}p{2.7cm}p{4.3cm}}
\toprule
Tier group & PER vs.\ Broadcast $\lambda^\star$ & Single-Agent vs.\ Broadcast $\lambda^\star$ & Interpretation \\
\midrule
Overall & $\approx 2.08\times10^5$ & $\approx 9.88\times10^5$ & Full-benchmark aggregate used by Figure~\ref{fig:supp_cost_breakeven}. \\
Tiers 1--4 & $\approx 1.32\times10^5$ & $\approx 1.05\times10^6$ & On easier problems, broadcast needs a relatively lower verifier-cost premium than in the global average to overtake PER, but single-agent remains hard to dislodge. \\
Tiers 5--6 & $\approx 1.92\times10^5$ & $\approx 1.04\times10^6$ & Mid-difficulty problems already move the PER/broadcast crossover upward relative to tiers 1--4. \\
Tiers 7--10 & $\approx 2.89\times10^5$ & $\approx 7.18\times10^5$ & Hard tiers shift the PER/broadcast crossover furthest right while making broadcast competitive against single-agent at a lower verifier-cost threshold. \\
\bottomrule
\end{tabular}
\end{table}

\begin{figure}[h]
\centering
\includegraphics[width=\linewidth]{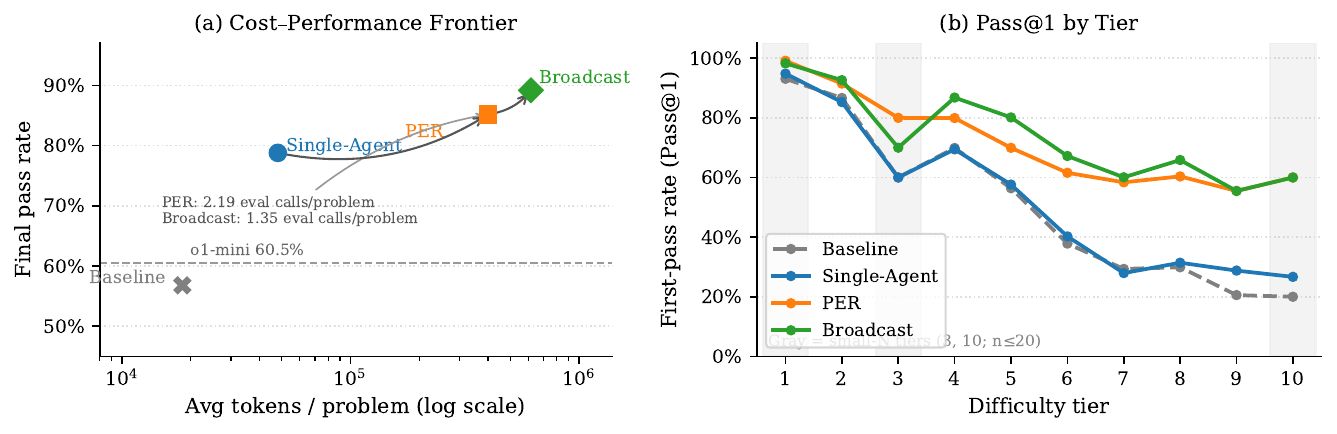}
\caption{\textbf{(a) Raw cost--performance frontier.} Generation tokens per problem versus FinalPassRate for all four protocols. Single-Agent achieves the best tokens-per-extra-solved-problem ratio; PER and Broadcast incur substantially higher generation costs for their respective accuracy gains. Notably, PER is not simply intermediate in cost: it uses \emph{more} verifier calls per problem than Broadcast (2.19 vs.\ 1.35), as annotated in the figure, so it is verifier-heavier despite generating fewer tokens overall. Dashed reference: o1-mini 60.5\%. \textbf{(b) First-pass rate (Pass@1) by difficulty tier.} Broadcast's first-pass advantage over PER and Single-Agent opens at harder tiers. Gray bands mark small-$N$ tiers (tiers 3 and 10, $n\le20$). Figure~\ref{fig:supp_cost_breakeven} extends this raw view to the scalarized $K_p(\lambda)$ model where verifier-call cost $\lambda$ is varied explicitly.}
\label{fig:cost_performance_frontier}
\end{figure}

\begin{figure}[h]
\centering
\includegraphics[width=\linewidth]{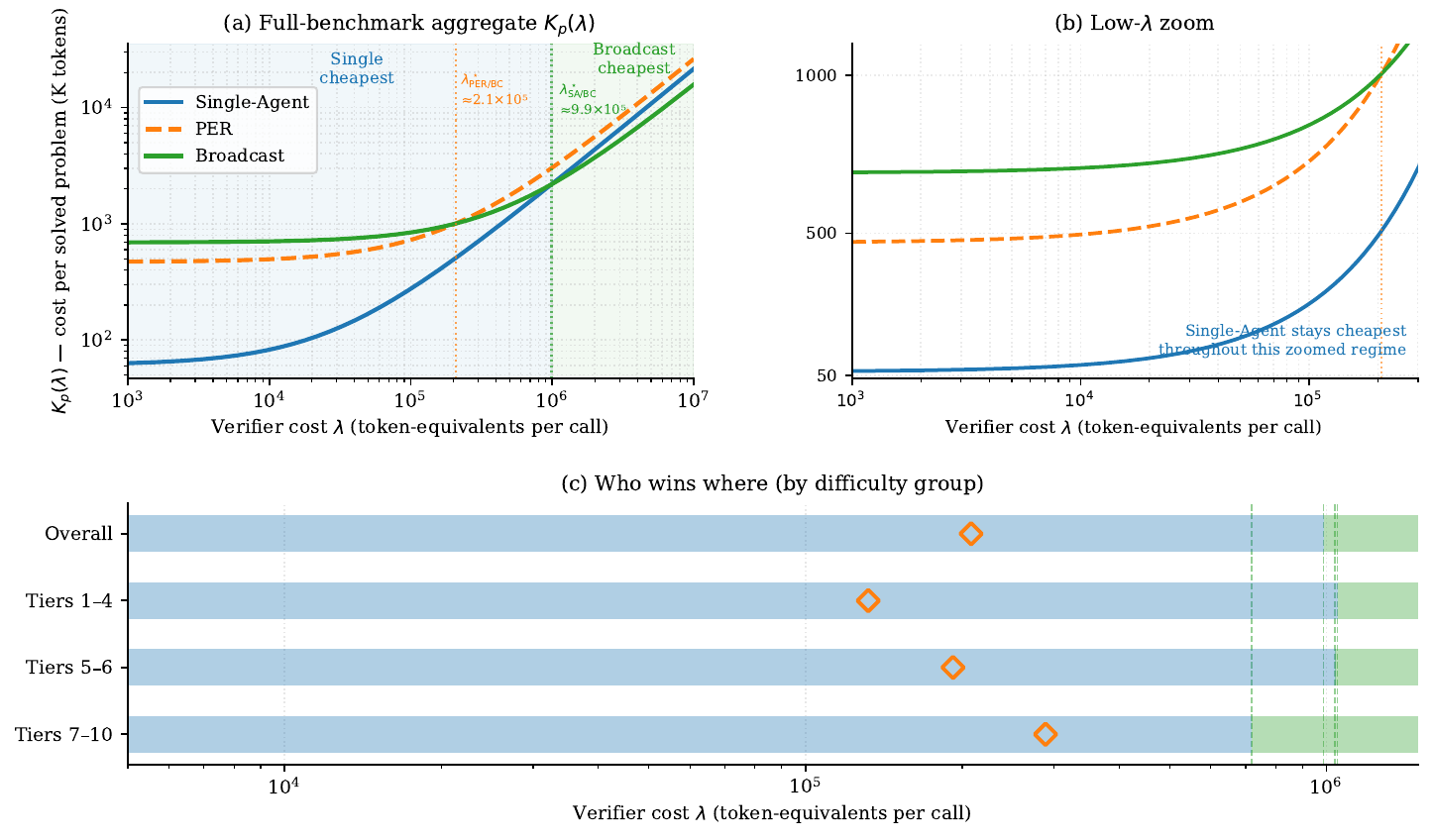}
\caption{Verifier-cost crossover analysis under the scalarized objective $K_p(\lambda)=(T_p+\lambda V_p)/A_p$. \textbf{Panel~(a):} full-benchmark curves for all three protocols. Broadcast overtakes PER when one verifier call costs about $2.1\times10^5$ token-equivalent units, and overtakes Single-Agent at about $9.9\times10^5$. Under this objective, PER never becomes cheaper than Single-Agent. \textbf{Panel~(b):} separate low-verifier-cost zoom, showing more clearly the regime in which Single-Agent remains cheapest. \textbf{Panel~(c):} difficulty-conditioned winner strip chart, with one row for the full benchmark and one for each tier group. Blue means Single-Agent is cheapest; green means Broadcast is cheapest; the orange diamond marks the PER/Broadcast pairwise crossover, not the actual winner switch. The main takeaway is that the Single-Agent/Broadcast boundary moves leftward on the hard tiers 7--10, making Broadcast defensible at a lower verifier cost for harder problems.}
\label{fig:supp_cost_breakeven}
\end{figure}

\begin{figure}[h]
\centering
\includegraphics[width=0.85\linewidth]{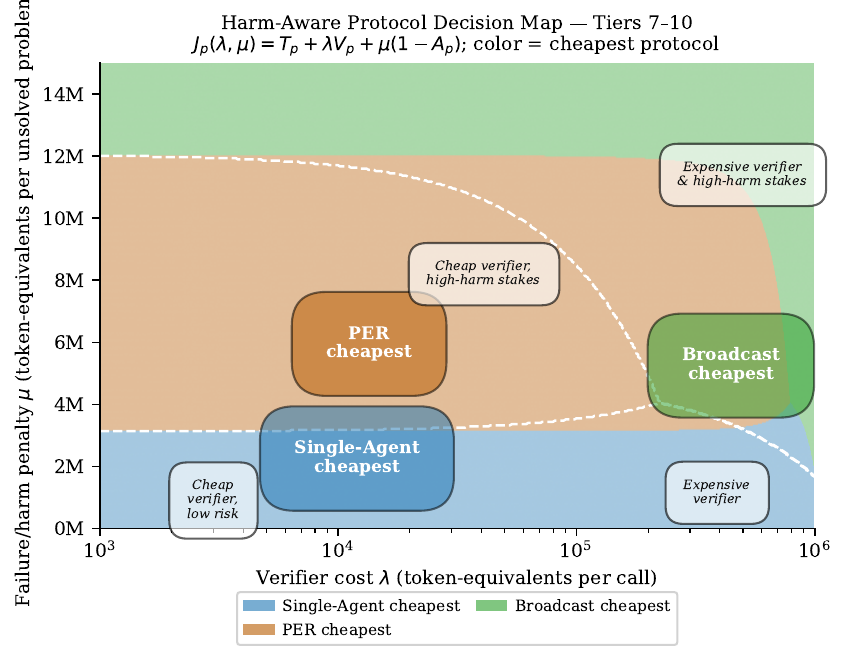}
\caption{Harm-aware protocol decision map for tiers 7--10 under the richer scalarized objective $J_p(\lambda,\mu)=T_p+\lambda V_p+\mu(1-A_p)$, where $\lambda$ is verifier cost and $\mu$ is the per-problem failure penalty. Color shows which protocol minimizes $J_p$. On the hard-tier aggregate (tiers 7--10), PER has a visible preference region at moderate $\mu$ and low-to-moderate $\lambda$, meaning it can be the best compromise when mistakes carry substantial downstream cost but verifier calls are not yet extremely expensive. Broadcast dominates at high $\lambda$ or very high $\mu$; Single-Agent dominates at low $\lambda$ and low $\mu$. On the full-benchmark aggregate, the PER preference region collapses to near-empty; the hard-tier conditioning shown here is required to expose the PER-as-compromise scenario.}
\label{fig:supp_cost_harmaware_map}
\end{figure}

\begin{figure}[h]
\centering
\includegraphics[width=\linewidth]{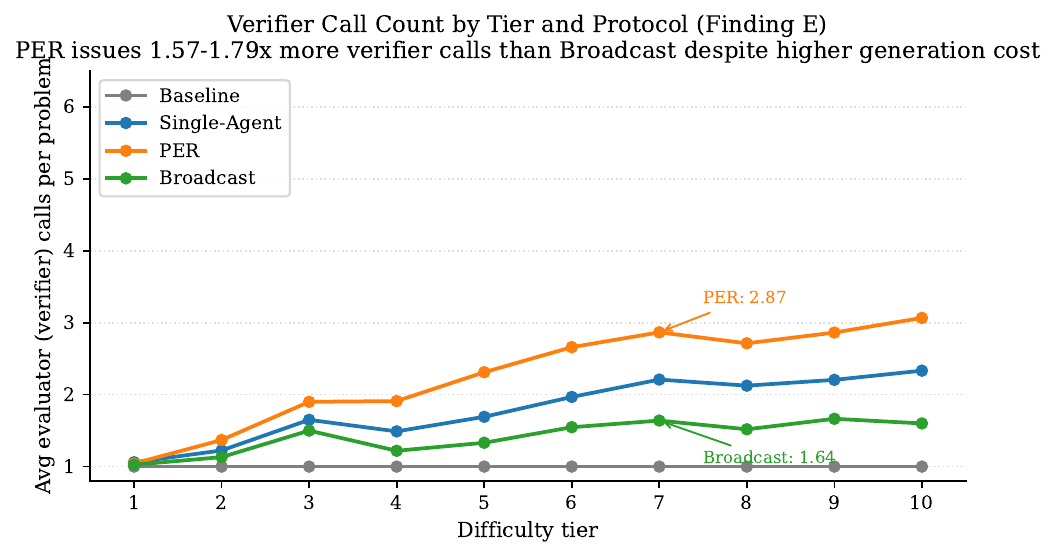}
\caption{Average evaluator (verifier) calls per problem by difficulty tier and protocol. Across tiers 4--9, PER issues about 1.6 to 1.8 times as many verifier calls as Broadcast.}
\label{fig:supp_eval_calls}
\end{figure}

\begin{figure}[h]
\centering
\includegraphics[width=\linewidth]{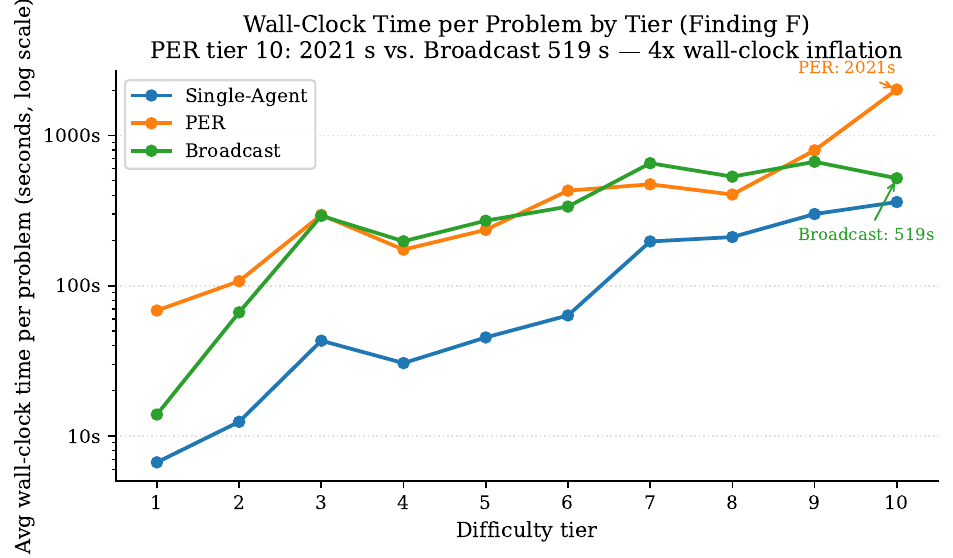}
\caption{Wall-clock time per problem by difficulty tier and protocol. At tier 10, PER averages 2{,}021 seconds per problem versus 519 seconds for Broadcast, or about four times as long. The gap is driven by PER's higher evaluator-call frequency and sequential reviewer--solver turns.}
\label{fig:supp_wall_time}
\end{figure}

\begin{figure}[h]
\centering
\includegraphics[width=\linewidth]{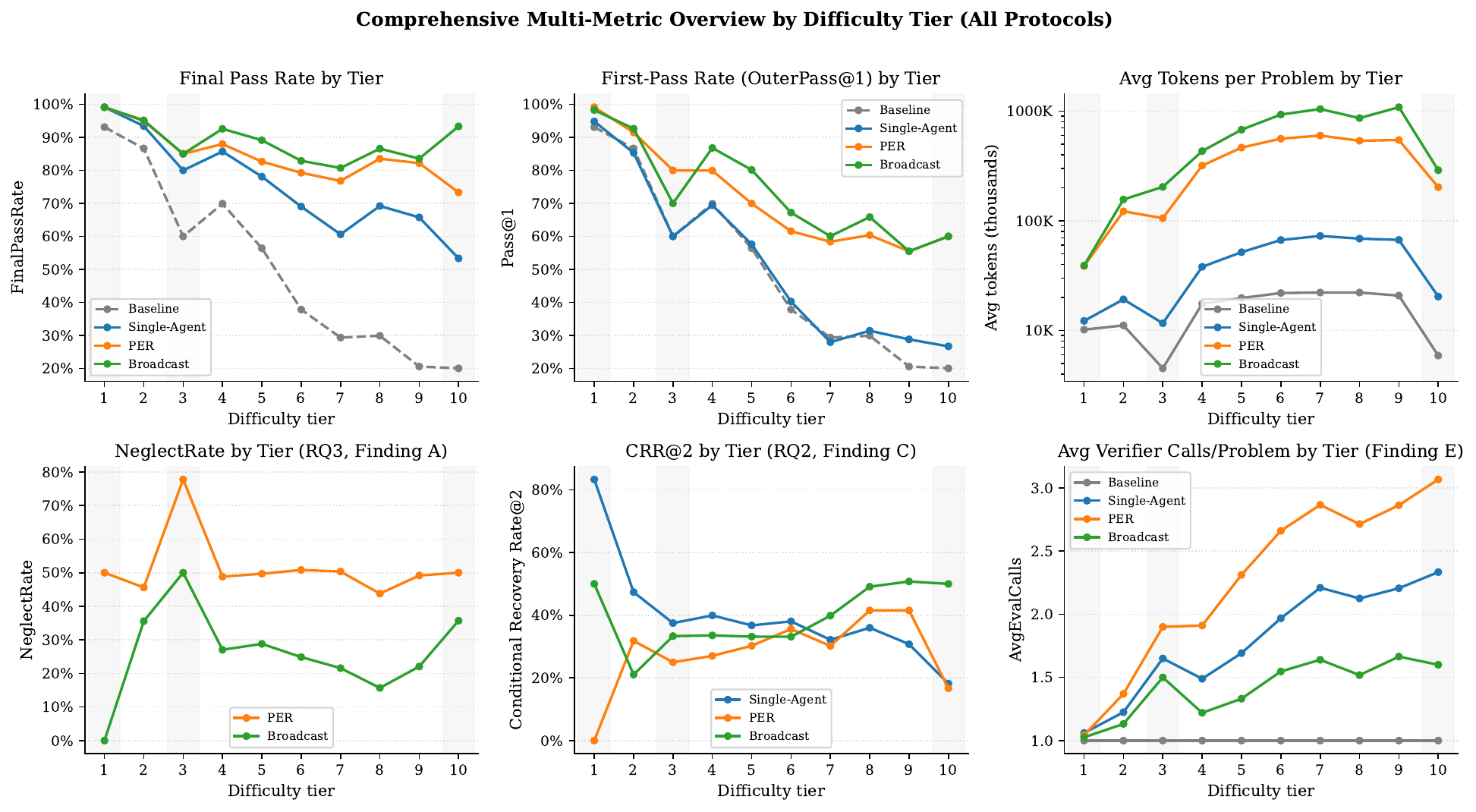}
\caption{Multi-metric overview across all tiers and protocols. The panels show FinalPassRate, Pass@1, pass rate after two outer attempts, pass rate after three outer attempts, average generated tokens, and average evaluator calls per problem. This figure is intended as a quick cross-reference for the paper's four research questions.}
\label{fig:supp_rq_overview}
\end{figure}

\begin{itemize}
\item Full per-tier AvgEvalCalls/problem showing PER issues 1.57--1.79$\times$ more verifier calls than broadcast.
\item Wall-clock time per problem by tier: PER tier 9 = 795\,s, broadcast = 668\,s; PER tier 10 = 2{,}021\,s, broadcast = 519\,s ($4\times$ inflation).
\item Verifier-cost breakeven analysis: the full-benchmark scalarized crossover occurs at roughly $2.1\times10^5$ token-equivalent units per verifier call for PER vs.\ broadcast, and at roughly $9.9\times10^5$ for single-agent vs.\ broadcast (Table~\ref{tab:verifier_cost_thresholds}).
\item Table~\ref{tab:tier_group_cost_thresholds} adds tier-group versions of the same crossover, showing that the verifier-cost decision boundary moves with difficulty instead of staying fixed at the global aggregate.
\item Figure~\ref{fig:cost_performance_frontier} shows the raw cost--performance frontier and first-pass rate by tier; Figure~\ref{fig:supp_cost_breakeven} extends that view to the scalarized $K_p(\lambda)$ crossover curves with a difficulty-conditioned regime-strip chart.
\item Figure~\ref{fig:supp_eval_calls} shows per-tier AvgEvalCalls/problem; Figure~\ref{fig:supp_wall_time} shows wall-clock time per problem by tier.
\item Figure~\ref{fig:supp_cost_harmaware_map} shows the harm-aware protocol decision map under $J_p(\lambda,\mu)$ for tiers 7--10, illustrating when a failure-penalty dimension makes PER a viable compromise.
\item Figure~\ref{fig:supp_rq_overview} provides the full multi-metric overview.
\end{itemize}

\section{Headline Outcome Confidence Intervals}
\label{app:headline_ci}

\begin{table}[h]
\centering
\small
\caption{Overall outcome uncertainty over $N=4{,}181$ problems. FinalPassRate and Pass@1 intervals are 95\% Wilson intervals over per-problem success indicators. The Broadcast--PER gap intervals are 95\% Newcombe intervals for the difference in independent proportions. These intervals capture per-problem sampling uncertainty under the fixed paper configs; they do not substitute for multi-seed variance across different stochastic runs.}
\label{tab:headline_outcome_ci}
\begin{tabular}{lcc}
\toprule
Protocol & FinalPassRate (95\% CI) & Pass@1 (95\% CI) \\
\midrule
Baseline LLM (one-shot) & 56.8\% [55.2, 58.3] & 56.8\% [55.2, 58.3] \\
Single-Agent Iterative & 78.8\% [77.5, 80.0] & 57.4\% [55.9, 58.9] \\
PER (role-based MAS) & 85.2\% [84.1, 86.2] & 72.8\% [71.5, 74.2] \\
Broadcast deliberation & 89.2\% [88.2, 90.1] & 78.6\% [77.3, 79.8] \\
\midrule
Broadcast $-$ PER gap & 4.0 pp [2.6, 5.5] & 5.8 pp [3.9, 7.6] \\
\bottomrule
\end{tabular}
\end{table}

\begin{table}[h]
\centering
\small
\caption{Episode-level uncertainty for the main review-conditioned rates. CouplingRate is the paper-facing immediate answer-change rate on evaluator-verified useful review episodes. Intervals are 95\% Wilson intervals over pooled eligible review episodes; the Broadcast--PER CouplingRate gap uses a 95\% Newcombe interval. Because multiple review episodes can come from the same problem across outer attempts, these should be read as descriptive episode-level uncertainty rather than as fully independent problem-level variance estimates.}
\label{tab:review_conditioned_ci}
\resizebox{\linewidth}{!}{
\begin{tabular}{lccc}
\toprule
Protocol & CouplingRate (95\% CI) & ReviewerGuidedRepairRate (95\% CI) & MisleadingResistance (95\% CI) \\
\midrule
PER (role-based MAS) & 33.6\% [32.3, 34.9] & 5.1\% [4.5, 5.7] & 92.1\% [90.1, 93.7] \\
Broadcast deliberation & 93.5\% [92.8, 94.1] & 28.6\% [27.4, 29.9] & 70.8\% [69.1, 72.4] \\
\midrule
Broadcast $-$ PER Coupling gap & 59.9 pp [57.9, 61.8] & --- & --- \\
\bottomrule
\end{tabular}
}
\end{table}

\begin{table}[h]
\centering
\small
\caption{Paired problem-clustered bootstrap uncertainty for the main PER--broadcast gaps. Each bootstrap replicate resamples the shared problem IDs with replacement and recomputes all associated per-problem outcomes or review-conditioned episodes jointly, preserving within-problem clustering. Intervals are 95\% percentile intervals over 500 bootstrap replicates on the matched 4{,}181-problem set.}
\label{tab:clustered_bootstrap_ci}
\resizebox{\linewidth}{!}{
\begin{tabular}{lccc}
\toprule
Metric & PER (clustered 95\% CI) & Broadcast (clustered 95\% CI) & Broadcast $-$ PER gap (clustered 95\% CI) \\
\midrule
FinalPassRate & 85.2\% [84.0, 86.2] & 89.2\% [88.3, 90.1] & +4.0 pp [3.1, 4.9] \\
Pass@1 & 72.8\% [71.6, 74.0] & 78.6\% [77.5, 79.8] & +5.8 pp [4.7, 7.0] \\
CouplingRate & 33.6\% [32.2, 35.0] & 93.5\% [92.9, 94.1] & +59.9 pp [58.2, 61.4] \\
ReviewerGuidedRepairRate & 5.1\% [4.4, 5.7] & 28.6\% [26.7, 30.7] & +23.5 pp [21.7, 25.5] \\
Strict CouplingRate & 33.6\% [32.4, 35.1] & 93.5\% [92.9, 94.1] & +59.9 pp [58.2, 61.4] \\
RepairGivenChangeRate & 15.1\% [13.5, 17.2] & 30.6\% [28.7, 32.6] & +15.4 pp [13.0, 18.2] \\
MisleadingResistance & 92.1\% [89.8, 94.0] & 70.8\% [68.4, 73.0] & $-21.3$ pp [$-24.0$, $-18.6$] \\
MisleadingReviewSusceptibility & 19.9\% [17.2, 22.7] & 64.8\% [63.0, 66.6] & +45.0 pp [41.8, 48.3] \\
\bottomrule
\end{tabular}
}
\end{table}

Appendix Table~\ref{tab:strict_coupling_audit} gives the pooled audit counts behind this robustness check. For both PER and broadcast, \texttt{StrictLegacyDisagreementRate} is 0, so the stricter deterministic answer-transition definition agrees exactly with the paper-facing CouplingRate on all eligible useful-revise episodes. We therefore use the strict audit as a measurement-stability check, not as a replacement metric.

\begin{figure}[h]
\centering
\includegraphics[width=0.54\linewidth]{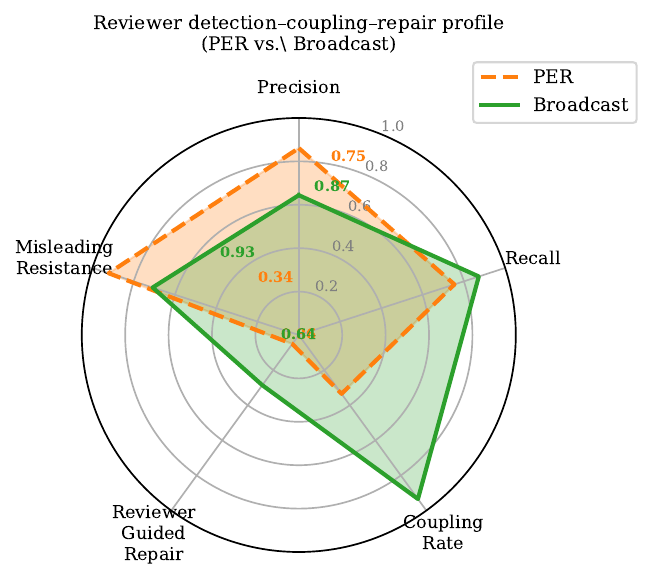}
\caption{Five-dimensional detection--coupling--repair profile for PER (dashed orange) and Broadcast (solid green). Axes show raw values on a 0--1 scale. PER dominates on Precision (0.861 vs.\ 0.644) and MisleadingResistance (0.921 vs.\ 0.708); Broadcast dominates on Recall (0.872 vs.\ 0.754), CouplingRate (0.935 vs.\ 0.336), and ReviewerGuidedRepairRate (0.286 vs.\ 0.051). The shape difference makes the precision--repair gap concrete: the protocols diverge most sharply on the coupling and repair dimensions, not on the detection dimensions alone.}
\label{fig:supp_reviewer_radar}
\end{figure}

\section{Literature Context: Published Omni-MATH Accuracy}
\label{app:literature_context}

\begin{itemize}
\item Published single-model accuracy numbers on Omni-MATH for reference context only. Evaluator setup, prompting, problem subset, and test-time budget are not matched to this paper's experiments.
\item Do not treat these as direct comparisons; they establish benchmark headroom motivation.
\end{itemize}

\begin{table}[h]
\centering
\small
\caption{Published Omni-MATH overall final pass rates \cite{omnimath} alongside this paper's four protocols. Reference rows are context only; experimental conditions are not matched.}
\label{tab:literature_context_appendix}
\begin{tabular}{lc}
\toprule
System & Overall FinalPassRate \\
\midrule
\multicolumn{2}{l}{\textit{Published single-model references \cite{omnimath}}} \\
Meta-Llama-3.1-70B-Instruct & 24.2\% \\
GPT-4o & 30.5\% \\
Qwen2.5-MATH-72B-Instruct & 36.2\% \\
OpenAI o1-preview & 52.6\% \\
OpenAI o1-mini & 60.5\% \\
\midrule
\multicolumn{2}{l}{\textit{This paper (\texttt{gpt-oss-120b}, $N=4{,}181$; baseline plain feedback, iterative/collaborative modes hinted)}} \\
Baseline LLM (one-shot) & 56.8\% \\
Single-Agent Iterative & 78.8\% \\
PER (role-based MAS) & 85.2\% \\
Broadcast deliberation & 89.2\% \\
\bottomrule
\end{tabular}
\end{table}

\section{Trace Examples}
\label{app:trace_examples}

The examples in this section are illustrative mechanism traces, not prevalence estimates. We selected them from the matched paper results directory using pre-specified criteria: we first preferred mid-tier cases (tiers 5--7), then preferred same-question PER/broadcast pairs when available, and then prioritized visible mechanism over dramatic outcome. For PER, the target pattern was substantively useful reviewer signal / solver-or-submission-unrepaired / evaluator-\texttt{FAIL}. For broadcast, the target pattern was visible peer critique followed by an explicit protocol candidate update and re-approval of the revised candidate. We found three same-question tier-6 pairs where PER failed and broadcast passed, but the cleanest reviewer-neglect trace and the cleanest visible candidate-revision trace did not coincide on a single problem, so the two excerpts below come from different mid-tier problems. All frequency claims in the paper remain metric-based.

For readability, the monospaced blocks below preserve stage labels and trace wording while eliding unrelated algebra; any bracketed stage tags and surrounding interpretation are ours.

\subsection{PER: Correct Review Signal, Unrepaired Submission}
\label{app:trace_per_neglect}

This tier-5 PER trace shows the decoupling pattern directly. The reviewer twice identifies a real flaw in the current derivation, and the second review even proposes a materially different threshold, but the submitted answer remains the earlier wrong candidate. This is the mechanism counted by \textsc{NeglectRate}: a substantively useful review signal exists, but it does not become the answer actually sent to the evaluator.

\paragraph{Problem context.} The task asks for the largest initial value $a_0$ such that the recurrence $a_{n+1}=a_n^2-2^{-(2020\cdot 2^n-1)}$ remains bounded. The evaluator label for this run is $1+\frac{1}{2^{2020}}$, while the submitted answer is $\boxed{1}$.

\begin{quote}\small\ttfamily\raggedright
[reviewer\_0] Diagnosis: Lemma~1 incorrectly claims $a_1 \ge 0$ for every $a_0$; in fact $a_1=a_0^2-\varepsilon_0$ can be negative. ... [Route:Executor]\par
[executor\_1] ... Thus the largest real number for which the sequence remains bounded is $\boxed{1}$.\par
[reviewer\_1] Diagnosis: The argument that any initial value greater than $1$ yields an unbounded sequence is incorrect; for $1<a_0\le \sqrt{1+\varepsilon_0}$ we have $a_1\le 1$ and the sequence remains bounded. ... Change final boxed answer to $\boxed{\sqrt{1+2^{-2019}}}$. [Route:Planner]\par
[evaluation\_submission] Actual extracted final answer sent to evaluator: $\boxed{1}$\par
[evaluation] Evaluation signal: FAIL
\end{quote}

The key point is not just that PER fails. The decisive detail is that the reviewer-generated correction remains a separable signal rather than becoming the active submission state. Even after a concrete corrected answer appears in the review channel, the pipeline still submits the earlier candidate. This is exactly the reviewer--solver uncoupling mechanism that the main text isolates.

\subsection{Broadcast: Peer Critique Replaces the Live Candidate}
\label{app:trace_broadcast_repair}

This tier-6 broadcast trace shows the complementary mechanism. The first public answer enters review with an overly large endpoint, peer approval produces a correction proposal, and the system explicitly replaces the current candidate before the next approval round. The corrected candidate is then the object of further scrutiny and is ultimately accepted by the evaluator.

\paragraph{Problem context.} The task asks for all possible final values after repeatedly pairing the numbers $1,\dots,1024$ and replacing each pair by its non-negative difference. The final accepted answer in this run is $\boxed{\{0,2,4,\dots,1022\}}$.

\begin{quote}\small\ttfamily\raggedright
[poll\_0 / Mira] Candidate Answer: $\boxed{\{0,2,4,\dots,524800\}}$\par
[poll\_0 / Talia] Candidate Answer: $\boxed{\{0,2,4,\dots,1022\}}$\par
[discussion\_0 / Mira] Candidate enters review as $\boxed{\{0,2,4,\dots,1024\}}$\par
[approval\_0 / Talia] Proposed Correction: $\boxed{\{0,2,4,\dots,1022\}}$\par
[system] Protocol candidate update. Previous Candidate: $\boxed{\{0,2,4,\dots,1024\}}$. Current Candidate For Further Review: $\boxed{\{0,2,4,\dots,1022\}}$.\par
[approval\_1 / Rowan] approve $\boxed{\{0,2,4,\dots,1022\}}$\par
[approval\_1 / Talia] approve $\boxed{\{0,2,4,\dots,1022\}}$\par
[evaluation] PASS
\end{quote}

This is the coupling pattern that the paper attributes to broadcast. The critique is not merely attached as external advice to a solver role. It becomes shared protocol state: the current candidate itself is replaced, re-reviewed, and then submitted in revised form. Even though the surrounding discussion is short, the repair signal is embedded in the live candidate trajectory rather than being left as a side-channel comment.

\section{Process Diagnostics (Flagged Subsets)}
\label{app:process_diagnostics}

\begin{itemize}
\item The main quantitative metrics are computed over the full evaluated set whenever candidate labels are available. Looping, drift, early convergence, wrong convergence, and productive convergence are diagnostic categories for flagged subsets of traces, not headline metrics.
\item Diagnostic flag criteria and planned reporting role:
  \begin{itemize}
  \item \emph{Looping}: repeated boxed answers, repeated review routes, $>$2 turns with no candidate change.
  \item \emph{Drift}: candidate revisions move away from problem constraints or abandon useful verifier feedback.
  \item \emph{Early convergence}: agreement reached with fewer than 2 distinct candidates; low substantive revision count.
  \item \emph{Wrong convergence}: unanimous approval followed by verifier \texttt{FAIL}.
  \item \emph{Productive convergence}: unanimous approval followed by verifier \texttt{PASS}.
  \end{itemize}
\item Role-specific trace slices: useful objection, error preservation, neglected errors, and redundant repetition.
\item Cross-protocol: compare how interaction structure shapes reasoning trajectory shape (not just final answer), including where structured review produces targeted refinement and where peer deliberation yields useful diversity vs.\ harmful drift \cite{multiagentdebate,masurvey,evalreasonaction}.
\end{itemize}

%% file: checklist.tex
\section*{NeurIPS Paper Checklist}

\begin{enumerate}

\item {\bf Claims}
    \item[] Question: Do the main claims made in the abstract and introduction accurately reflect the paper's contributions and scope?
    \item[] Answer: \answerYes{}
    \item[] Justification: The abstract and introduction state the paper as a mechanism-focused evaluation of reviewer--solver coupling, not as a new SOTA method, and explicitly qualify the claims to matched configs and verifier-grounded math (Abstract; Secs.~1 and~7).

\item {\bf Limitations}
    \item[] Question: Does the paper discuss the limitations of the work performed by the authors?
    \item[] Answer: \answerYes{}
    \item[] Justification: The paper includes a dedicated limitations section covering domain scope, shared-model dependence, prompt/interface scope, token cost, and trace-parsed diagnostics (Sec.~7).

\item {\bf Theory assumptions and proofs}
    \item[] Question: For each theoretical result, does the paper provide the full set of assumptions and a complete (and correct) proof?
    \item[] Answer: \answerNA{}
    \item[] Justification: The paper is empirical and does not claim formal theorems or proofs.

    \item {\bf Experimental result reproducibility}
    \item[] Question: Does the paper fully disclose all the information needed to reproduce the main experimental results of the paper to the extent that it affects the main claims and/or conclusions of the paper (regardless of whether the code and data are provided or not)?
    \item[] Answer: \answerYes{}
    \item[] Justification: We disclose the public benchmark, matched protocol structure, evaluator contract, budgets, temperature-$0$ paper configs, prompt-surface/output contracts, routing and memory details, and concrete reproduction paths. The paper includes a public release website; code and data will be released after institutional approval. The actor/evaluator model \texttt{gpt-oss-120b} is public and open-weight rather than a private API-only dependency, and the reported runs host those public weights on GPU-backed HPC infrastructure through an OpenAI-compatible server (Secs.~3--4; App.~\ref{app:evaluator}, App.~\ref{app:protocol_configs}, and App.~\ref{app:artifact_release}).

\item {\bf Open access to data and code}
    \item[] Question: Does the paper provide open access to the data and code, with sufficient instructions to faithfully reproduce the main experimental results, as described in supplemental material?
    \item[] Answer: \answerNo{}
    \item[] Justification: The benchmark inputs are public, and the paper links the public release website. Code and data will be released after institutional approval. The appendix lists the planned artifact contents and relative paths for the staged release package (App.~\ref{app:artifact_release} and App.~\ref{app:artifact_paths}).
    \item[] Guidelines:
    \begin{itemize}
        \item The answer \answerNA{} means that paper does not include experiments requiring code.
        \item Please see the NeurIPS code and data submission guidelines (\url{https://neurips.cc/public/guides/CodeSubmissionPolicy}) for more details.
        \item While we encourage the release of code and data, we understand that this might not be possible, so \answerNo{} is an acceptable answer. Papers cannot be rejected simply for not including code, unless this is central to the contribution (e.g., for a new open-source benchmark).
        \item The instructions should contain the exact command and environment needed to run to reproduce the results. See the NeurIPS code and data submission guidelines (\url{https://neurips.cc/public/guides/CodeSubmissionPolicy}) for more details.
        \item The authors should provide instructions on data access and preparation, including how to access the raw data, preprocessed data, intermediate data, and generated data, etc.
        \item The authors should provide scripts to reproduce all experimental results for the new proposed method and baselines. If only a subset of experiments are reproducible, they should state which ones are omitted from the script and why.
        \item For public release, the authors should release institutionally approved public versions when applicable.
        \item Providing as much information as possible in supplemental material (appended to the paper) is recommended, but including URLs to data and code is permitted.
    \end{itemize}

\item {\bf Experimental setting/details}
    \item[] Question: Does the paper specify all the training and test details (e.g., data splits, hyperparameters, how they were chosen, type of optimizer) necessary to understand the results?
    \item[] Answer: \answerYes{}
    \item[] Justification: The paper and appendix specify the benchmark, matched protocol ladder, outer/inner-loop budgets, evaluator contract, model family, temperature, and protocol-specific memory/routing details needed to understand the results (Secs.~3--5; App.~\ref{app:evaluator} and App.~\ref{app:memory_contracts}).

\item {\bf Experiment statistical significance}
    \item[] Question: Does the paper report error bars suitably and correctly defined or other appropriate information about the statistical significance of the experiments?
    \item[] Answer: \answerYes{}
    \item[] Justification: The paper now reports headline 95\% confidence intervals for FinalPassRate and Pass@1, including a Newcombe interval for the Broadcast--PER gap, and the appendix reports Wilson intervals for tier-wise NeglectRate and related diagnostics together with the assumptions behind those intervals (Sec.~5; App.~\ref{app:headline_ci} and App.~\ref{app:neglect_by_tier}).

\item {\bf Experiments compute resources}
    \item[] Question: For each experiment, does the paper provide sufficient information on the computer resources (type of compute workers, memory, time of execution) needed to reproduce the experiments?
    \item[] Answer: \answerYes{}
    \item[] Justification: The appendix now records the concrete compute path: public \texttt{gpt-oss-120b} weights served through a public OpenAI-compatible inference endpoint on GPU-backed HPC infrastructure, standard GPU-backed orchestration nodes used only for benchmark coordination, worker parallelism by protocol, and representative first-pass wall-clock times, while the main paper reports protocol-level token, evaluator-call, and hard-tier wall-clock costs (Sec.~5; App.~\ref{app:artifact_release} and App.~\ref{app:memory_contracts}).
    
\item {\bf Code of ethics}
    \item[] Question: Does the research conducted in the paper conform, in every respect, with the NeurIPS Code of Ethics \url{https://neurips.cc/public/EthicsGuidelines}?
    \item[] Answer: \answerYes{}
    \item[] Justification: The work is a benchmark-based protocol evaluation with no human subjects, no deceptive data collection, and an explicit discussion of limitations and broader impacts (Sec.~7).
    \item[] Guidelines:
    \begin{itemize}
        \item The answer \answerNA{} means that the authors have not reviewed the NeurIPS Code of Ethics.
        \item If the authors answer \answerNo, they should explain the special circumstances that require a deviation from the Code of Ethics.
        \item The authors should make sure institutional, legal, and release-review requirements are satisfied.
    \end{itemize}

\item {\bf Broader impacts}
    \item[] Question: Does the paper discuss both potential positive societal impacts and negative societal impacts of the work performed?
    \item[] Answer: \answerYes{}
    \item[] Justification: The paper includes a broader-impact discussion covering both the positive value of coupling-aware evaluation and the potential capability gains from stronger feedback uptake (Sec.~7).
    
\item {\bf Safeguards}
    \item[] Question: Does the paper describe safeguards that have been put in place for responsible release of data or models that have a high risk for misuse (e.g., pre-trained language models, image generators, or scraped datasets)?
    \item[] Answer: \answerNA{}
    \item[] Justification: The paper does not release a new model, new scraped dataset, or other high-misuse asset.

\item {\bf Licenses for existing assets}
    \item[] Question: Are the creators or original owners of assets (e.g., code, data, models), used in the paper, properly credited and are the license and terms of use explicitly mentioned and properly respected?
    \item[] Answer: \answerYes{}
    \item[] Justification: The paper and appendix explicitly credit and document the external assets used in the experiments: \texttt{gpt-oss-120b} is identified as a public OpenAI checkpoint released under Apache 2.0 with the OpenAI usage policy; the public Omni-MATH / Omni-MATH-2 dataset cards used for benchmarking are cited and noted as MIT-licensed; and the auxiliary public models used only for evaluator replay and cross-family robustness checks (\texttt{Meta-Llama-3.1-70B-Instruct} and \texttt{gemma-3-27b-it}) are also credited and used subject to their respective model licenses and usage terms. The benchmark papers and repositories are credited directly (Sec.~3; App.~\ref{app:artifact_release} and App.~\ref{app:evaluator}).

\item {\bf New assets}
    \item[] Question: Are new assets introduced in the paper well documented and is the documentation provided alongside the assets?
    \item[] Answer: \answerYes{}
    \item[] Justification: The staged reproducibility artifact is planned to contain the exact paper configs, prompt-bearing YAMLs, compact per-problem tables, analysis scripts, packaged report tables, and reproduction instructions. It does not introduce a new model or dataset, but the planned release artifact is documented alongside the paper (App.~\ref{app:artifact_release}).
    \item[] Guidelines:
    \begin{itemize}
        \item The answer \answerNA{} means that the paper does not release new assets.
        \item Researchers should communicate the details of the dataset\slash code\slash model as part of their submissions via structured templates. This includes details about training, license, limitations, etc. 
        \item The paper should discuss whether and how consent was obtained from people whose asset is used.
        \item For public release, remember to use approved public assets when applicable.
    \end{itemize}

\item {\bf Crowdsourcing and research with human subjects}
    \item[] Question: For crowdsourcing experiments and research with human subjects, does the paper include the full text of instructions given to participants and screenshots, if applicable, as well as details about compensation (if any)? 
    \item[] Answer: \answerNA{}
    \item[] Justification: The paper does not involve crowdsourcing or research with human subjects.

\item {\bf Institutional review board (IRB) approvals or equivalent for research with human subjects}
    \item[] Question: Does the paper describe potential risks incurred by study participants, whether such risks were disclosed to the subjects, and whether Institutional Review Board (IRB) approvals (or an equivalent approval/review based on the requirements of your country or institution) were obtained?
    \item[] Answer: \answerNA{}
    \item[] Justification: The paper does not involve crowdsourcing or research with human subjects.
    \item[] Guidelines:
    \begin{itemize}
        \item The answer \answerNA{} means that the paper does not involve crowdsourcing nor research with human subjects.
        \item Depending on the country in which research is conducted, IRB approval (or equivalent) may be required for any human subjects research. If you obtained IRB approval, you should clearly state this in the paper. 
        \item We recognize that the procedures for this may vary significantly between institutions and locations, and we expect authors to adhere to the NeurIPS Code of Ethics and the guidelines for their institution. 
        \item For public release, include institutional information only after the relevant review requirements are satisfied.
    \end{itemize}

\item {\bf Declaration of LLM usage}
    \item[] Question: Does the paper describe the usage of LLMs if it is an important, original, or non-standard component of the core methods in this research? Note that if the LLM is used only for writing, editing, or formatting purposes and does \emph{not} impact the core methodology, scientific rigor, or originality of the research, declaration is not required.
    \item[] Answer: \answerYes{}
    \item[] Justification: LLMs are the core acting agents and evaluator in the studied protocols, and the paper describes their role, model family, feedback contract, and protocol usage throughout the main text and appendix (Secs.~3--5; App.~\ref{app:evaluator} and App.~\ref{app:memory_contracts}).

\end{enumerate}

%% file: references.bib
@article{multiagentdebate,
  title = {Improving Factuality and Reasoning in Language Models through Multiagent Debate},
  author = {Du, Yilun and Li, Shuang and Torralba, Antonio and Tenenbaum, Joshua B. and Mordatch, Igor},
  journal = {arXiv preprint arXiv:2305.14325},
  year = {2023},
  url = {https://arxiv.org/abs/2305.14325}
}

@article{masurvey,
  title = {Multi-Agent Collaboration Mechanisms: A Survey of LLMs},
  author = {Tran, Khanh-Tung and Dao, Dung and Nguyen, Minh-Duong and Pham, Quoc-Viet and O'Sullivan, Barry and Nguyen, Hoang D.},
  journal = {arXiv preprint arXiv:2501.06322},
  year = {2025},
  url = {https://arxiv.org/abs/2501.06322}
}

@inproceedings{wei2022chainofthought,
  title = {Chain-of-Thought Prompting Elicits Reasoning in Large Language Models},
  author = {Wei, Jason and Wang, Xuezhi and Schuurmans, Dale and Bosma, Maarten and Ichter, Brian and Xia, Fei and Chi, Ed and Le, Quoc V. and Zhou, Denny},
  booktitle = {Advances in Neural Information Processing Systems},
  year = {2022},
  url = {https://arxiv.org/abs/2201.11903}
}

@inproceedings{wang2023selfconsistency,
  title = {Self-Consistency Improves Chain of Thought Reasoning in Language Models},
  author = {Wang, Xuezhi and Wei, Jason and Schuurmans, Dale and Le, Quoc V. and Chi, Ed H. and Narang, Sharan and Chowdhery, Aakanksha and Zhou, Denny},
  booktitle = {International Conference on Learning Representations},
  year = {2023},
  url = {https://arxiv.org/abs/2203.11171}
}

@inproceedings{yao2023treeofthoughts,
  title = {Tree of Thoughts: Deliberate Problem Solving with Large Language Models},
  author = {Yao, Shunyu and Yu, Dian and Zhao, Jeffrey and Shafran, Izhak and Griffiths, Thomas L. and Cao, Yuan and Narasimhan, Karthik},
  booktitle = {Advances in Neural Information Processing Systems},
  year = {2023},
  url = {https://arxiv.org/abs/2305.10601}
}

@article{autogen,
  title = {AutoGen: Enabling Next-Gen {LLM} Applications via Multi-Agent Conversation},
  author = {Wu, Qingyun and Bansal, Gagan and Zhang, Jieyu and Wu, Yiran and Li, Beibin and Zhu, Erkang and Jiang, Li and Zhang, Xiaoyun and Zhang, Shaokun and Liu, Jiale and Awadallah, Ahmed Hassan and White, Ryen W. and Burger, Doug and Wang, Chi},
  journal = {arXiv preprint arXiv:2308.08155},
  year = {2023},
  url = {https://arxiv.org/abs/2308.08155}
}

@article{camel,
  title = {{CAMEL}: Communicative Agents for Mind Exploration of Large Scale Language Model Society},
  author = {Li, Guohao and Hammoud, Hasan Abed Al Kader and Itani, Hani and Khizbullin, Dmitrii and Ghanem, Bernard},
  journal = {arXiv preprint arXiv:2303.17760},
  year = {2023},
  url = {https://arxiv.org/abs/2303.17760}
}

@inproceedings{agentbench,
  title = {AgentBench: Evaluating LLMs as Agents},
  author = {Liu, Xiao and Yu, Hao and Zhang, Hanchen and Xu, Yifan and Lei, Xuanyu and Lai, Hanyu and Gu, Yu and Ding, Hangliang and Men, Kai and Yang, Kejuan and Zhang, Shudan and Deng, Xiang and Zeng, Aohan and Du, Zhengxiao and Zhang, Chenhui and Shen, Sheng and Zhang, Tianjun and Su, Yu and Sun, Huan and Huang, Minlie and Dong, Yuxiao and Tang, Jie},
  booktitle = {International Conference on Learning Representations},
  year = {2024},
  url = {https://arxiv.org/abs/2308.03688}
}

@article{aiagentsmatter,
  title = {AI Agents That Matter},
  author = {Kapoor, Sayash and Stroebl, Benedikt and Siegel, Zachary S. and Nadgir, Nitya and Narayanan, Arvind},
  journal = {arXiv preprint arXiv:2407.01502},
  year = {2024},
  url = {https://arxiv.org/abs/2407.01502}
}

@article{zhang2025ttssurvey,
  title = {A Survey on Test-Time Scaling in Large Language Models: What, How, Where, and How Well?},
  author = {Zhang, Qiyuan and Lyu, Fuyuan and Sun, Zexu and Wang, Lei and Zhang, Weixu and Hua, Wenyue and Wu, Haolun and Guo, Zhihan and Wang, Yufei and Muennighoff, Niklas and King, Irwin and Liu, Xue},
  journal = {arXiv preprint arXiv:2503.24235},
  year = {2025},
  url = {https://arxiv.org/abs/2503.24235}
}

@article{yehudai2025agentsurvey,
  title = {Survey on Evaluation of LLM-based Agents},
  author = {Yehudai, Asaf and Eden, Lilach and Li, Alan and Uziel, Guy and Zhao, Yilun and Bar-Haim, Roy and Cohan, Arman and Shmueli-Scheuer, Michal},
  journal = {arXiv preprint arXiv:2503.16416},
  year = {2025},
  url = {https://arxiv.org/abs/2503.16416}
}

@inproceedings{selfrefine,
  title = {Self-Refine: Iterative Refinement with Self-Feedback},
  author = {Madaan, Aman and Tandon, Niket and Gupta, Prakhar and Hallinan, Skyler and Gao, Luyu and Wiegreffe, Sarah and Alon, Uri and Dziri, Nouha and Prabhumoye, Shrimai and Yang, Yiming and Gupta, Shashank and Majumder, Bodhisattwa Prasad and Hermann, Katherine and Welleck, Sean and Yazdanbakhsh, Amir and Clark, Peter},
  booktitle = {Advances in Neural Information Processing Systems},
  year = {2023},
  url = {https://arxiv.org/abs/2303.17651}
}

@inproceedings{reflexion,
  title = {Reflexion: Language Agents with Verbal Reinforcement Learning},
  author = {Shinn, Noah and Cassano, Federico and Berman, Edward and Gopinath, Ashwin and Narasimhan, Karthik and Yao, Shunyu},
  booktitle = {Advances in Neural Information Processing Systems},
  year = {2023},
  url = {https://arxiv.org/abs/2303.11366}
}

@article{selfcorrection,
  title = {Confidence v.s. Critique: A Decomposition of Self-Correction Capability for LLMs},
  author = {Yang, Zhe and Zhang, Yichang and Wang, Yudong and Xu, Ziyao and Lin, Junyang and Sui, Zhifang},
  journal = {arXiv preprint arXiv:2412.19513},
  year = {2024},
  url = {https://arxiv.org/abs/2412.19513}
}

@article{cannotselfcorrect,
  title = {Large Language Models Cannot Self-Correct Reasoning Yet},
  author = {Huang, Jie and Chen, Xinyun and Mishra, Swaroop and Zheng, Huaixiu Steven and Yu, Adams Wei and Song, Xinying and Zhou, Denny},
  journal = {arXiv preprint arXiv:2310.01798},
  year = {2024},
  url = {https://arxiv.org/abs/2310.01798}
}

@article{evalreasonaction,
  title = {On Evaluating the Integration of Reasoning and Action in {LLM} Agents with Database Question Answering},
  author = {Nan, Linyong and Zhang, Ellen and Zou, Weijin and Zhao, Yilun and Zhou, Wenfei and Cohan, Arman},
  journal = {arXiv preprint arXiv:2311.09721},
  year = {2023},
  url = {https://arxiv.org/abs/2311.09721}
}

@article{xverify,
  title = {xVerify: Efficient Answer Verifier for Reasoning Model Evaluations},
  author = {Chen, Ding and Yu, Qingchen and Wang, Pengyuan and Hu, Mengting and Zhang, Wentao and Wang, Zhengren and Tang, Bo and Xiong, Feiyu and Li, Xinchi and Wang, Chao and Yang, Minchuan and Li, Zhiyu},
  journal = {arXiv preprint arXiv:2504.10481},
  year = {2025},
  url = {https://arxiv.org/abs/2504.10481}
}

@article{jetts,
  title = {Evaluating Judges as Evaluators: The JETTS Benchmark of LLM-as-Judges as Test-Time Scaling Evaluators},
  author = {Zhou, Yilun and Xu, Austin and Wang, Peifeng and Xiong, Caiming and Joty, Shafiq},
  journal = {arXiv preprint arXiv:2504.15253},
  year = {2025},
  url = {https://arxiv.org/abs/2504.15253}
}

@inproceedings{strongverifiers,
  title = {Small Language Models Need Strong Verifiers to Self-Correct Reasoning},
  author = {Zhang, Yunxiang and Khalifa, Muhammad and Logeswaran, Lajanugen and Kim, Jaekyeom and Lee, Moontae and Lee, Honglak and Wang, Lu},
  booktitle = {Findings of the Association for Computational Linguistics: ACL 2024},
  year = {2024},
  url = {https://aclanthology.org/2024.findings-acl.924/}
}

@article{scalableoversight,
  title = {On Scalable Oversight with Weak LLMs Judging Strong LLMs},
  author = {Kenton, Zachary and Siegel, Noah Y. and Kramar, Janos and Brown-Cohen, Jonah and Albanie, Samuel and Bulian, Jannis and Agarwal, Rishabh and Lindner, David and Tang, Yunhao and Goodman, Noah D. and Shah, Rohin},
  journal = {arXiv preprint arXiv:2407.04622},
  year = {2024},
  url = {https://arxiv.org/abs/2407.04622}
}

@article{omnimath,
  title = {Omni-MATH: A Universal Olympiad Level Mathematic Benchmark For Large Language Models},
  author = {Gao, Bofei and Song, Feifan and Yang, Zhe and Cai, Zefan and Miao, Yibo and Dong, Qingxiu and Li, Lei and Ma, Chenghao and Chen, Liang and Xu, Runxin and Tang, Zhengyang and Wang, Benyou and Zan, Daoguang and Quan, Shanghaoran and Zhang, Ge and Sha, Lei and Zhang, Yichang and Ren, Xuancheng and Liu, Tianyu and Chang, Baobao},
  journal = {arXiv preprint arXiv:2410.07985},
  year = {2024},
  url = {https://arxiv.org/abs/2410.07985}
}

@article{omnimath2,
  title = {Benchmarks Saturate When The Model Gets Smarter Than The Judge},
  author = {Ballon, Marthe and Algaba, Andres and Verbeken, Brecht and Ginis, Vincent},
  journal = {arXiv preprint arXiv:2601.19532},
  year = {2026},
  url = {https://arxiv.org/abs/2601.19532}
}

@inproceedings{mrr,
  title = {The {TREC}-8 Question Answering Track Report},
  author = {Voorhees, Ellen M.},
  booktitle = {Proceedings of the Eighth Text REtrieval Conference ({TREC}-8)},
  year = {1999},
  url = {https://trec.nist.gov/pubs/trec8/papers/qa_report.pdf}
}

@article{malt,
  title = {MALT: Improving Reasoning with Multi-Agent {LLM} Training},
  author = {Motwani, Sumeet Ramesh and Smith, Chandler and Das, Rocktim Jyoti and Rafailov, Rafael and Laptev, Ivan and Torr, Philip H. S. and Pizzati, Fabio and Clark, Ronald and de Witt, Christian Schroeder},
  journal = {arXiv preprint arXiv:2412.01928},
  year = {2024},
  url = {https://arxiv.org/abs/2412.01928}
}

@article{dpsdp,
  title = {Reinforce {LLM} Reasoning through Multi-Agent Reflection},
  author = {Yuan, Yurun and Xie, Tengyang},
  journal = {arXiv preprint arXiv:2506.08379},
  year = {2025},
  url = {https://arxiv.org/abs/2506.08379}
}

@article{canagentsdebate,
  title = {Can {LLM} Agents Really Debate? A Controlled Study of Multi-Agent Debate in Logical Reasoning},
  author = {Wu, Haolun and Li, Zhenkun and Li, Lingyao},
  journal = {arXiv preprint arXiv:2511.07784},
  year = {2025},
  url = {https://arxiv.org/abs/2511.07784}
}

@article{whymasfail,
  title = {Why Do Multi-Agent {LLM} Systems Fail?},
  author = {Cemri, Mert and Pan, Melissa Z. and Yang, Shuyi and Agrawal, Lakshya A. and Chopra, Bhavya and Tiwari, Rishabh and Keutzer, Kurt and Parameswaran, Aditya and Klein, Dan and Ramchandran, Kannan and Zaharia, Matei and Gonzalez, Joseph E.},
  journal = {arXiv preprint arXiv:2503.13657},
  year = {2025},
  url = {https://arxiv.org/abs/2503.13657}
}

@article{correctbench,
  title = {Can LLMs Correct Themselves? A Benchmark of Self-Correction in LLMs},
  author = {Tie, Guiyao and Yuan, Zenghui and Zhao, Zeli and Hu, Chaoran and Gu, Tianhe and Zhang, Ruihang and Zhang, Sizhe and Wu, Junran and Tu, Xiaoyue and Jin, Ming and Wen, Qingsong and Chen, Lixing},
  journal = {arXiv preprint arXiv:2510.16062},
  year = {2025},
  url = {https://arxiv.org/abs/2510.16062}
}

@article{selfcorrectionblindspot,
  title = {Self-Correction Bench: Uncovering and Addressing the Self-Correction Blind Spot in Large Language Models},
  author = {Tsui, Ken},
  journal = {OpenReview},
  year = {2025},
  url = {https://openreview.net/forum?id=7K1kXowjK1}
}

@article{decomposingselfcorrection,
  title = {Decomposing {LLM} Self-Correction: The Accuracy-Correction Paradox and Error Depth Hypothesis},
  author = {Li, Yin},
  journal = {arXiv preprint arXiv:2601.00828},
  year = {2026},
  url = {https://arxiv.org/abs/2601.00828}
}

@article{scientificagentssurvey,
  title = {Towards Scientific Intelligence: A Survey of LLM-based Scientific Agents},
  author = {Ren, Shuo and Xie, Can and Jian, Pu and Ren, Zhenjiang and Leng, Chunlin and Zhang, Jiajun},
  journal = {arXiv preprint arXiv:2503.24047},
  year = {2025},
  url = {https://arxiv.org/abs/2503.24047}
}

@article{iterativeselfrepair,
  title = {How Many Tries Does It Take? {I}terative Self-Repair in {LLM} Code Generation Across Model Scales and Benchmarks},
  author = {Arimbur, Johin Johny},
  journal = {arXiv preprint arXiv:2604.10508},
  year = {2026},
  url = {https://arxiv.org/abs/2604.10508}
}

@article{scalingagentsystems,
  title = {Towards a Science of Scaling Agent Systems},
  author = {Kim, Yubin and Gu, Ken and Park, Chanwoo and Park, Chunjong and Schmidgall, Samuel and Heydari, A. Ali and Yan, Yao and Zhang, Zhihan and Zhuang, Yuchen and Liu, Yun and Malhotra, Mark and Liang, Paul Pu},
  journal = {arXiv preprint arXiv:2512.08296},
  year = {2025},
  url = {https://arxiv.org/abs/2512.08296}
}

@article{zhu2025abc,
  title = {Establishing Best Practices for Building Rigorous Agentic Benchmarks},
  author = {Zhu, Yuxuan and Jin, Tengjun and Pruksachatkun, Yada and Zhang, Andy and Liu, Shu and Cui, Sasha and Kapoor, Sayash and Longpre, Shayne and Meng, Kevin and Weiss, Rebecca and Barez, Fazl and Gupta, Rahul and Dhamala, Jwala and Merizian, Jacob and Giulianelli, Mario and Coppock, Harry and Ududec, Cozmin and Sekhon, Jasjeet and Steinhardt, Jacob and Kellermann, Antony and Schwettmann, Sarah and Zaharia, Matei and Liang, Percy and Kang, Daniel},
  journal = {arXiv preprint arXiv:2507.02825},
  year = {2025},
  url = {https://arxiv.org/abs/2507.02825}
}

@misc{openai_gptoss_release,
  author = {{OpenAI}},
  title = {Introducing gpt-oss},
  year = {2025},
  url = {https://openai.com/index/introducing-gpt-oss/},
  note = {Accessed: 2026-05-01}
}

@misc{openai_gptoss_model,
  author = {{OpenAI}},
  title = {gpt-oss-120b Model},
  year = {2025},
  url = {https://developers.openai.com/api/docs/models/gpt-oss-120b},
  note = {Accessed: 2026-05-01}
}

@article{cohen1960agreement,
  author  = {Cohen, Jacob},
  title   = {A Coefficient of Agreement for Nominal Scales},
  journal = {Educational and Psychological Measurement},
  year    = {1960},
  volume  = {20},
  number  = {1},
  pages   = {37--46},
  doi     = {10.1177/001316446002000104}
}

@article{landis1977observer,
  author  = {Landis, J. Richard and Koch, Gary G.},
  title   = {The Measurement of Observer Agreement for Categorical Data},
  journal = {Biometrics},
  year    = {1977},
  volume  = {33},
  number  = {1},
  pages   = {159--174},
  doi     = {10.2307/2529310}
}

@article{mchugh2012kappa,
  author  = {McHugh, Mary L.},
  title   = {Interrater Reliability: The Kappa Statistic},
  journal = {Biochemia Medica},
  year    = {2012},
  volume  = {22},
  number  = {3},
  pages   = {276--282},
  doi     = {10.11613/BM.2012.031}
}
